# Scalable Scene Modeling from Perspective Imaging:

# Physics-based Appearance and Geometry Inference

Dissertation

Presented in Partial Fulfillment of the Requirements for the Degree Doctor of Philosophy in the Graduate School of The Ohio State University

By

Shuang Song

Graduate Program in Civil Engineering

The Ohio State University

2024

Dissertation Committee

Dr. Rongjun Qin, Advisor

Dr. Charles Toth

Dr. Alper Yilmaz

## Abstract

3D scene modeling techniques serve as the bedrocks in the geospatial engineering and computer science, which drives many applications ranging from automated driving, terrain mapping, navigation, virtual, augmented, mixed, and extended reality (for gaming and movie industry etc.). This dissertation presents a fraction of contributions that advances 3D scene modeling to its state of the art, in the aspects of both appearance and geometry modeling. In contrast to the prevailing deep learning methods, as a core contribution, this thesis aims to develop algorithms that follow first principles, where sophisticated physic-based models are introduced alongside with simpler learning and inference tasks. The outcomes of these algorithms yield processes that can consume much larger volume of data for highly accurate reconstructing 3D scenes at a scale without losing methodological generality, which are not possible by contemporary complex-model based deep learning methods. Specifically, the dissertation introduces three novel methodologies that address the challenges of inferring appearance and geometry through physics-based modeling.

Firstly, we address the challenge of efficient mesh reconstruction from unstructured point clouds—especially common in large and complex scenes. The proposed solution employs a cutting-edge framework that synergizes a learned virtual view visibility with graph-cut based mesh generation. We introduce a unique three-step deep network that leverages depth completion for visibility prediction in virtual views, and an adaptive visibility weighting in the graph-cut based surface. This hybrid approach enables robust mesh reconstruction, overcoming the limitations of traditional methodologies and showing superior generalization capabilities across various scene types and sizes, including large indoor and outdoor environments.

Secondly, we delve into the intricacies of combining multiple 3D mesh models, particularly those obtained through oblique photogrammetry, into a unified high-resolution site model. This novel methodology circumvents the complexity of traditional conflation by using a panoramic virtual camera field and Truncated Signed Distance Fields. The result is a seamless handling of full-3D mesh conflations, which is a daunting task in standard geoscience applications due to intricate topologies and manifold geometries. The developed technique substantially enhances the accuracy and integrity of resultant 3D site models, empowering geoscience and environmental monitoring efforts.

Thirdly, the dissertation introduces a physics-based approach for the recovery of albedo from aerial photogrammetric images. This general albedo recovery method is grounded in a sophisticated inverse rendering framework that capitalizes on the specifics of photogrammetric collections—such as known solar position and estimable scene geometry—to recover albedo information accurately. The effectiveness of the approach is demonstrated through significant improvements not only in the realism of the rendered models for VR/AR applications but also in the primary photogrammetric processes. Such advancements have the potential to refine feature extraction, dense matching procedures, and the overall quality of synthetic environment creation.

Overall, the research encapsulated in this dissertation marks a series of methodological triumphs in the processing of complex datasets. By navigating the confluence of deep learning, computational geometry, and photogrammetry, this work lays down a robust framework for future exploration and practical application in the rapidly evolving field of 3D scene reconstruction. The outcomes of these studies are evidenced through rigorous experiments and comparisons with existing state-of-the-art methods, demonstrating the efficacy and scalability of the proposed approaches.

# Acknowledgments

I am profoundly grateful to a multitude of people whose support has been pivotal in the fruition of this dissertation.

Firstly, I extend my deepest appreciation to my advisor, Dr. Rongjun Qin, whose guidance, patience, and scholastic rigor provided the foundation upon which this work rests. His relentless pursuit of excellence drove me to push boundaries, and his unwavering support offered solace during the most challenging times. I would also like to thank my dissertation committee members, Dr. Charles Toth and Dr. Alper Yilmaz, for their insightful feedback and invaluable support throughout this process.

A heartfelt thanks goes to my colleagues and collaborators in the Geospatial Data Analytics Lab, who provided an outstanding work environment and collaborative support.

The endless love, patience, and encouragement I've received from my family, namely Mr. Binghui Song and Mrs. Yongxiang Cheng, have been my stronghold. My uncle, Dr. Yongqiang Cheng, has always believed in me, fueling my ambition to pursue a Ph.D. as aa monumental achievement in my life.

To my friends, Xupei Zhang, Yuan Yang, Yang Tang, thank you for the breaks from work that were vital to my wellbeing, the laughter that lightened my spirit, and the constant reassurance that the end goal was within reach.

This journey was not a solitary one, and the support I received along the way has left a lasting impact on both my professional work and personal development. Thank you.

## Vita

| | |
|---|---|
| **2014** | B.S., Geographic Information Science, Guangzhou University |
| **2017** | M.S., Photogrammetry and Remote Sensing, Wuhan University |
| **2018** | Algorithm Engineer, Shenzhen Intelligence Ally Technology Co., Ltd. |
| **2023** | M.S., Civil Engineering, The Ohio State University |
| **2018 - Present** | Ph.D. student & candidate, Civil Engineering, The Ohio State University |

## Publications

### Book Chapter

[1] Rongjun Qin, **Shuang Song**, Xiao Ling, Mostafa Elhashash (2020). 3D reconstruction through fusion of cross-view images. In Kwan, C. (Eds.), Recent Advances in Image Restoration with Applications to Real World Problems. IntechOpen.

### Journal Papers

[1] Sean Downey, Matthew Walker, Jacob Moschler, Filiberto Penados, William Peterman, Juan Pop, Rongjun Qin, Shane Scaggs, and **Shuang Song** (2023). An intermediate level of disturbance with customary agricultural practices increases

species diversity in Maya community forests in Belize. Communications Earth & Environment. DOI: 10.1038/s43247-023-01089-6.

[2] **Shuang Song**, Rongjun Qin (2023). Mesh Conflation of Oblique Photogrammetric Models using Virtual Cameras and Truncated Signed Distance Field. IEEE Geoscience and Remote Sensing Letters. DOI 10.1109/LGRS.2023.3298321

[3] Guixiang Zhang[*], **Shuang Song**[*], Jenny Panescu, Nicholas Shapiro, Karen C. Dannemiller, Rongjun Qin (2023). A novel systems solution for accurate colorimetric measurement through smartphone-based augmented reality. PLOS ONE 18, no. 6 (2023): e0287099.

[4] Ningli Xu, Rongjun Qin, **Shuang Song** (2023). Point Cloud Registration for LiDAR and Photogrammetric Data: a Critical Synthesis and Performance Analysis on Classic and Deep Learning Algorithms. ISPRS Open Journal of Photogrammetry and Remote Sensing, 100032.

[5] Yang Tang, **Shuang Song**, Shengxi Gui, Weilun Chao, Chinmin Cheng, Rongjun Qin (2023). Active and Low-Cost Hyperspectral Imaging for the Spectral Analysis of a Low-Light Environment. Sensors, 23(3), 1437.

[6] Xu, Ningli, Debao Huang, **Shuang Song**, Xiao Ling, Chris Strasbaugh, Alper Yilmaz, Halil Sezen, and Rongjun Qin (2021). A volumetric change detection framework using UAV oblique photogrammetry–a case study of ultra-high-resolution monitoring of progressive building collapse. International Journal of Digital Earth: 1-16.

---

[*] Equal contributions

[7] Yang Xu[*], Adil M. Rather[*], **Shuang Song**[*], Jen-Chun Fang, Robert L. Dupont, Ufuoma I. Kara, Yun Chang, Joel A. Paulson, Rongjun Qin, Xiaoping Bao, Xiaoguang Wang (2020). Ultrasensitive and Selective Detection of SARS-CoV-2 using Thermotropic Liquid Crystals and Image-based Machine Learning. Cell Reports Physical Science: 100276.

[8] **Shuang Song**, Rongjun Qin. A General Albedo Recovery Approach for Aerial Photogrammetric Images through Inverse Rendering, ISPRS Journal of Photogrammetry and Remote Sensing (submitted).

**Conference Papers**

[1] **Shuang Song**, Rongjun Qin (2022). A Novel Intrinsic Image Decomposition Method to Recover Albedo for Aerial Images in Photogrammetry Processing. ISPRS. Annals. Photogramm. Remote Sens. Spatial Inf. Sci.

[2] **Shuang Song**, Zhaopeng Cui, Rongjun Qin. Vis2Mesh: Efficient Mesh Reconstruction from Unstructured Point Clouds of Large Scenes with Learned Virtual View Visibility. International Conference on Computer Vision (ICCV) 2021.

[3] **Shuang Song**, Rongjun Qin (2020). Optimizing mesh reconstruction and texture mapping generated from a combined side-view and over-view imagery. ISPRS. Annals. Photogramm. Remote Sens. Spatial Inf. Sci. 2, 403-409.

---

[*] Equal contributions

## Fields of Study

Major Field: Civil Engineering

Studies in:

Topic 1: Photogrammetry

Topic 2: Surface reconstruction

Topic 3: Deep learning in Photogrammetry Applications

Topic 4: Inverse Rendering

## Table of Contents

# List of Tables

## List of Figures

# Chapter 1. Introduction

This chapter covers the motivation of work presented in this dissertation. Firstly, Section 1.1 presents the main motivations behind the work and challenges for the topic 3D modeling and scene reconstruction. Further, this chapter states the research questions to define the research scope of the dissertation (Section 1.2). The main contributions of the dissertation are presented in Section 1.3. Finally, Section 1.4 describes the structure of the rest of this dissertation.

## 1.1 Motivation

The last few decades have seen remarkable progress in the field of 3D modeling and scene reconstruction, primarily driven by advancements in drone photogrammetry, machine learning and computer vision. These technologies are widely used in numerous of applications, ranging from virtual and augmented reality on personal devices to professional geospatial sciences and urban planning. The ability to accurately and efficiently create 3D representations of physical environments has profound implications for both theoretical research and practical applications. Scaling research work to accommodate other datasets with varying scales and modality requires a thoughtfully designed framework. The primary hurdle in scaling is often the intensive global optimization or complex end-to-end neural networks. Therefore, the three pieces of research presented in this dissertation share a common design philosophy: breaking down the problem into manageable parts to enable an iterative approach that can handle data of any size. To tackle challenges in scene modeling, we have explored the integration of classic methods with deep learning and physics-based modeling.

**Challenges in 3D Scene Reconstruction from Point Clouds**

With the advent of sophisticated 3D sensors, such as LiDAR and depth cameras, capturing detailed 3D point clouds has become more accessible. However, transforming these point clouds into usable 3D models, especially for large and complex scenes, remains a formidable challenge. Traditional methods struggle with issues like varying point densities, incomplete data, and the need for scalability to large scenes. End-to-end schema are highly favored by machine learning community due to its simplicity and modularity. However, most of them suffers from overfitting and extremely limited running memory issue. There is a significant need for a hybrid approach that leverages both traditional techniques and modern deep learning methods to address these challenges. Our novel solution proposed focuses on solving visibility determination problem of 2D depth image using the rather mature binary classification CNN (Convolutional Neural Network) architectures. Then, we utilized the learned virtual view visibility to enhance the quality and scalability of mesh reconstruction, offering a promising direction in overcoming the current limitations.

**The Complexity of Mesh Conflation**

The next critical aspect of this dissertation focuses on mesh model conflation. In fields like geoscience, where large-scale 3D models are essential, the process of combining multiple mesh models into a coherent and accurate representation is fraught with difficulties. The complex topology of full-3D triangle meshes, differences in resolution, and varying levels of uncertainty among models from different sources add layers of complexity to the conflation process. To tackle on this challenge, we introduce an innovative method using virtual cameras and truncated signed distance fields (TSDF) to achieve a more accurate and reliable mesh conflation, a significant advancement over traditional overlay and averaging techniques.

**Albedo Recovery in Aerial Photogrammetry**

The third part shifts focus to a specific challenge in aerial photogrammetry: the recovery of albedo from high-resolution images taken under natural light. Albedo recovery is crucial for enhancing the realism and applicability of photogrammetric models in simulations and immersive applications. The traditional texturing methods, which often carry lighting artifacts from source images, limit the utility of these models. This work presents a physics-based approach to albedo recovery, leveraging the unique aspects of photogrammetric image collection, such as known capture times and locations, to model outdoor lighting effectively. This approach has significant implications for improving the quality of photogrammetric models and expanding their application scope.

The dissertation unifies these distinct yet interrelated areas under the broader theme of enhancing 3D modeling and scene reconstruction. Each chapter not only addresses specific challenges within its domain but also contributes to the overarching goal of advancing the field of photogrammetry and computer vision. By integrating traditional methods with novel computational approaches, those research pushes the boundaries of what is possible in 3D modeling and scene reconstruction, paving the way for more accurate, efficient, and versatile applications.

## 1.2 Research Scope

This dissertation explores different techniques to enhance the appearance and geometry of 3D scene based on challenges mentioned above (Section 1.1). In addition, this dissertation provides algorithms that tackle these challenges. Precisely, the primary research questions are:

- What makes classic photogrammetry success in mesh reconstruction from multi-view image set? (Chapter 2)
- How to reconstruct a mesh as good as photogrammetry pipeline if only point clouds (only contains XYZ coordinates, without any other attributes) present? (Chapter 2)
- How can we efficiently utilize the deep learning techniques in scale-agnostic mesh reconstruction? (Chapter 2)
- How to create a seamless, topologically correct mesh from multiple triangle meshes with different resolutions and levels of uncertainty. (Chapter 3)
- What are shading components of the outdoor natural illumination? (Chapter 4)
- Given aerial image set acquired under outdoor light, how to estimate the nature lighting effects and remove it from image? (Chapter 4)

## 1.3 Contributions

Our research has made substantial progress in scalable scene modeling. We developed innovative physics-based methods for 3D scene reconstruction, mesh conflation and albedo recovery. These contributions provide a strong foundation for future research and practical use. In particular:

For ***3D Scene Reconstruction***, we introduce a pioneering framework for mesh reconstruction from unstructured point clouds, coupling traditional approaches with learning techniques for enhanced performance. The systematic approach overcomes hurdles associated with varying point density, complexities of different scene contexts, and scalability, thereby providing a comprehensive solution for robust and generalized mesh reconstruction. The method's favorability towards transferability, robustness, and superior performance highlights its potential to revolutionize 3D surface reconstruction, fostering broad practical application across computer vision and virtual reality realms.

For ***Mesh Conflation***, we contribute an innovative method for conflation of full 3D oblique photogrammetry models, enabling the seamless merging of intricate mesh models into a unified, high-resolution landscape representation. Overcoming the complexities associated with differing resolutions and uncertainties, this solution leverage TSDF and virtual cameras to address a long-standing challenge in geoscience applications. The approach presents a breakthrough, unlocking new possibilities for creating accurate full 3D mesh models, thereby expanding the horizons in geoscience and environmental fields.

For ***Albedo Recovery***, a pioneering approach for albedo recovery from aerial photogrammetric images through inverse rendering is proposed to address the critical need for realistic scene modeling. By devising a physics-based model leveraging estimable solar illumination and scene geometry, we proposed an approach to resolve the albedo information from source images. Significantly extending the current state-of-the-art, this work not only enhances the use of photogrammetry models across a wide range of applications, but also presents a step-change in the enhancement of photogrammetry processing techniques.

## 1.4 Organization

In this dissertation, we have explored several key issues in enhancing the quality of textured meshes through photogrammetry, computer vision, and computer graphics. The following is the structure of the dissertation: Chapter 2 discusses our hybrid 3D scene reconstruction framework, which combines traditional graph-cut based surface reconstruction with deep neural networks. Chapter 3 presents our TSDF mesh conflation workflow, which merges multiple meshes into a single, topologically correct, and seamless mesh. In Chapter 4, we introduce our approach to albedo recovery using inverse rendering techniques. Finally, Chapter 5 concludes the dissertation with a summary of the major findings and a discussion of the research questions.

# Chapter 2. Efficient Mesh Reconstruction from Unstructured Point Clouds of Large Scenes with Learned Virtual View Visibility

This chapter is based on the paper called “Vis2Mesh: Efficient Mesh Reconstruction From Unstructured Point Clouds of Large Scenes With Learned Virtual View Visibility” that was published in the “IEEE/CVF International Conference on Computer Vision (ICCV 2021)” by Shuang Song, Zhaopeng Cui, and Rongjun Qin. Figure 2.1 shows two reconstruction examples with our approach.

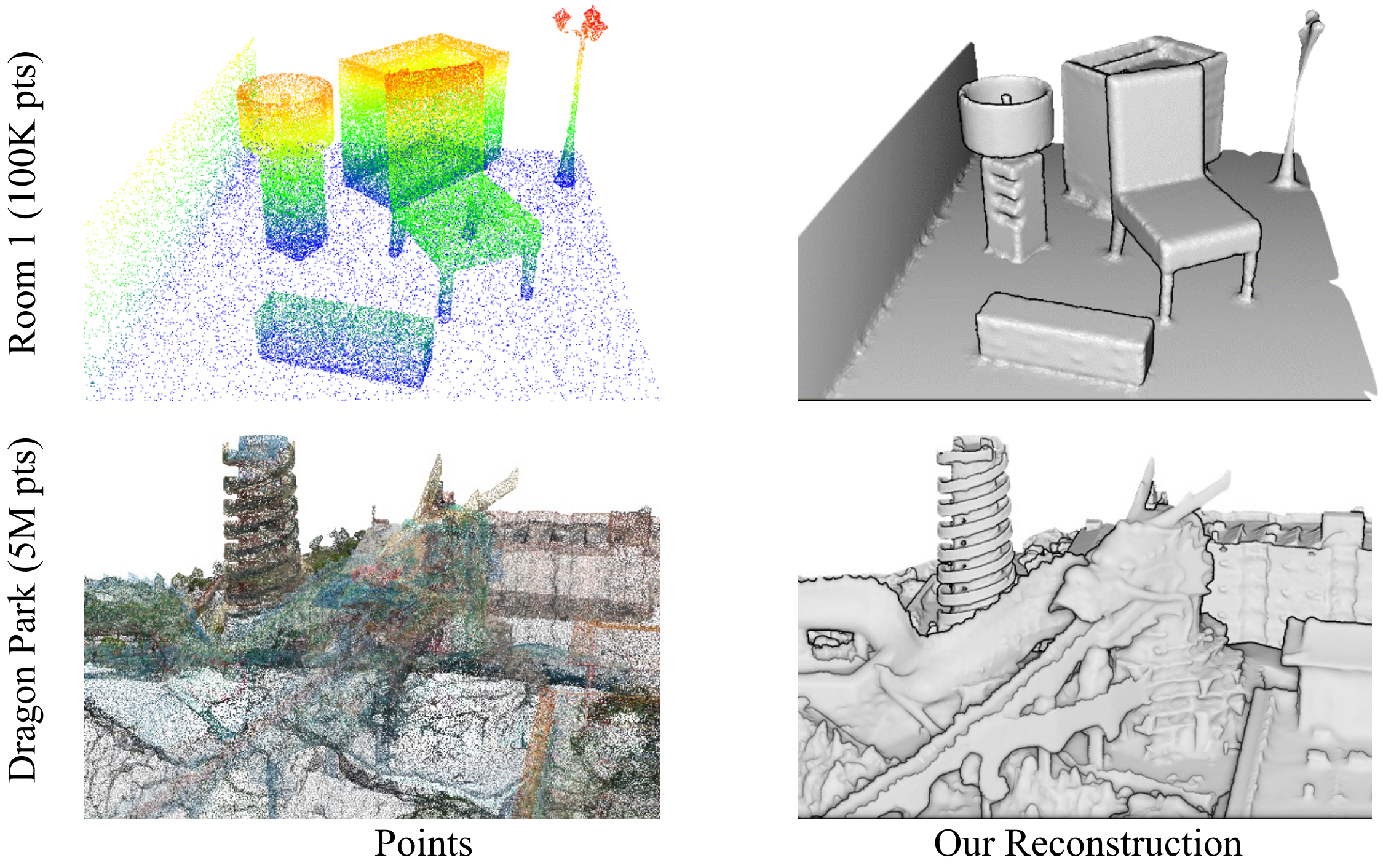

Figure 2.1 Examples of reconstructed surfaces with our approach on both indoor and large outdoor scenes. The number of points ranges from thousands to millions. Colors of point cloud are attached for visualization purpose.

## 2.1 Chapter Abstract

We present a novel framework for mesh reconstruction from unstructured point clouds by taking advantage of the learned visibility of the 3D points in the virtual views and traditional graph-cut based mesh generation. Specifically, we first propose a three-step network that explicitly employs depth completion for visibility prediction. Then the visibility information of multiple views is aggregated to generate a 3D mesh model by solving an optimization problem considering visibility in which a novel adaptive visibility weighting in surface determination is also introduced to suppress line of sight with a large incident angle. Compared to other learning-based approaches, our pipeline only exercises the learning on a 2D binary classification task, i.e., points visible or not in a view, which is much more generalizable and practically more efficient and capable to deal with a large number of points. Experiments demonstrate that our method with favorable transferability and robustness, and achieve competing performances w.r.t state-of-the-art learning-based approaches on small complex objects and outperforms on large indoor and outdoor scenes.

## 2.2 Introduction

3D surface reconstruction is essential to drive many computer vision and VR/AR applications, such as vision-based localization, view rendering, animation, and autonomous navigation. With the development of 3D sensors, LiDAR or depth sensors, we can directly capture accurate 3D point clouds.

However, determining surfaces from unstructured point clouds remains a challenging problem, especially for point clouds of objects of complex shapes, varying point density, completeness, and volumes. A favorable mesh reconstruction method should be 1) capable of recovering geometric details captured by the point clouds; 2) robust and generalizable to different scene contexts; 3) scalable to large scenes and tractable in terms of memory and computation.

Typical reconstruction methods either explicitly explore the local surface recovery through connecting neighboring points (e.g., Delaunay Triangulation (Boissonnat and Yvinec 1998)), or sort solutions globally through implicit surface determination (e.g., Screened Poisson Surface Reconstruction (Kazhdan, Bolitho, and Hoppe 2006; Kazhdan and Hoppe 2013)). However, both of them favor high-density point clouds. With the development of 3D deep learning, many learning-based methods have demonstrated their strong capability when reconstructing mesh surfaces from moderately dense or even comparatively sparse point clouds of complex objects. However, these end-to-end deep architectures, on one hand, encode the contextual/scene information which often runs into generalization issues; on the other hand, the heavy networks challenge the memory and computations when scaling up to large scenes.

The point visibility in views has been shown as a good source of information, which can greatly benefit surface reconstruction (S. Song and Qin 2020). For example, a 3D point visible on a physical image view alludes to the fact the line of sight from the 3D point to the view perspective center, travels through free space (no objects or parts occludes), which as a result, can serve as a strong constraint to guide the mesh reconstruction. Intuitively, the more visible a point is in more views, the more information about the free space between points and views that one can explore (M. Jancosek and Pajdla 2011). This principle has been practiced in multi-view stereo (MVS) and range images and shows promising results (Labatut, Pons, and Keriven 2007; 2009; Vu et al. 2012), in which the visibility of each 3D point to each physical is recorded through dense matching (in MVS) or decoded directly from the sensors. However, in both cases, the visibility is limited by the number of physical views and accounts for only limited knowledge of the free space, thus leading to the lack of reconstruction details.

To address it, we propose a novel solution by generating a large number of virtual views around the point clouds and utilize the learned visibility through a graph-cut based surface reconstruction approach. This will literally give us unlimited visibility information for high-quality mesh generation through traditional approaches based on visibility. However, it is non-trivial to design such a system. First, visibility prediction in virtual views is not easy as the projected pixels from the point clouds to different views can have varying sparsity. Second, unlike the MVS views which normally assume a good incidence angle to the generated point clouds, the generated virtual views may present a large variation in terms of their incident angle and distance to the surface.

To solve these problems, we design a network for visibility prediction, which utilizes partial convolution to model the sparsity of the input in the neural network. To further improve the accuracy, we propose a cascade network that explicitly employs depth completion as an intermediate task. Secondly, to suppress the side effect of unfavored virtual rays (large-incidence angle to the surface), we propose a novel adaptive visibility weighting term that adaptively weights the virtual rays. Compared to the end-to-end learning-based methods, our method has better generalization across different scene contexts because a very simple and learning-friendly task, i.e., visibility prediction, is involved in our pipeline. Moreover, our method also maintains the capability to process an extremely large volume of outdoor point clouds with high quality.

Our contributions can be summarized as follows:

- Firstly, we present a surface reconstruction framework from unstructured point clouds, that exploits the power of both traditional and learning-based methods.
- Secondly, we propose a three-step network that explicitly employs the depth completion for the visibility prediction of 3D points.
- Thirdly, we propose a novel adaptive visibility weighting term into the traditional graph-cut based meshing pipeline, to increase the geometric details.
- Lastly, we demonstrate through our experiments that our proposed method yields better generalization capability, and can be scaled up to process point clouds of a large scene. It is also robust against different types of noises and incomplete data and has better performance than the state-of-the-art methods.

### 2.3 Related Works

**Classic surface reconstruction** from unstructured point clouds exploits both local and global surface smoothness, global regularity, visibility, etc. The *local surface smoothness* based methods (Edelsbrunner and Mücke 1994; Hoppe et al. 1992; Alexa et al. 2003) seek for smooth surface only in close proximity to the data.

The *global surface smoothness* based methods (Kazhdan 2005; Carr et al. 2001; Kazhdan, Bolitho, and Hoppe 2006; Kazhdan and Hoppe 2013; Kolluri, Shewchuk, and O'Brien 2004) tend to find a field function (signed distance function, indicator function) approximating the point cloud. The most common global smoothness algorithm is Screened Poisson Surface Reconstruction (SPSR) (Kazhdan, Bolitho, and Hoppe 2006; Kazhdan and Hoppe 2013) which can generate smooth, void-free surfaces with high-quality oriented normals, while it can be oversensitive to incorrect normals, as well as non-watertight shapes.

The *global regularity* based methods (Oesau, Lafarge, and Alliez 2016; Zheng et al. 2010; Y. Li et al. 2011) deal with the man-made objects or architectural shapes that feature symmetry, orthogonality, repetition, and parallelism.

The *visibility* based methods assume the information of the sensors from which the point clouds are collected, and this yields additional cues as rays and visibility of each point at sensor locations in which they are observable. Varying forms of visibility (includes lines of sight (Curless and Levoy 1996; Labatut, Pons, and Keriven 2007) , exterior visibility (Shalom et al. 2010; Katz, Tal, and Basri 2007) parity (Nagai, Ohtake, and Suzuki 2015)) have been explored. The most widely used methods utilize lines of sight (Labatut, Pons, and Keriven 2007; 2009; M. Jancosek and Pajdla 2011; Michal Jancosek and Pajdla 2014; Y. Zhou, Shen, and Hu 2019) that determine the surface between interior and exterior tetrahedron of the Delaunay triangulation of a point set by solving a *s-t* graph which is weighted by lines of sight. Nonetheless, the quality of results relies on the number and quality of rays and it may fall short in concave areas on the reconstructed surface if insufficient rays are identified (Y. Zhou, Shen, and Hu 2019).

**Learning-based surface reconstruction** takes a data-driven approach that takes advantage of the available examples and the high-capability network structures (Charles

R Qi et al. 2017; Charles Ruizhongtai Qi et al. 2017; Çiçek et al. 2016; Y. Wang et al. 2019) to build surfaces for challenging and complex point clouds that classic methods have poor performance on. Several methods (Chen and Zhang 2019; Mescheder et al. 2019; Park et al. 2019) encode the entire object to an embedding vector and decode it with a neural network, as a general problem in such deep networks, the representation often cannot be generalized to objects of unseen categories or scenes. CONet (Peng et al. 2020) proposed alternative volumetric representations and enabled convolution in the implicit field making the representations invariant to translation, and the sliding-window nature of this approach makes CONet capable of scaling up to large scenes, while the method being local does not capture global shape information and the volumetric representations are still memory demanding for 3D data generation. Points2Surf (Erler et al. 2020) proposed a patch-based learning framework combining outputs of local-scale and global-scale neural networks to construct a signed distance function (SDF) and shows a certain level of generalization capability. Yet the query on the per-voxel level and the inference over the voxels field are computationally intensive (4x slower than SPSR), which limits it to only process small scenes.

The Meshing with Intrinsic-Extrinsic Ratio (MIER) method (M. Liu, Zhang, and Su 2020) uses the explicit representation of input points and cast the connectivity to a surrogate comparing geodesic/Euclidean distances. A neural network-based classifier is trained to determine surface triangles, which has shown to be capable of preserving fine-grained details and works well on ambiguous structures. However, the per-triangle classification overlooks the connectivity of adjacent triangles, causing holes on the reconstructed surfaces.

Point2Mesh (Hanocka et al. 2020) proposed a self-prior to detect self-similarity, which contributes to reconstructing surface as well as removing noises and completing missing parts. Nonetheless, Point2Mesh only works with watertight and simple models, requires an initial mesh model, and comparably needs a long running time to converge. As compared to the classic methods, the learning-based methods, in general, demand a high volume of memory and can be scene-specific.

## 2.4 Preliminary

This section introduces the formula of Delaunay triangulation and graph-cut optimization based surface reconstruction (Labatut, Pons, and Keriven 2009).

The surface reconstruction problem is considered as the outer space/inner space labeling problem of the connected graph built based on Delaunay triangulation, to be solved by minimum *s-t* cut algorithm (Boykov, Veksler, and Zabih 2001). The reconstructed surface, denoted by $S$ in the following, is a union of oriented Delaunay triangles which are guaranteed to be watertight and intersection free as it bounds a volume consist of tetrahedra. The energy function of the reconstruction problem is formulated by the basic method (Labatut, Pons, and Keriven 2009) as Equation (2.1) and Equation (2.2).

$$E(S) = E_{vis}(S,V) + \lambda_{vis} E_{ql}(S) \,, \tag{2.1}$$

$$E(S) = \sum_{v \in V} \left[ E_s(v) + E_t(v) + E_{ij}(v) \right] + \lambda_{ql} E_{ql}(S) \,, \tag{2.2}$$

where $E_{vis}(S, V)$ is a sum of penalties for conflicts and wrong orientations of the surface $S$ concerning the constraints imposed by all lines of sight $V$, $E_{ql}(S)$ is an additional smoothness term to suppress skinny triangles (Labatut, Pons, and Keriven 2009). In graph-cut, $E_s$ and $E_t$ are data terms for outer and inner space respectively, and $E_{ij}$ is a smoothness term. $E_{ql}(S)$ is another smoothness term independent of visibility.

$$E_s(v) = \infty \cdot \delta[T_1 \in t], \tag{2.3}$$

$$E_t(v) = \alpha_{vis} \cdot \delta[T_{M+1} \in s], \tag{2.4}$$

$$E_{ij}(v) = \sum_{i=1}^{M-1} \alpha_{vis} \cdot \delta[T_i \in s \wedge T_{i+1} \in t], \tag{2.5}$$

where $T_i$ is the $i$-th tetrahedron intersects with the line of sight $v$, $M$ is the size of the intersection set. So that $T_1$ and $T_{M+1}$ denote the first and the one right behind the last tetrahedra along $v$. The conditional operator $\delta[\cdot]$ is 1 if the condition is true, otherwise is 0. $\alpha_{vis}$ is the confidence of the visibility, which is proportional to the distance to the endpoint, which is called soft visibility constraint.

To illustrate equations with their geometric meaning, As shown in Figure 2.2, a line of sight traverse a series of tetrahedra (where $M = 5$), from $T_1$ to $T_5$ and extended to $T_6$ ($T_{M+1}$ in the main paper) which is the tetrahedron right behind the end point $p$. The weighting process starts from the point of view ($c$) marking it as the source node (Equation (2.3)). Along the line of sight, accumulate the value $\alpha_v$ to the connectivity of facets that line of sight passes through from the front (Equation (2.5)). As for $T_6$ behind the point, it is marked as a sink node with constant weight (Equation (2.4)).

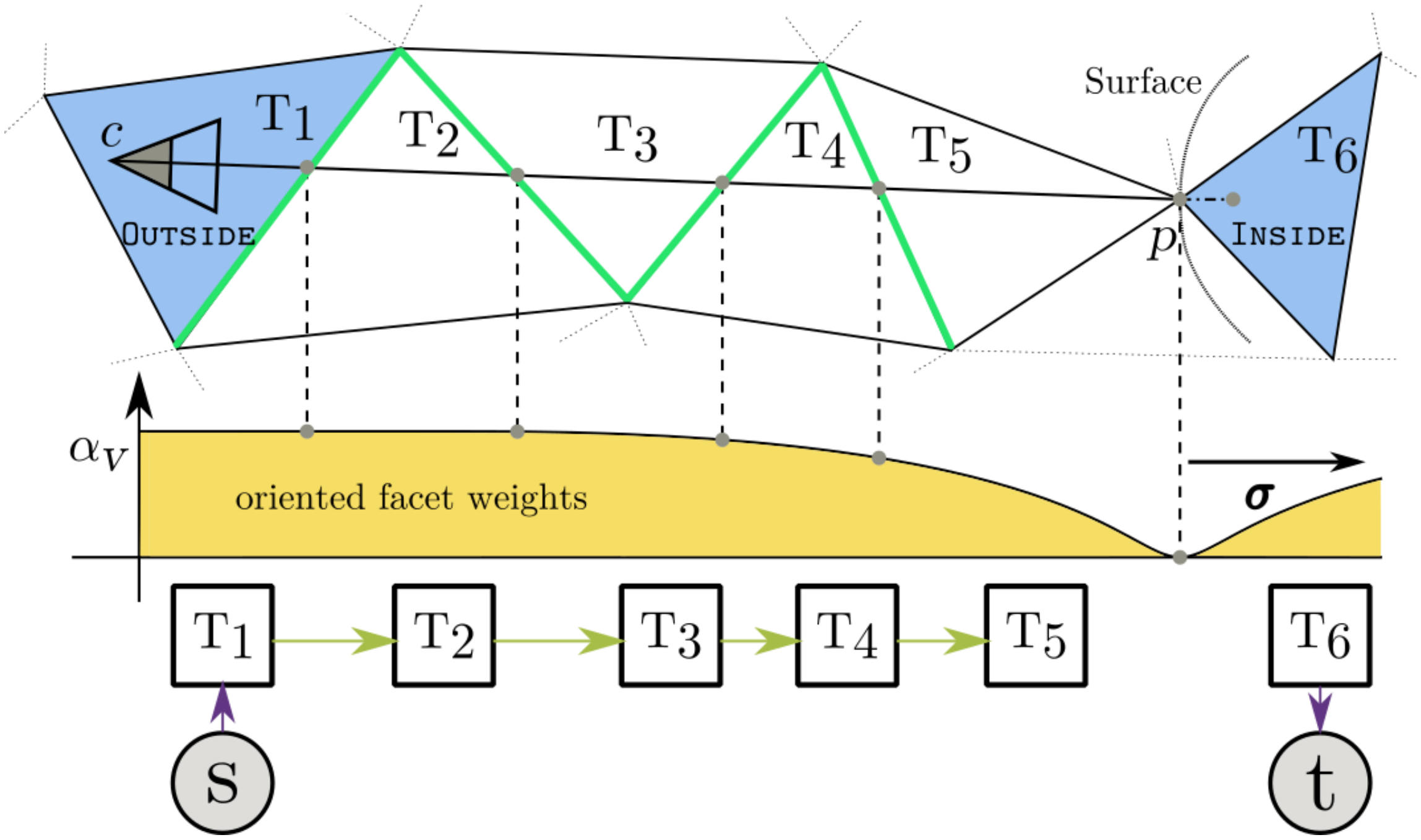

Figure 2.2 Visibility and graph construction (Labatut, Pons, and Keriven 2009). From top to bottom, **Upper:** A line of sight from a reconstructed 3D point traverses a sequence of tetrahedra, the graph construction, and the assignment of weights to the tetrahedron and oriented facets. **Middle:** Soft visibility decay along the line of sight which is inversely proportional to the distance to the end point. **Lower:** Corresponding s-t graph and the cut solution.

### 2.5 Method

In this section, we introduce our framework of surface reconstruction from unstructured point clouds (Figure 2.3). Firstly, we sample feasible 6 DoF (Degree of Freedom) poses according to input point clouds. Then, a 3D renderer will take poses as viewing parameters and project points onto that virtual image plane based on its projection matrix. Next, a three-step network that explicitly employs depth completion to assist visibility estimation and predicts the visible and occluded points to build visibility information. Finally, the graph-cut based solver processes the input points associated with predicted visibility to reconstruct the surface. Our overall framework is shown in Figure 2.3, and the individual components will be discussed in the following subsections.

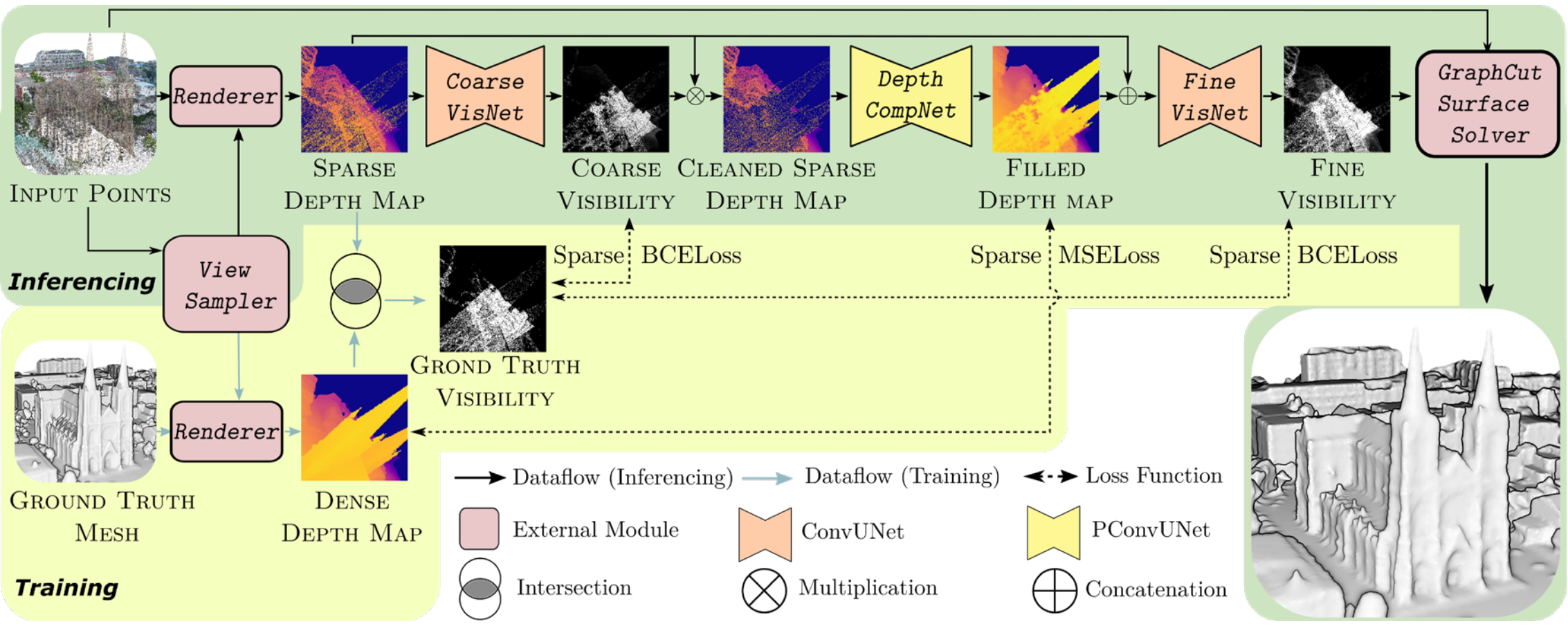

Figure 2.3 The proposed Vis2Mesh framework. Our framework reconstructs surfaces in *four* steps shown in inference region from left to right follow the flow of arrows: 1) Virtual view sampling 2) Rendering 3) Visibility determination and 4) Delaunay and graph-cut based surface reconstruction. Visibility determination composed by three networks: **CoarseVisNet**, predicting visibility from sparse depth input, **DepthCompNet** completing dense depth map based on coarse predicted visible points and **FineVisNet** refining visibility prediction with sparse depth and completed dense depth map, detailed in Section 2.5.2.

### 2.5.1 Virtual View Sampling

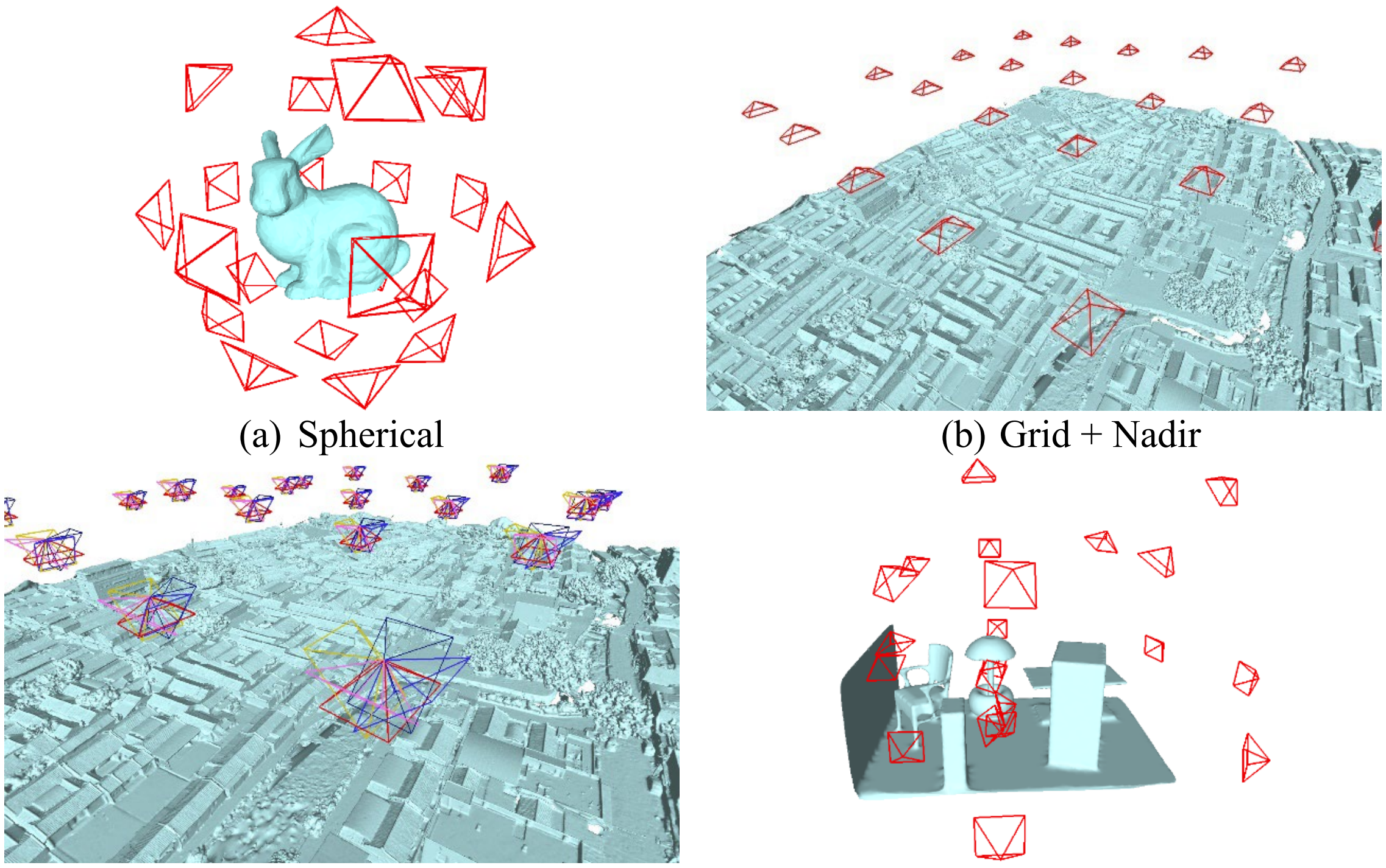

(a) Spherical (b) Grid + Nadir (c) Grid + Oblique (d) Customized

Figure 2.4 Virtual view generators. **Spherical Pattern Generator**: samples around the object while targeting it, always used for small objects. **Grid + Nadir Generator**: samples in grid fashion that parallel to the ground plane for the terrain-like large-scale scenario, uses above ground height and overlap rate to control the density of sampled views. **Grid + Oblique generator**: beside nadir views, samples concentric oblique views to capture facades in the urban area. **Customized**: we create an interactive tool to let users pick views. Customized views can improve the quality of ambiguous structures of complex objects.

Given the input unstructured point cloud, we first generate the images under virtual views that simulate the input images in the MVS method. Virtual view sampling given a presumed object or scene is a challenging problem, and automatic view selection for tasks such as stereo reconstruction or surveillance is an active topic in computer vision and robotics community (Pito 1999; N. Smith et al. 2018; Zeng, Zhao, and Liu 2020;

Zeng et al. 2020), and the performance of these methods are task and scene-specific. In our work, we utilize several general generators as shown in Figure 2.4 and empirically find that they work very well with our method when dealing with different objects and scene contexts. In practice, our method shows a certain level of robustness regarding varieties of virtual view sampling, and the number of virtual views can affect surface quality, the analysis of which can be found in Section 2.6.4.

### 2.5.2 Visibility Estimation Network

Given the projected image with sparse depth information, we exploit the deep neural network to predict the visibility of 3D points in the virtual views. The biggest challenge of the visibility estimation task is that it requires pixel-wise accuracy rather than region-wise. To this end, the standard CNNs (Convolutional Neural Networks) with small kernel size is preferred for the task because it is sensitive to local structure. However, for data with varying sparsity levels, the standard CNNs have poor performance since invalid pixels or voids produced by projecting sparse points make it difficult to learn robust representations (Uhrig et al. 2017).

Partial Convolution (PConv) (G. Liu, Shih, et al. 2018; G. Liu, Reda, et al. 2018) is proposed for image inpainting from an incomplete input image. The proposed re-weighted convolution operation is particularly suitable for processing sparse data since it models the validity of each pixel and propagates it along with the convolutional operation. A similar idea has been used in depth completion (Uhrig et al. 2017) from sparse laser scan data. Visibility prediction exploits the local sensitivity of neural networks, while depth completion requires local smoothness. Therefore, we design an intermediate task with a dedicated network module for depth completion and proposed a cascade network VDVNet (Visibility-Depth-Visibility).

The input of our visibility estimation network is a $H \times W \times 2$ feature map in which the first channel is a normalized depth image, and the second is a binary mask that indicates whether the pixel is valid. Our frame shown in Figure 2.3 includes an example input and intermediate results between different network components (which will be detailed in the following sections).

**CoarseVisNet.** We train the sub-network to predict the visibility with supervised learning. As shown in Figure 2.3, we apply Sparse BCELoss (binary cross-entropy loss with mask) between the predicted coarse visibility map and the ground truth visibility map, where the mask indicates the valid pixels.

**DepthCompNet.** The depth completion sub-network is designed to convert a sparse depth map to a dense one. We filter raw depth map with coarse visibility predicted by CoarseVisNet before feeding it to DepthCompNet. The supervision of DepthCompNet is depth maps of ground truth surface. Mean squared error loss function is applied between them (Sparse MSELoss in Figure 2.3).

**FineVisNet.** This module takes the input of CoarseVisNet and the output of DepthCompNet to predict the fine visibility. The loss function is applied as same as CoarseVisNet.

**Training Data.** The training data used for visibility prediction can be easily simulated from any type of 3D model. We select a few well-reconstructed textured mesh models of large-scale scenes from a public MVS reconstruction dataset (Yao et al. 2020) as the *ground truth surfaces* to generate synthetic visibility. Points are uniformly and sparsely sampled from surfaces. Then, the sampled points are further augmented with additional Gaussian white noises and outliers. In the next step, we sample virtual views from the point clouds with the combination of patterns shown in Figure 2.4. For each view, a depth map of both point clouds and surface are rendered and combined to generate the *ground truth visibility*. For more details, please refer to Section 2.6.1.

If the depth value of the point rendered image and the surface rendered image are matched, the point is labeled as visible, otherwise, as occluded. Depth maps of the surfaces serve as *ground truth depth* for the training of DepthCompNet.

### 2.5.3 Adaptive Visibility Weighting of Graph-Cut based Surface Reconstruction

Once the visibility of points in all views is predicted, we utilize the graph-cut based mesh generation (Labatut, Pons, and Keriven 2009) to reconstruct the final mesh model. Due to the different characteristics of virtual visibility, the basic method (Labatut, Pons, and Keriven 2009) suffers from the over-smoothing issue of creases on the surface, which motivates us to propose adaptive visibility weighting. The formulas of the basic method can be found in Section 2.4.

Compared with the visibility from instruments, the virtual view visibility generated by our VDVNet from unordered points and sampled views has very different characteristics in terms of density, direction, and distance. As Equation (2.4) indicates, the basic method has an issue that the strong constraint imposed on the tetrahedra right behind the point along the line of sight and might overlook details of the surface especially when the lines of sight have a large incident angle. Since the large incident angle is rare in traditional applications, the issue of the baseline method was not discovered before.

Since our virtual visibility is predicted by the neural network from unordered points directly, the angle of the incidence can range from 0 to 90 degrees, because there is no implicit filter of the multi-stereo view method. The visibility with a large incident angle (Figure 2.5a) is equivalent to a sharp surface in the depth map of the view, which is always considered as a sign of self-occlusion (Bódis-Szomorú, Riemenschneider, and Van Gool 2015).

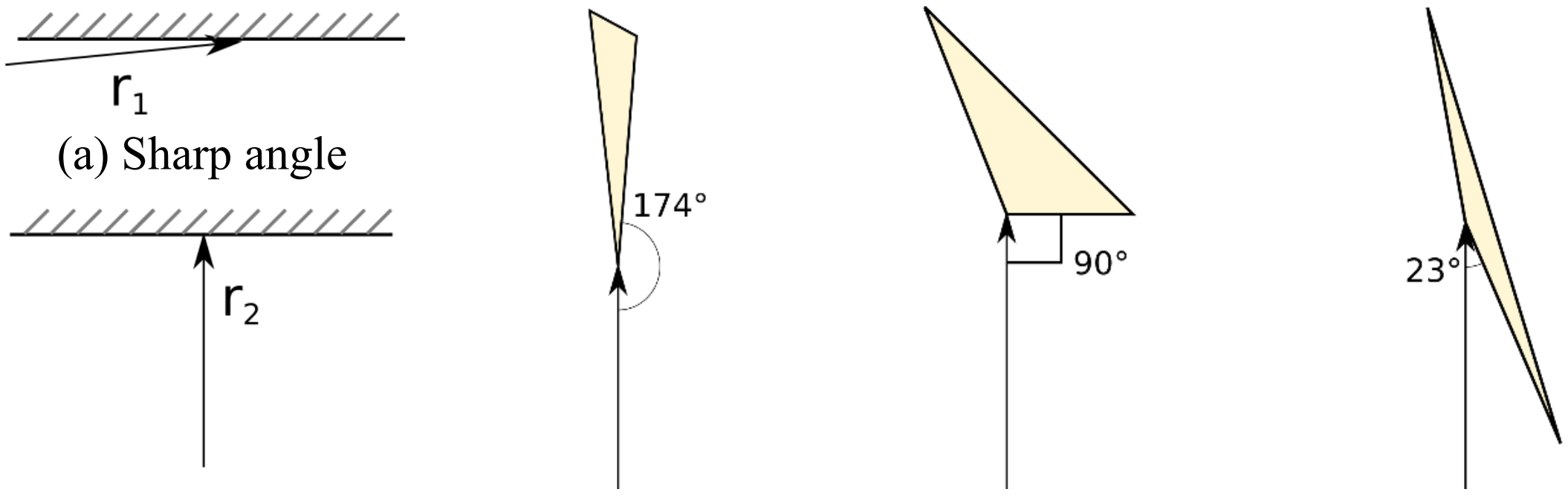

(b) Fronto-parallel (c) $\gamma = 0.1$ (d) $\gamma = 1.0$ (e) $\gamma = 0.4$

Figure 2.5 Geometric illustration of adaptive visibility weighting, where $\lambda_{avw} = 1$.

The key idea of our adaptive visibility weighting (AVW) is illustrated in Figure 2.5. Apparently, for surface reconstruction, a measurement for a sharp angle surface (Figure 2.5a) is considered as the lowest confidence, while for the fronto-parallel surface (Figure 2.5b) is considered the best confidence. We use cosine similarity between the direction of the line of sight and the surface normal to construct a smooth weighting function. The weight is regarded as the confidence of visibility, its definition overlaps with $\alpha_{vis}$, and it changes terms $E_t$ and $E_{ij}$. We use the product of two factors $\gamma$ and $\alpha_{vis}$ as the new confidence of line of sight. In order to apply the weighting function to a tetrahedron with 3 triangles where the vertex and the line of sight intersect, we consider the one most likely to be on the surface, that is, the triangle whose normal is closest to the direction of the ray. We took the maximum value of the cosine similarity of the three angles, as shown in Equation (2.6) - (2.8) , where the cosine similarity is equivalent to the dot product because the normal is of unit length. Figure 2.5a, Figure 2.5b and Figure 2.5c show examples of our adaptive visibility weighting term.

$$E_t^{OURS}(v) = \gamma \cdot \alpha_{vis} \cdot \delta[T_{M+1} \in s], \tag{2.6}$$

$$E_{ij}^{OURS}(v) = \sum_{i=1}^{M-1} \gamma \cdot \alpha_{vis} \cdot \delta[T_i \in s \wedge T_{i+1} \in t], \tag{2.7}$$

$$\gamma = (1 - \lambda_{avw}) + \lambda_{avw} \max\left(N_v^T \cdot \left[N_{f1}, N_{f2}, N_{f3}\right]\right), \tag{2.8}$$

where $N_v$ denotes direction of the incident line of sight $v$, $N_{f\#}$ are normal of faces of tetrahedron $T_{M+1}$ incident to the endpoint of $v$, $\lambda_{avw}$ is the damping factor to adjust the adaptive visibility weighting γ. Our method is equivalent to Labatut's method (Labatut, Pons, and Keriven 2009) when $\lambda_{avw} = 0$.

**2.6 Experiments**

We evaluate our proposed Vis2Mesh method through a series of quantitative and qualitative experiments covering a variety of small objects, indoor scenarios, large-scale outdoor scenarios either acquired by multi-view stereo, laser scanning, or simulated images. We compare our method with several state-of-the-art learning-based reconstruction approaches including Point2Mesh (P2M) (Hanocka et al. 2020),

Points2Surf (P2S) (Erler et al. 2020), Meshing Point Cloud with IER (MIER) (M. Liu, Zhang, and Su 2020), Convolutional Occupancy Networks(CONet) (Peng et al. 2020), and a classical approach Screened Poisson Surface Reconstruction (SPSR) (Kazhdan and Hoppe 2013). To demonstrate the generalization capacity, we widely collect data from multiple datasets with varying sensors and platforms, including BlendedMVS (Yao et al. 2020), COSEG (Van Kaick et al. 2011), Thingi10K (Q. Zhou and Jacobson 2016), P2M (Hanocka et al. 2020), CONet (Peng et al. 2020), senseFly (senseFly 2020), KITTI (Geiger et al. 2013), Mai City Dataset (StachnissLab 2021), Airborne LiDAR data of Columbus, Ohio (Office of Information Technology Ohio Department of Administrative Services 2020) and Airborne LiDAR data of Toronto (Rottensteiner et al. 2012).

Our method does not assume surface normal information, while it is required for some of the comparing methods (SPSR, P2M, etc.). High-quality surface normals are an important source of information to guide the reconstruction method but are not easily obtainable or directly measured by most of the sensors. In the experiment, the surface normal used by these comparing methods is estimated by a Minimal Spanning Tree (MST) based local plane fitting method, implemented by the open-source software CloudCompare (CloudCompare 2020). The software also integrated SPSR (Kazhdan and Hoppe 2013) implementation which we use in our comparative study. We report the performance of the comparing methods quantitatively in Section 2.6.3, and qualitatively in Section 2.6.5.

### 2.6.1 Training Dataset

Our training dataset is built based on textured mesh models of public Multi-View Stereo (MVS) reconstruction dataset BlendedMVS (Yao et al. 2020). We select 5 textured models as ground truth surfaces for training and 2 models for testing. For each model, we uniformly sample points from the surface. The number of sampled points depends on the actual size and complexity of the structure, ranging from 500K to 10M points. We prefer sparse sampling since the more occluded points have been projected to the virtual views, the more contrastive samples we will have. We collect datasets by sampling virtual views (Section 2.5.1) and rendering them with our renderer. Our renderer is implemented with OpenGL, which not only provides regular color images and depth images but also records the original index of each projection point, as shown in Figure 2.6.

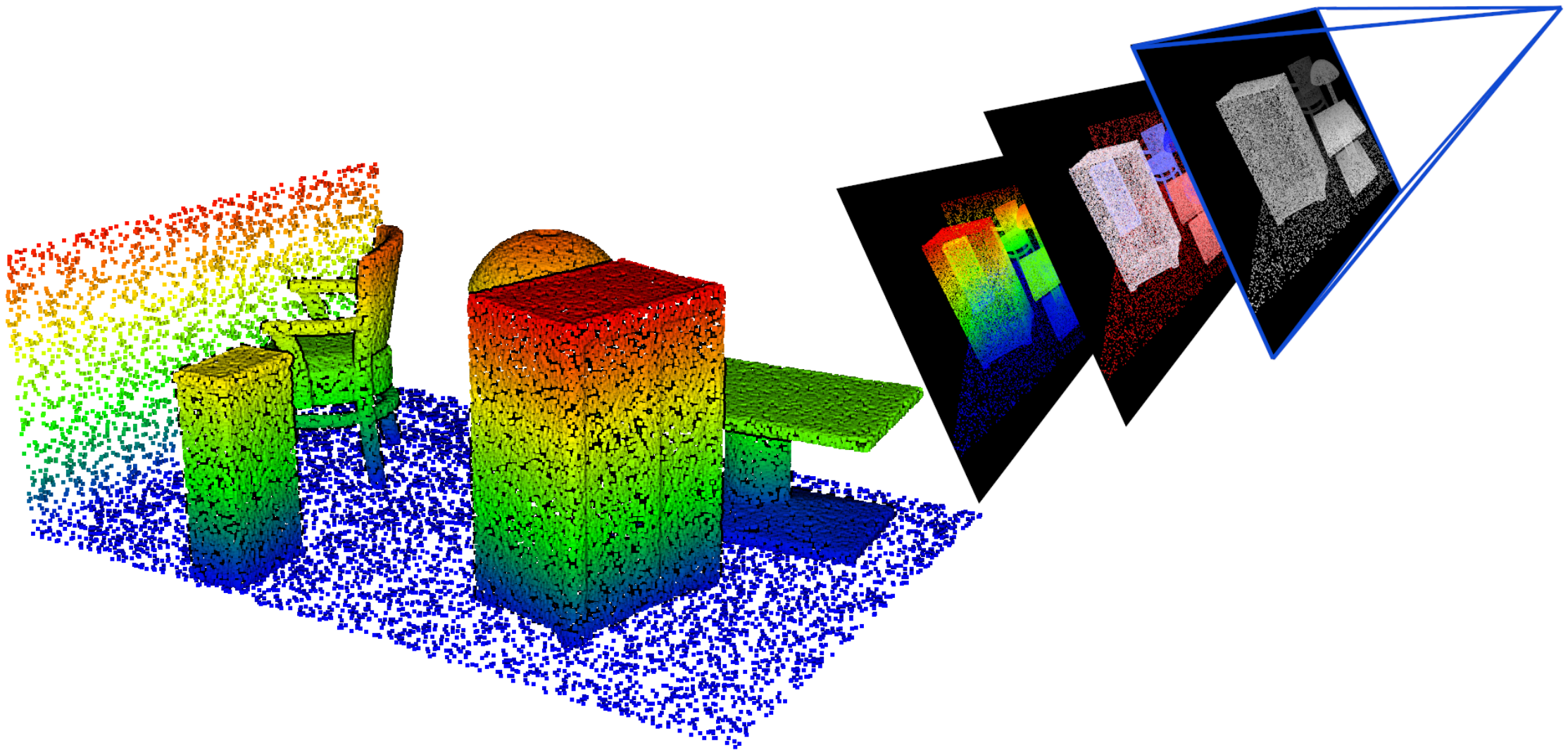

Figure 2.6 Record visibility information in an image by rendering: While points were rendered, the global index of the points been recorded simultaneously.

Figure 2.7 shows an example out of 1414 virtual views. The ground truth visibility is generated by comparing the point-rendering depth map and the surface-rendering depth map. If the difference of depth is less than $\epsilon = 0.05$, it will be marked as visible, otherwise, it will be marked as occluded.

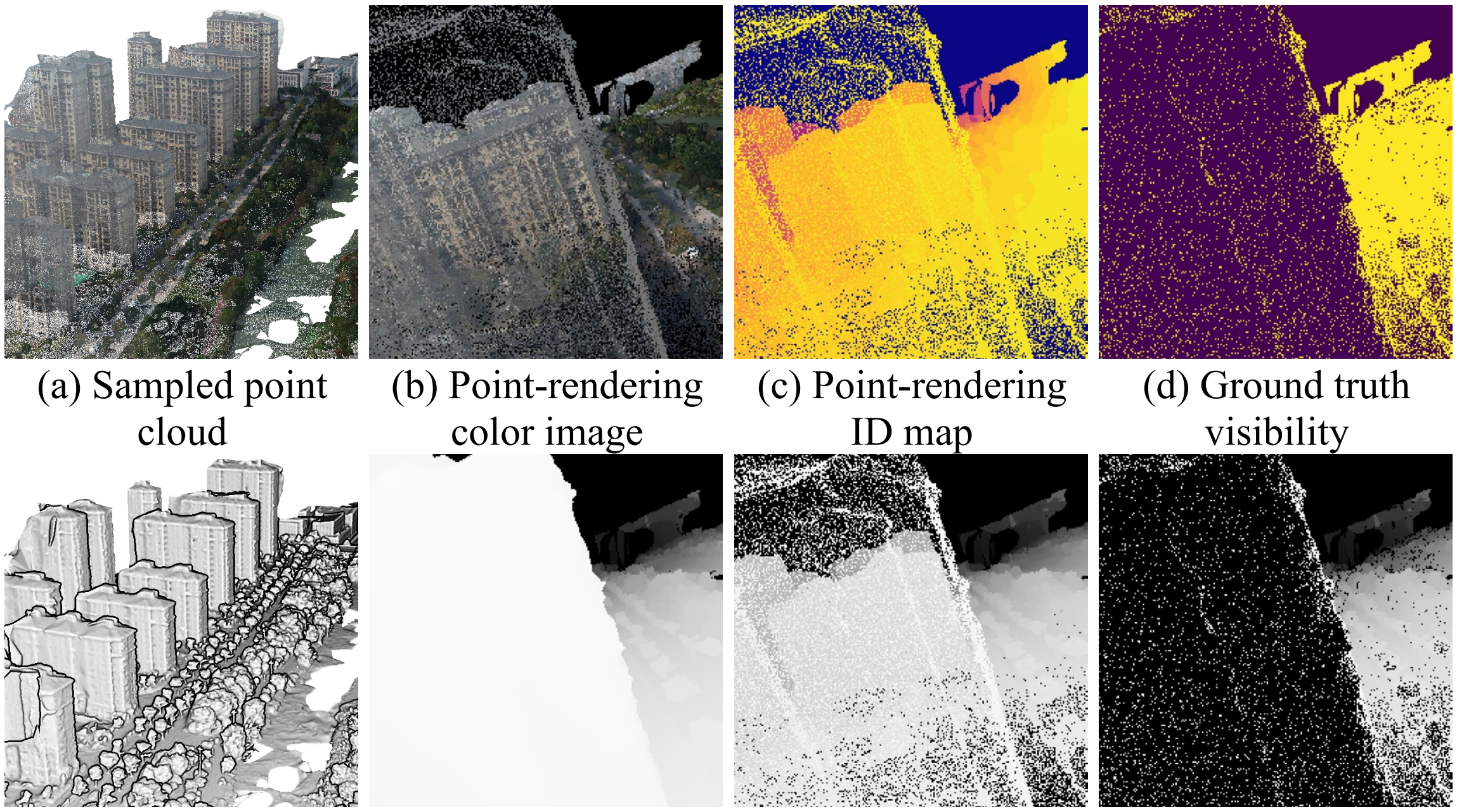

(a) Sampled point cloud (b) Point-rendering color image (c) Point-rendering ID map (d) Ground truth visibility

(e) Ground truth surface (f) Surface-rendering depth map (g) Point-rendering depth map (h) Ground truth cleaned depth map

Figure 2.7 The content of our training dataset.

### 2.6.2 Network Architecture

Table 2.1 provides a detailed description of the input and output tensor sizes of each module in our workflow. And Table 2.2 presents the detailed architecture of the network, including the size of each buffer.

| Module | IO | Description | Data Dimension |
|---|---|---|---|
| Virtual View Sampler | Input | Given Point Cloud | $n \times 3$ |
| | Input | The flag indiate the pattern of view generator | 1 |
| | Output | 6 DoF poses | $m \times 6 \times 1$ |
| Renderer | Input | Given Point Cloud | $n \times 3$ |
| | Input | 6 DoF poses | $6 \times 1$ |
| | Output | Rendered sparse color image | $H \times W \times 3$ |
| | Output | Rendered sparse depth map | $H \times W \times 1$ |
| | Output | Rendered sparse point ID map | $H \times W \times 1$ |
| Coarse VisNet | Input | Normalized and binary mask attached sparse depth map | $H \times W \times 2$ |
| | Output | Predicted mask of visible pixels | $H \times W \times 1$ |
| Multiplication | Input | Rendered sparse depth map | $H \times W \times 1$ |
| | Input | Predicted mask of visible pixels | $H \times W \times 1$ |
| | Output | Cleaned sparse depth map | $H \times W \times 1$ |
| Depth CompNet | Input | Normalized and binary mask attached cleaned sparse depth map | $H \times W \times 2$ |
| | Output | Completed dense depth map | $H \times W \times 1$ |
| Concatenation | Input | Normalized and binary mask attached sparse depth map | $H \times W \times 2$ |
| | Input | Normalized and binary mask attached cleaned sparse depth map | $H \times W \times 1$ |
| | Output | Concatenated raw depth and completed depth map | $H \times W \times 3$ |
| FineVisNet | Input | Concatenated raw depth and completed depth map | $H \times W \times 3$ |
| | Output | Predicted mask of visible pixels | $H \times W \times 1$ |
| Graph-cut based Surface Reconstruction | Input | Given Point Cloud | $n \times 3$ |
| | Input | All rendered sparse point ID map | $m \times H \times W \times 1$ |
| | Input | Predicted mask of visible pixels | $m \times H \times W \times 1$ |
| | Output | Reconstructed triangle surface | Irregular |

Table 2.1 Modules overview. We detail the input and output of all modules that appeared in our workflow. $n$ is the number of input points, $m$ is the number of generated virtual views. $H$ and $W$ are the customized values which are set to $256 \times 256$ in our experiments.

| Part | Layer | Parameters | Output Dimension |
|---|---|---|---|
| Encoder | Double PConv1BnReLU | 3x3,64,64 | $H \times W \times 64$ |
| | MaxPool + Double PConv2BnReLU | 2,3x3,128,128 | $H/2 \times W/2 \times 128$ |
| | MaxPool + Double PConv3BnReLU | 2,3x3,256,256 | $H/4 \times W/4 \times 256$ |
| | MaxPool + Double PConv4BnReLU | 2,3x3,512,512 | $H/8 \times W/8 \times 512$ |
| | MaxPool + Double PConv5BnReLU | 2,3x3,512,512 | $H/16 \times W/16 \times 512$ |
| Decoder | Bilinear Upsample1 | 2 | $H/8 \times W/8 \times 512$ |
| | Concat1 | cat(PConv4, Upsample1) | $H/8 \times W/8 \times 1024$ |
| | Double PConv6BnReLU | 3x3, 256, 256 | $H/8 \times W/8 \times 256$ |
| | Bilinear Upsample2 | 2 | $H/4 \times W/4 \times 256$ |
| | Concat2 | cat(PConv3, Upsample2) | $H/4 \times W/4 \times 512$ |
| | Double PConv7BnReLU | 3x3, 128, 128 | $H/4 \times W/4 \times 128$ |
| | Bilinear Upsample3 | 2 | $H/2 \times W/2 \times 128$ |
| | Concat3 | cat(PConv2, Upsample3) | $H/2 \times W/2 \times 256$ |
| | Double PConv8BnReLU | 3x3, 64, 64 | $H/2 \times W/2 \times 64$ |
| | Bilinear Upsample4 | 2 | $H \times W \times 64$ |
| | Concat4 | cat(PConv1, Upsample4) | $H \times W \times 128$ |
| | Double PConv9BnReLU | 3x3, 64, 64 | $H \times W \times 64$ |
| | Sigmoid | - | $H \times W \times 1$ |

Table 2.2 Detailed encoder-decoder network architecture used for our **CoarseVisNet**, **DepthCompNet**, and **FineVisNet**. **PConv**: partial convolution layer. **Double PConvBnReLU**: PConv+BatchNorm+ReLU+PConv+BatchNorm+ReLU. Parameter of **Double PConvBnReLU**: kernel size, number of filters for the first PConv, and the second PConv.

### 2.6.3 Quantitative Evaluation

We evaluate our method on points sampled from ground truth mesh from COSEG, CONet, and BlendedMVS datasets. Since the existing methods have certain limitations on the amount of data, we divide the data set into object/indoor/outdoor scenes, and the complexity and scene scale are gradually increasing. We uniformly sample 25K points from the ground truth surface for objects, 100K points for indoor data, and 500K points for outdoor data. In order to evaluated with actual data, we generated ground truth for the sensefly (senseFly 2020) dataset by using a commercial MVS pipeline (Pix4D 2023) that employs SPSR for surface reconstruction. Table 2.3 shows F-score (Knapitsch et al. 2017) and Chamfer distance (CD) (Barrow et al. 1977) metrics on each category. To be noted that, we use the sliding window manner of CONet for all scales, and we trim SPSR output by the density value, and the threshold is set to $max_density/2$.

Compared with several state-of-the-art learning-based reconstruction approaches and the SPSR method, our virtual view visibility-based methods present outstanding performance in terms of both metrics on each level of datasets, which also proves the insensitivity of our method to the type of scene. Furthermore, although our VDVNet is trained from the synthetic dataset collected from outdoor-level models, and the quantitative evaluation shows our method has achieved competitive results on object-level and indoor-level datasets. Regarding efficiency, our method has similar speed and memory as SPSR when reaching the same level of detail.

| | | F-score | | | | | | |
|---|---|---|---|---|---|---|---|---|
| | | P2M | P2S | MIER | CONet | SPSR (Trimmed) | Ours ($\gamma_{av} = 0$) | Ours ($\gamma_{av} = 1$) |
| Object 25K | DSLR | 0.9790 | 0.9014 | **0.9935** | 0.9041 | 0.9517 | 0.9844 | 0.9841 |
| | Bull | 0.9990 | 0.9946 | 0.9851 | 0.8203 | **0.9993** | 0.9929 | 0.9991 |
| | Giraffe | 0.9969 | 0.9525 | 0.9938 | 0.9776 | **0.9999** | 0.9989 | 0.9996 |
| | *Average* | *0.9916* | *0.9495* | *0.9908* | *0.9007* | *0.9836* | *0.9921* | ***0.9943*** |
| Indoor 100K | room0 | / | 0.7334 | 0.8885 | 0.9112 | 0.9049 | **0.9345** | 0.9340 |
| | room1 | / | 0.5379 | 0.8453 | 0.8695 | 0.8065 | 0.8747 | **0.8775** |
| | *Average* | / | *0.6356* | *0.8669* | *0.8903* | *0.8557* | *0.9046* | ***0.9058*** |
| Outdoor 500K | Church | / | / | / | 0.4086 | 0.8292 | 0.9206 | **0.9221** |
| | Archway | / | / | / | 0.9133 | 0.9815 | **0.9840** | 0.9772 |
| | Pedestrian street | / | / | / | 0.7494 | 0.9401 | **0.9889** | 0.9860 |
| | Eco Park | / | / | / | 0.7416 | 0.8470 | 0.9431 | **0.9553** |
| | Dragon Park | / | / | / | 0.7700 | 0.8702 | 0.9804 | **0.9866** |
| | *Average* | / | / | / | *0.7166* | *0.8936* | *0.9634* | ***0.9654*** |
| MVS | Hotel | / | / | / | 0.3577 | 0.7732 | 0.8229 | **0.8350** |
| | GSM Tower | / | / | / | 0.1863 | 0.6261 | 0.6321 | **0.6354** |
| | UThammasat | / | / | / | 0.4656 | **0.8046** | 0.7094 | 0.7259 |
| | *Average* | / | / | / | *0.3365* | ***0.7346*** | *0.7215* | *0.7321* |
| | | **Chamfer distance** | | | | | | |
| | | P2M | P2S | MIER | CONet | SPSR (Trimmed) | Ours ($\gamma_{av} = 0$) | Ours ($\gamma_{av} = 1$) |
| Object 25K | DSLR | 0.1593 | 0.3558 | 0.1440 | 0.2570 | 0.1288 | **0.0702** | 0.0707 |
| | Bull | 0.3328 | 0.3576 | 0.3421 | 1.3881 | 0.2914 | 0.3037 | **0.2893** |
| | Giraffe | 0.2227 | 0.4325 | 0.3276 | 0.3831 | **0.1893** | 0.2136 | 0.1939 |
| | *Average* | *0.2383* | *0.3819* | *0.2712* | *0.6761* | *0.2032* | *0.1959* | ***0.1846*** |
| Indoor 100K | room0 | / | 2.0151 | 0.3403 | 0.6476 | 0.8272 | **0.3296** | 0.3571 |
| | room1 | / | 2.8026 | 0.6714 | 0.7814 | 1.2546 | 0.5971 | **0.5457** |
| | *Average* | / | *2.4088* | *0.5059* | *0.7145* | *1.0409* | *0.4634* | ***0.4514*** |
| Outdoor 500K | Church | / | / | / | 19.825 | 1.7784 | 0.6811 | **0.6657** |
| | Archway | / | / | / | 0.2753 | 0.0869 | **0.0677** | 0.0756 |
| | Pedestrain street | / | / | / | 0.8008 | 0.3561 | **0.1590** | 0.1620 |
| | Eco Park | / | / | / | 0.8813 | 0.9463 | 0.3273 | **0.2961** |
| | Dragon Park | / | / | / | 0.8696 | 0.7398 | 0.2096 | **0.1859** |
| | *Average* | / | / | / | *4.4809* | *0.7815* | *0.2890* | ***0.2770*** |
| MVS | Hotel | / | / | / | 1.3053 | 0.9369 | 0.3660 | **0.3479** |
| | GSM Tower | / | / | / | 6.6513 | 0.8051 | 0.6935 | **0.6888** |
| | UThammasat | / | / | / | 1.4951 | 0.3741 | 0.3982 | **0.3737** |
| | *Average* | / | / | / | *1.7845* | *0.7054* | *0.4859* | ***0.4701*** |

Table 2.3 Quantitative comparison of reconstruction on COSEG, CONet, and BlendedMVS dataset, grouped by the size of the scene and number of points.(``F-score'': Higher is better. ``Chamfer distance'': Lower is better. ``/'': Not applicable.)

### 2.6.4 Ablation Studies

We first demonstrate the advantage of virtual view visibility in surface reconstruction even if the actual visibility is available (Figure 2.9) and show the quality improvement by the proposed adaptive visibility weighting (Figure 2.10). Then, we analyze the network architecture with a quantitative evaluation (Table 2.4).

**Importance of virtual view visibility.** In this experiment, we compare surfaces reconstructed by real visibility (obtained from physical views) and virtual view visibility to prove the importance of virtual visibility. The data in this experiment is LiDAR scans associated with 6DoF pose trajectory from the Mai City Dataset (Geiger et al. 2013). We present three methods as shown in Figure 2.8. *Per Point* is generated by associating the corresponding scan pose with each point, and each point has only one line of sight (or visibility) collected from the sensor. For *Grid Fusion*, we use a voxel grid filter to aggregate nearby points' visibility to obtain points with multiple actual visibility. Finally, we use the proposed network to generate virtual view visibility, where the virtual views are sampled along the recorded trajectory.

The visual results shown in Figure 2.9 demonstrates that the virtual views visibility place a significant improvement in the mesh quality. Per Point actual visibility failed to reconstruct a complete scene because only one line of sight associated with each point is not geometrical stable, and is not sufficient to create a reliable weighted graph for the optimization. Grid Fusion shows better reconstruction quality since more lines of sight are included (4/points on average), but the details of reconstruction are poor, and noises seriously affect the surface. In contrast, our pure virtual view visibility method creates smooth, detailed, and noise-free surfaces (thanks to the denoising capability of VDVNet).

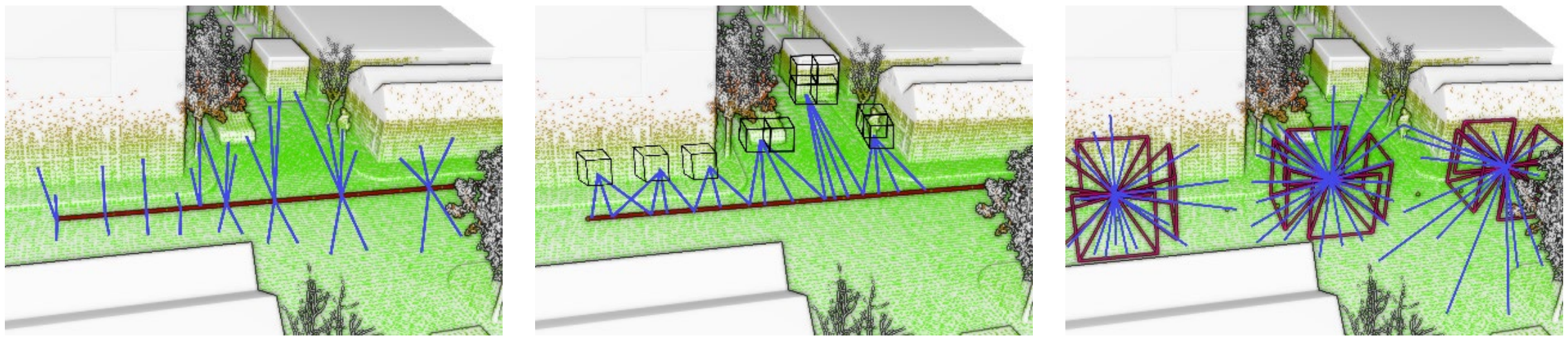

(a) Per Point (b) Grid Fusion (c) Ours

Figure 2.8 Methods of modeling visibility from mobile LiDAR scans. (Per Point: each LiDAR point only connects to the trajectory once. Grid Fusion: points within the voxel of the grid fuse into a centroid and connect it to the trajectory. Ours: pure virtual visibility generated from sampled views and VDVNet.)

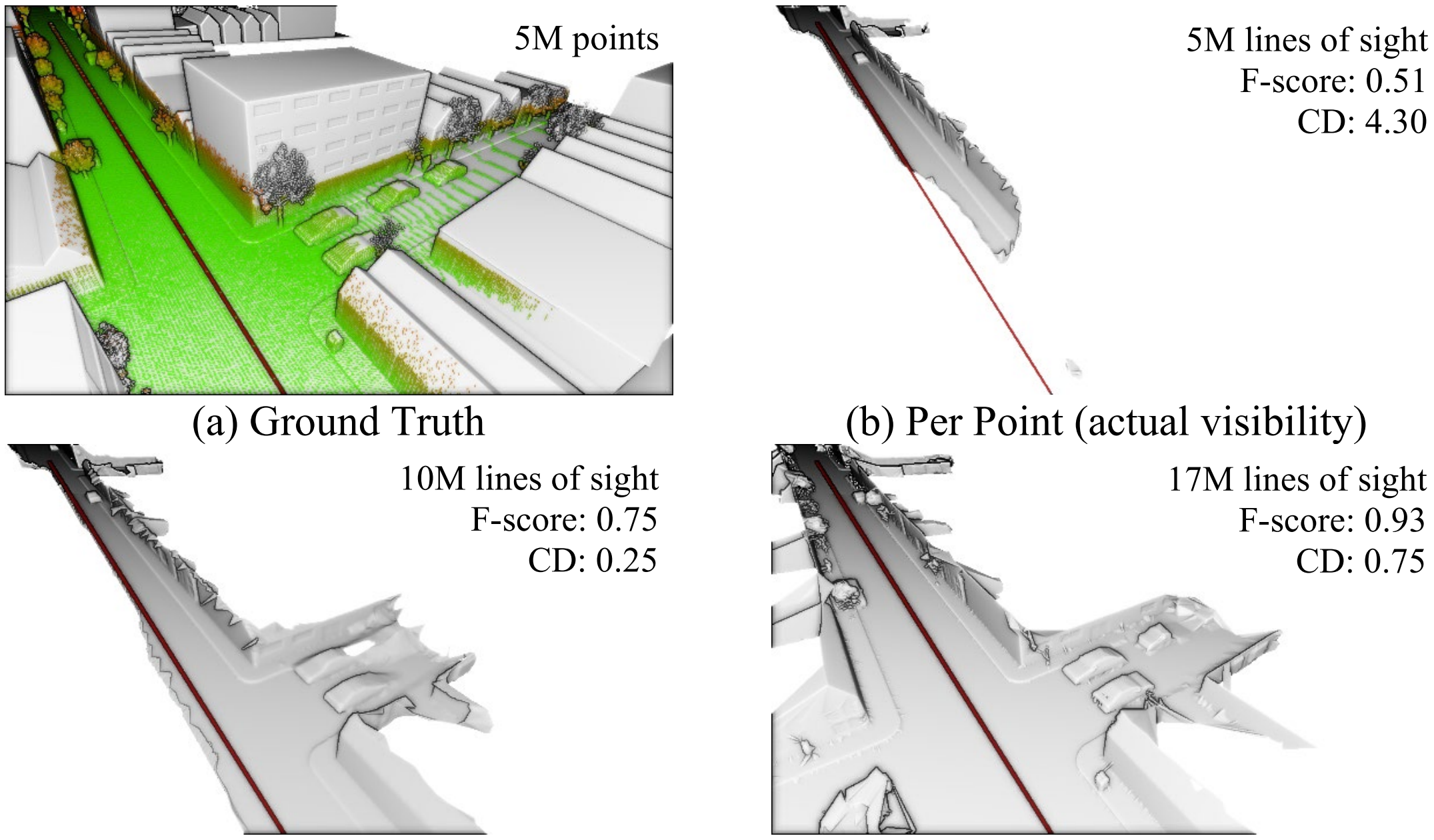

(a) Ground Truth (b) Per Point (actual visibility)

(c) Grid Fusion (actual visibility) (d) Ours (virtual view visibility)

Figure 2.9 Importance of virtual view visibility. Reconstruction using our method for each visibility dataset in Figure 2.8.

**Importance of adaptive visibility weighting.** Figure 2.10 demonstrates the impact of proposed adaptive visibility weighting. To be noted that when setting $\lambda_{avw} = 0$, the visibility weighted graph is exactly equivalent to the basic method (Labatut, Pons, and Keriven 2009). We observe the method $\lambda_{avw} = 1$ works better than the basic method for objects with pole-like and notches on the surface structure, since such a complex structure may introduce large incidence angles for our virtual lines of sight. The adaptive visibility weighting works as we expected in Section 2.5.3. The quantitative results in Table 2.3 show that in most cases, adaptive visibility weighting works better than the basic method.

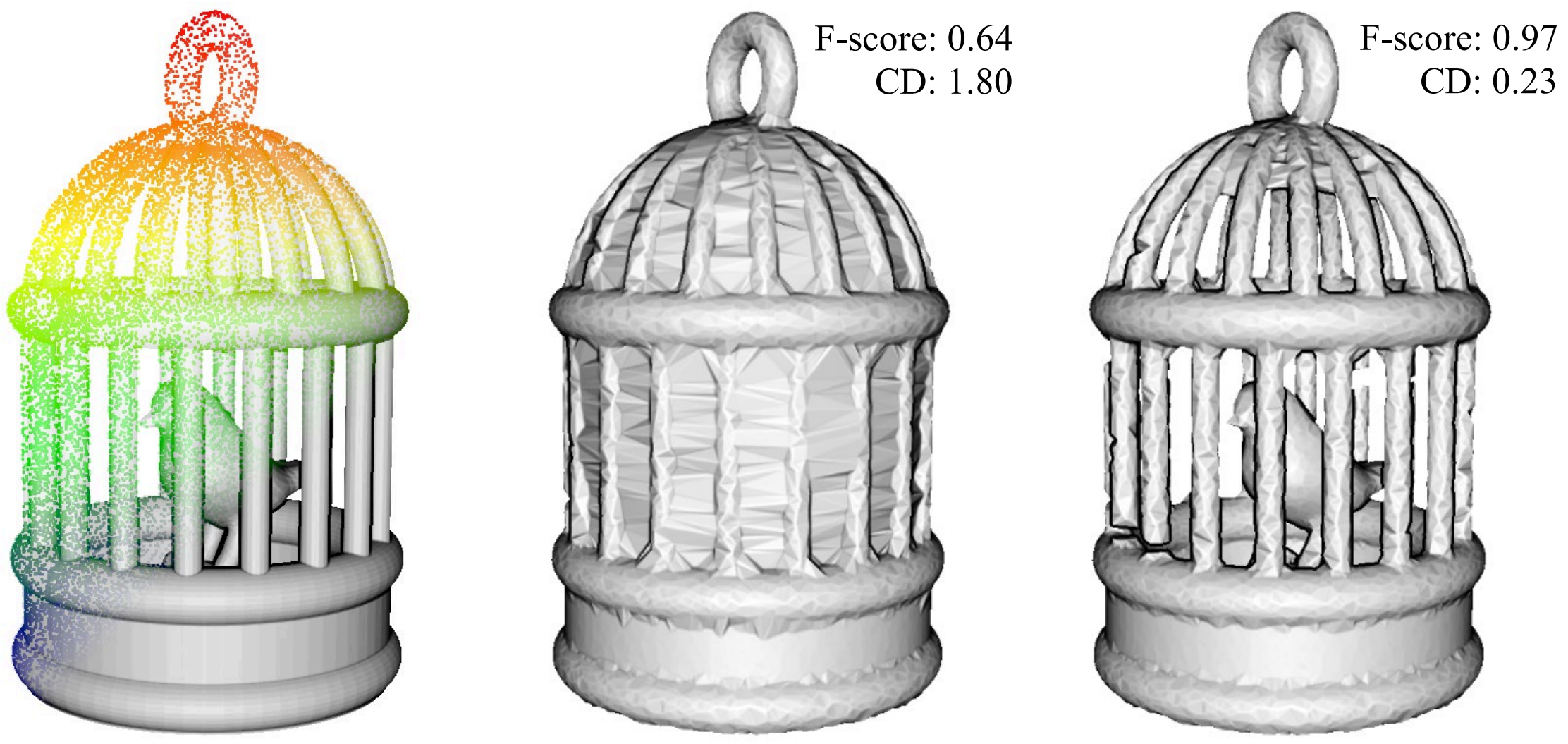

(a) Points (b) $\lambda_{avw} = 0$ (c) $\lambda_{avw} = 1$

Figure 2.10 Impact of AVW on synthetic data with complex shape. The data is *Birdcage*.

**Impact of the number of virtual views.** To evaluate the effect of the view configuration, we demonstrate the reconstruction results w.r.t. different numbers of virtual views in Figure 2.11. Firstly, the single view on the top center of the scene generates a very coarse surface. Secondly, 7 virtual views on the spherical orbit around the target carved out objects of the scene, the base of the lamp, legs of the chair starting emerging. With more virtual views, the reconstructed surface yields more geometric details. The last configuration contains virtual views selected interactively by the user to reconstruct the complex structure of the chair.

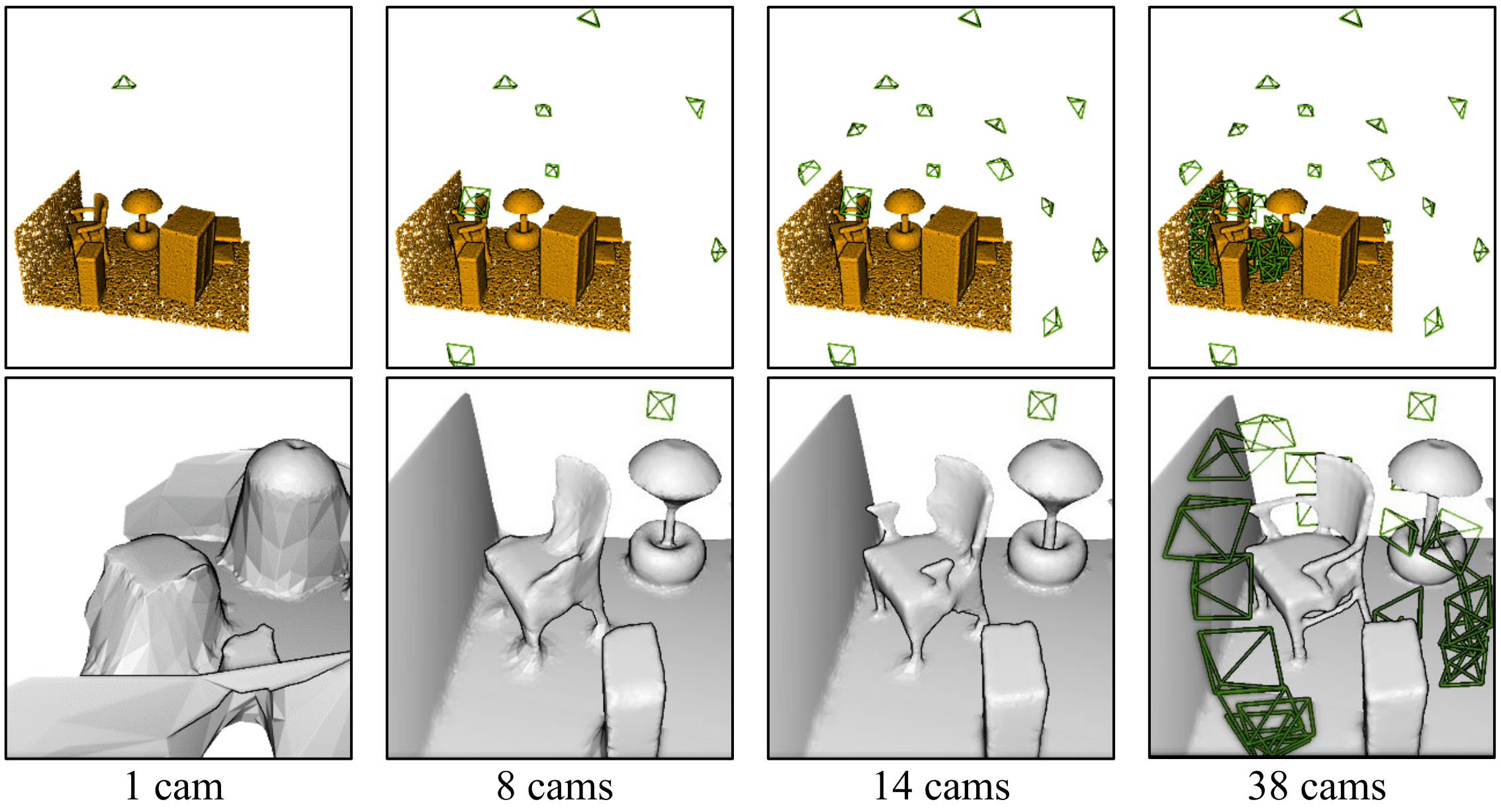

Figure 2.11 Impact of the number of virtual views.

**Visibility estimation networks.** We independently evaluate the performance of our VDVNet in the task of visibility prediction comparing with geometric-based method Hidden Point Removal (HPR) (Katz, Tal, and Basri 2007) implemented by Open3D (Q.-Y. Zhou, Park, and Koltun 2018) and the baseline learning method UNet (Ronneberger, P.Fischer, and Brox 2015). The VisibNet proposed by InvSFM (Pittaluga et al. 2019) is designed for the same task as us, and its architecture is UNet with convolutional layers at the end of the decoder.

We report the size and the performance of classifiers in Table 2.4 on our pixel-wise visibility dataset sampled from BlendedMVS. We observe the networks with partial convolutional layers outperform standard convolutional networks without parameter increasing. The variants of our networks present the contribution of the depth completion module and partial convolutional layers w.r.t. visibility estimation task.

| **Visibility Estimator** | #Param | %P | %R | %F1 | %AUC |
|---|---|---|---|---|---|
| HPR | / | 85.79 | 85.01 | 85.24 | 82.22 |
| UNet | 51M | 90.42 | 87.10 | 88.68 | 88.52 |
| UNet + PConv | 51M | 90.73 | 88.03 | 89.32 | 89.59 |
| VisibNet | 52M | 90.15 | 91.50 | 90.78 | 90.22 |
| Ours w/o DepthComp or PConv | 153M | 91.84 | 84.08 | 87.71 | 89.29 |
| Ours w/o DepthComp | 153M | 91.38 | 93.10 | 92.21 | 92.62 |
| Ours w/o PConv | 153M | 91.86 | 94.24 | 93.01 | 93.14 |
| Ours | 153M | 92.37 | 94.95 | 93.63 | 94.17 |

Table 2.4 Quantitative analysis of methods on binary visibility classification task. #Param is the number of parameters, %P is precision, %R is recall, %F1 is F1 score, and %AUC is Area Under the Curve of ROC curves. DepthComp represents the depth completion module and PConv denotes partial convolutional layers.

**Impact of noises and missing data.** Since the real 3D point clouds normally have noises and incomplete data, it is very important for mesh generation to deal with these problems. We designed two experiments to evaluate the robustness of different methods against the additional noises and incompleteness.

We evaluated the impact of additional noises (random noise and outliers) on CONet, SPSR, and our method on indoor scenes, as shown in Figure 2.12. The added noises may affect both the position of the points and the normals of the points. The middle column of Figure 2.12 shows outliers introduced undesired objects to CONet, but it has much less impact on other methods and ours. However, noises can be problematic for SPSR since it relies on high-quality normal estimation, which can be easily affected by noises. CONet and our method have demonstrated a certain level of noises robustness in the reconstructed meshes. Incomplete data or missing data is a common problem in practice, such as those from laser scanning. Figure 2.13 compares the results of CONet and ours in large-scale scenes, and we note that our approach is able to complete the missing facades with triangle faces in the incomplete region. Thanks to Delaunay triangulation and tetrahedron labeling, our method is able to close gaps and holes caused by missing data.

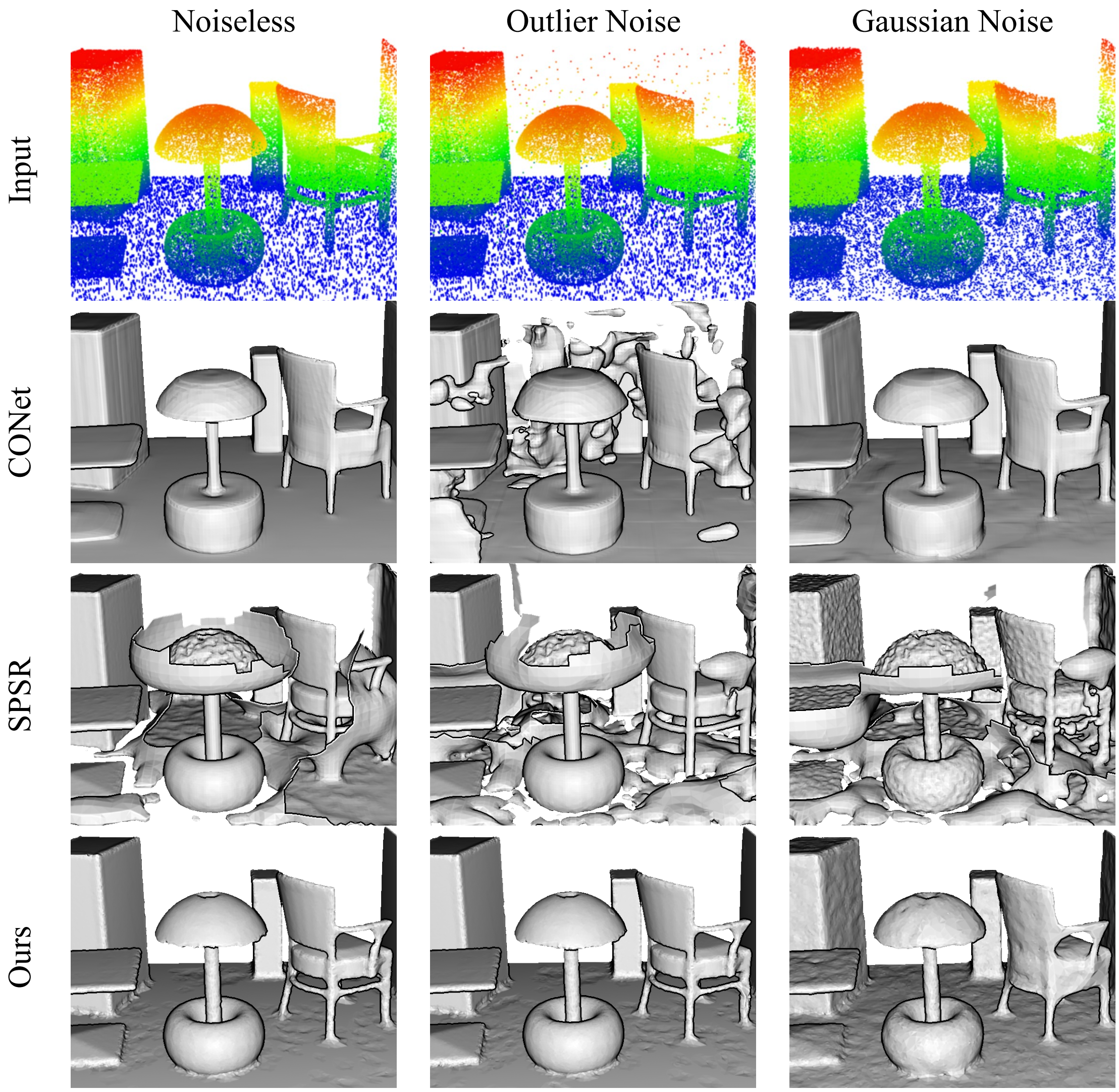

Figure 2.12 Evaluation of the methods subject to additional noises on the point clouds. The data is *room0*.

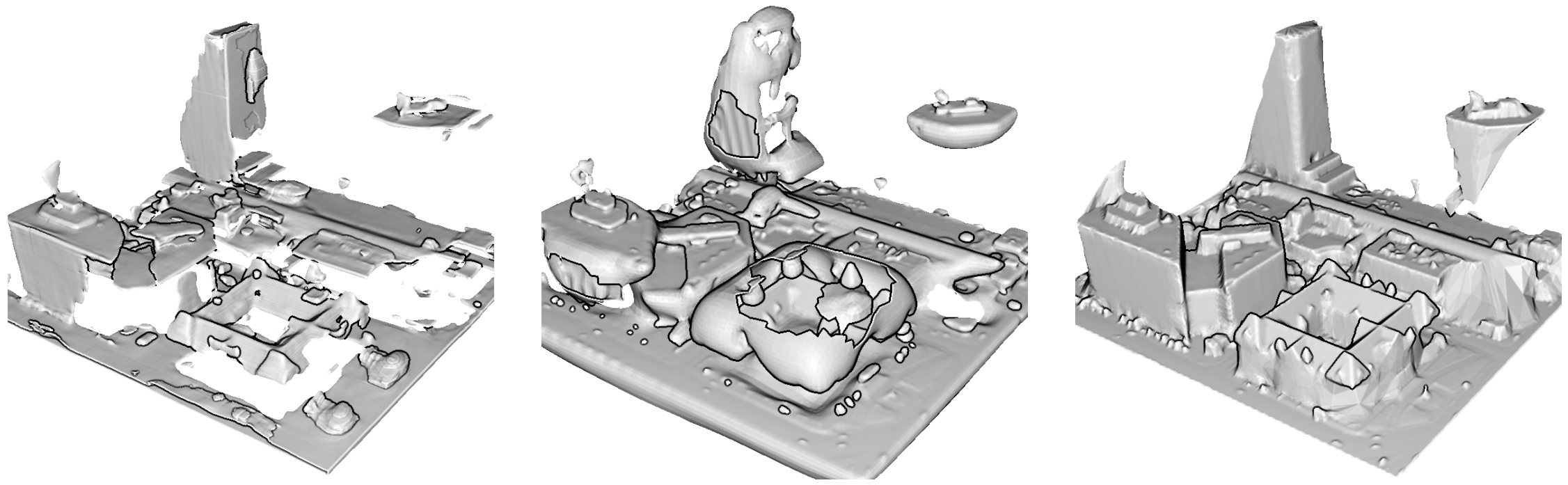

(a) CONet (b) SPSR (c) Ours

Figure 2.13 Evaluation of the methods on incomplete point clouds. The data is *Toronto downtown*.

### 2.6.5 Qualitative Evaluation

Our qualitative comparison is divided into three scales as well, shown in Figure 2.14 , the first column of Figure 2.12, and Figure 2.15 respectively. P2M (Hanocka et al. 2020) and P2S (Erler et al. 2020) have poor reconstruction quality in terms of open surface (DSLR). Triangles generated by MIER (M. Liu, Zhang, and Su 2020) are close to the ground truth surface, but the completeness is poor. CONet (Peng et al. 2020) usually produces a flat surface, which is a good feature for specific scenes (i.e., indoors), but it is not good for curved surfaces. While SPSR (Kazhdan and Hoppe 2013) also produces reasonable reconstructions, it tends to close the resulting meshes and present bubbles event after trimming with the threshold $max_density/2$. A carefully chosen trimming parameter could reduce such artifacts. Moreover, instead of uniform sampling from the ground truth surface, we use real-world datasets derived from multi-view stereo (senseFly 2020) or laser scan data (Geiger et al. 2013) to demonstrate the generality of the proposed method.

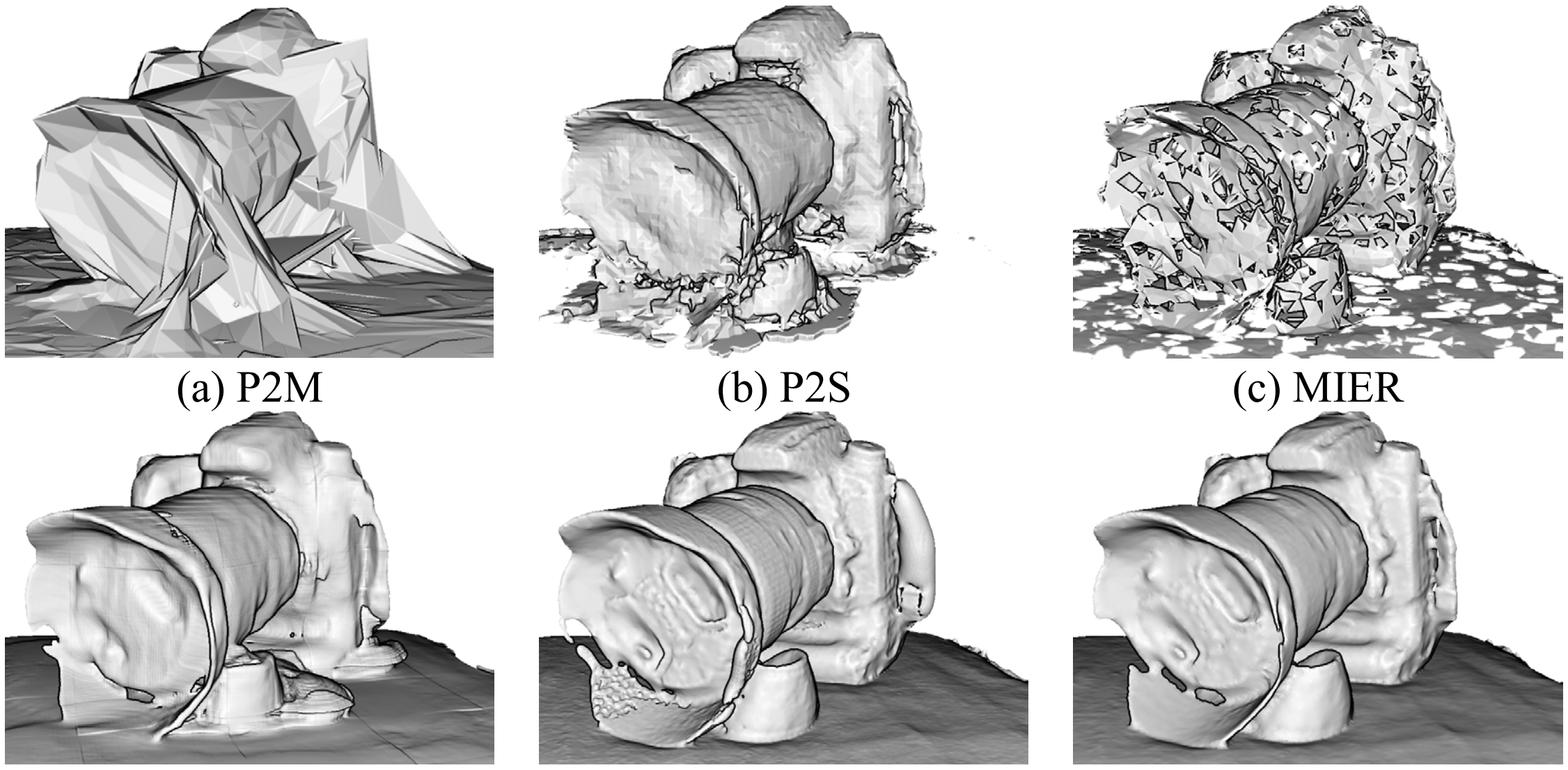

(a) P2M (b) P2S (c) MIER

(d) CONet (e) SPSR (f) Ours

Figure 2.14 Qualitative comparison on small object level datasets. The data is *DSLR*.

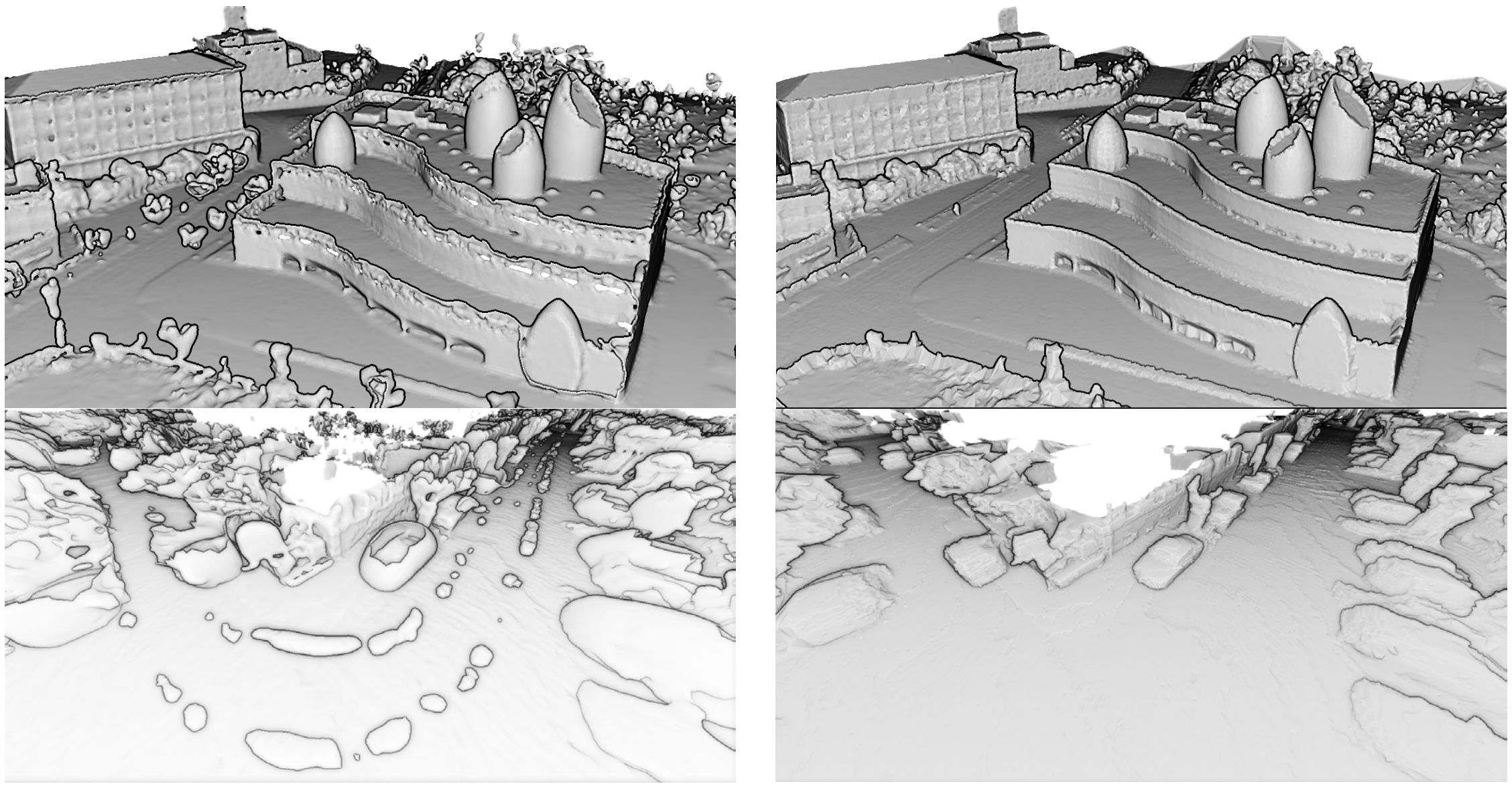

(a) SPSR (b) Ours ($\lambda_{avw} = 1$)

Figure 2.15 Qualitative comparison on the large-scale outdoor dataset. The data are *Eco Park* (top) and *Crossroad* (bottom).

## 2.7 Conclusion

We presented a novel surface reconstruction framework that combines the traditional graph-cut based method and learning-based method. In our method, the neural network only focuses on a very simple task - learning to predict the visibility of the point clouds given a view which thus present a much better generalization capability over existing learning-based approaches, while maintaining the efficiency of the traditional method. Specifically, we proposed a three-step network that explicitly employs depth completion for the visibility prediction of 3D points. Furthermore, we have improved graph-cut based formulation by proposing a novel adaptive visibility weighting term. Experiments show that our proposed method can be generalized to point clouds in different scene contexts, and can be extended to point clouds in large scenes. On datasets of varying scales, our framework presents better performance than the comparing classic reconstruction methods and state-of-the-art learning-based approaches.

# Chapter 3. Mesh Conflation of Oblique Photogrammetric Models using Virtual Cameras and Truncated Signed Distance Field

This chapter is based on the paper called "Mesh Conflation of Oblique Photogrammetric Models Using Virtual Cameras and Truncated Signed Distance Field" that was published in the "IEEE Geoscience and Remote Sensing Letters (Volume: 20)" by Shuang Song and Rongjun Qin, 2023.

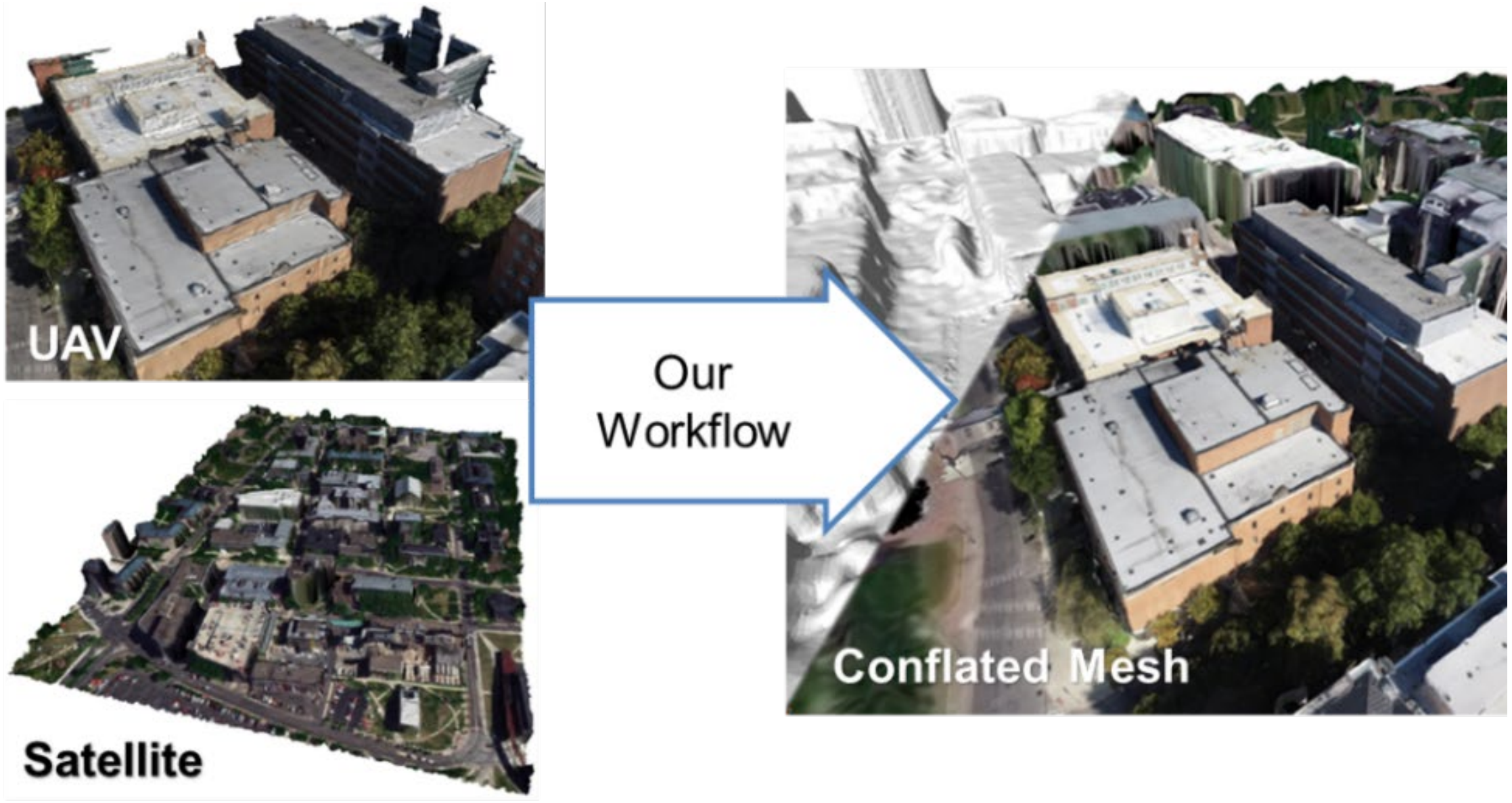

Figure 3.1 Conflation result of textured meshes with different resolutions.

### 3.1 Chapter Abstract

Conflating/stitching 2.5D raster digital surface models (DSM) into a large one has been a running practice in geoscience applications, however, conflating full-3D mesh models, such as those from oblique photogrammetry, is extremely challenging. In this letter, we propose a novel approach to address this challenge by conflating multiple full-3D oblique photogrammetric models into a single, and seamless mesh for high-resolution site modeling. Given two or more individually collected and created photogrammetric meshes, we first propose to create a virtual camera field (with a panoramic field of view) to incubate virtual spaces represented by Truncated Signed Distance Field (TSDF), an implicit volumetric field friendly for linear 3D fusion; then we adaptively leverage the truncated bound of meshes in TSDF to conflate them into a single and accurate full 3D site model. With drone-based 3D meshes, we show that our approach significantly improves upon traditional methods for model conflations, to drive new potentials to create excessively large and accurate full 3D mesh models in support of geoscience and environmental applications.

### 3.2 Introduction

Typical geoscience applications rely on large-scale and wide-area 2.5D digital surface models (DSM) to characterize the earth surfaces. As a common practice, a large DSM is produced through stitching or conflating multiple individual and smaller DSMs, which could be generated from means such as photogrammetry, LiDAR, and SAR (Synthetic Aperture Radar) interferometry. In recent years, Oblique Photogrammetry is attracting increasing attention, due to its capability to characterize sites in full 3D, which greatly facilitates applications such as digital twin, disaster responses to landslides, and earthquakes that large-scale site models with detailed façade geometry. However, unlike model conflation for 2.5D raster DSM, conflating full 3D site models is extremely challenging, as these models are often represented in full-3D triangle meshes, whose topology among vertices is intricate, and oftentimes associated with complex manifold geometry. This is made further even more challenging when these individual mesh models are from different sources, with different resolutions, accuracy, and level of uncertainty. Therefore, when presenting multiple mesh models, the current practices simply overlay these models for visualization, and at most, moderate the mesh conflation process through simple averaging (Katzil and Doytsher 2006; Kyriakidis, Shortridge, and Goodchild 1999; Ruiz et al. 2011), followed by manual or semi-automatic mesh editing through Boolean operations, thus leaving mesh conflation a largely underserved task. Recent advances in 3D scene representations, such as through implicit surfaces, occupancy grid, signed distance field (SDF), and truncated signed distance field (TSDF) (Curless and Levoy 1996), have made it possible to manipulate 3D assets under continuous fields. Among these, TSDF has been a well-supported 3D implicit representation model in the Computer Graphics (CG), due to its simplicity in representing compact spaces, its scalability for large scene processing due to the locality of the updating process and marching cubes (Newman and Yi 2006), and its capability to set forth 3D geometry fusion/conflation frameworks.

Therefore, in this chapter, we present a novel approach that uses a virtual camera (Shuang Song, Cui, and Qin 2021) induced TSDF to fuse multiple mesh models into a single and coherent mesh. Specifically, our approach fills a few gaps of prior works in view-based large-scale mesh fusion/conflation: 1) we propose a novel panoramic virtual camera field to minimize the chances of occlusion at complex objects. 2) we introduce a scheme concerning non-uniform weights with respect to the quality of individual meshes, thus it provides a more reliable and accurate means for mesh model conflation. 3) we adapt the ground-plane assumption oblique photogrammetric models to automate the virtual camera placement.

### 3.3 Related Works

*Mesh conflation* aims to combine mesh models from different sources or scans into a single and coherent mesh model (Ruiz et al. 2011). As mentioned earlier, this is a rather underserved task, yet is still highly relevant to a number of existing approaches (Abdollahi and Riyahi Bakhtiari 2017; Butenuth et al. 2007; Samsonov 2020; Sledge et al. 2011; W. Song et al. 2013; M. Zhang, Yao, and Meng 2016). Remeshing is a family of methods in computer graphics, which has been used to create a new mesh with a desired topology conditioned on the original surface geometry. It is categorized into two types: explicit remeshing and implicit remeshing. Explicit remeshing primarily preserves the 3D vertices and generates triangular networks subjected to desired constraints, such as number of faces and conformation (Gueziec et al. 2001). Algorithms in this category are primarily built based on heuristics and can be hardly extended to meshes representing complex objects (Hilton et al. 1998); The implicit remeshing first converts the original mesh to an intermediate volumetric representation, such as UDF (unsigned distance field), SDF (signed distance field), TSDF (truncated SDF), and then extracting a new mesh from the volumetric representation (Newman and Yi 2006). Most of these volumetric representations favor close surfaces (Zhao 2005), while being unsuitable for photogrammetric mesh models, which are typically open, with unknown manifoldness, and oftentimes with holes and artifacts. TSDF was considered as a useful tool to address such challenges, for example, TSDF can be built based on individual views to minimize the topological complexity in fusion and has been widely used in surface reconstruction applications (Newcombe et al. 2011). Yet this approach requires known image views to guide the fusion and conflation process, which might not be always available.

### 3.4 Method

We present a novel approach that directly conflates full-3D mesh models based on TSDF (Section 3.4.1), through means of virtual cameras. Our approach does not require known image views and can benefit from the efficiency of TSDF presentations.

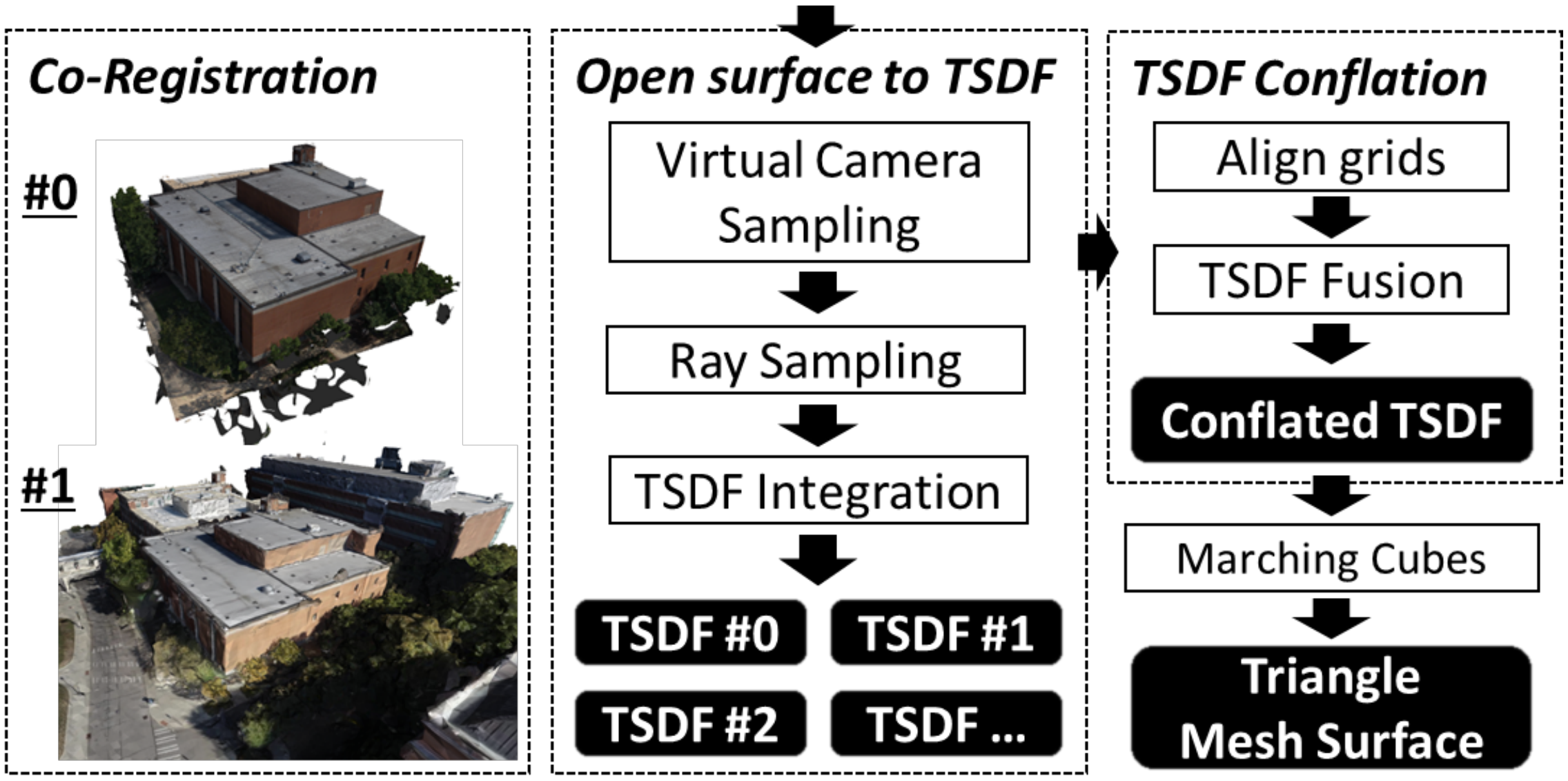

Figure 3.2 The workflow of our proposed mesh conflation approach.

Figure 3.2 presents the workflow of our proposed method. In our approach, we assume well-registered meshes (Xu, Qin, and Song 2023) and start with sampling virtual cameras (Section 3.4.2) in the outer space defined by mesh surfaces. Then we generate rays for each pixel originating from the virtual cameras and find the first intersection with the mesh surface. Since the mesh is a continuous surface, this process can yield pixel-wise depth values as the depth images. Then we integrate these rays into view-dependent TSDF and conflate them in the object space leveraging weights of different meshes based on their respective local geometry. This process is done for each view to update the TSDF field in the object space. Finally, we extract triangle meshes from the implicit surface representation using marching cubes methods (Newman and Yi 2006).

### 3.4.1 Open Surface to TSDF

Signed Distance Function (SDF) defines a function D(x) of the 3D space, where its function value at a 3D location x defines the Euclidean distance of this x location to the actual surface, and the sign indicates if x is inner or outer the surface (Malladi, Sethian, and Vemuri 1995). In a fusion framework, it usually couples with a weight function W(x) defining the importance of a point at fusion. TSDF is a more memory-efficient version of SDF, since it truncates the field only near the surfaces at a bandwidth m (see equation (1-3)). This feature makes it suitable for large-scale applications with limited resources. The TSDF can be easily constructed from closed surfaces using sweep (Zhao 2005) and flood-filling (Rong and Tan 2006) algorithms. Yet, these methods do not apply to open surfaces, as flood-filling yields nothing due to the presence of a single connected components.

The method of (Curless and Levoy 1996) performs TSDF integration given a few depth images. Basically, the method presupposes iterative updating of TSDF distance $D(x): R^3 \rightarrow R$ and weight $W(x): R^3 \rightarrow R$ functions. Both are real-valued functions defined for every voxel of the voxel grid. They are initialized with zeros and updated for each ray (Equation (3.1) - (3.3)) in every depth image.

$$r_i(t) = o_i + t \cdot v_i \,, \tag{3.1}$$

$$d\big(r_i(t)\big) = \max\big(-m, \min(m, t_i - t)\big)\,, \tag{3.2}$$

$$w\big(r_i(t)\big) = \begin{cases} C, & \text{if } |t_i - t| \le m \\ 0, & \text{otherwise} \end{cases}, \tag{3.3}$$

where $o_i \in R^3$ is the position of camera center, $v_i \in R^3$ is the normalized direction of the ray, $t$ is the length of the ray, $d(x)$ and $w(x)$ define TSDF along the ray $r_i$, $t_i$ is the distance from ray to nearest surface, and a distance m defines the vicinity of the surface. In common practice, m is set to 3 times the voxel width. Then, the TSDF can be update using Equation 3 and 4 in (Curless and Levoy 1996).

In our case, the weight function (Curless and Levoy 1996) is not suitable for a wide band width $m$. The problem is resolved by introducing adaptive weights (Section 3.4.3).

### 3.4.2 Virtual Ray Sampling

Converting explicit mesh models to TSDF requires ray sampling through the known geometry to the view space. However, typically, view space information does not exist (e.g., LiDAR data), or is not stored in mesh models. We hence propose to construct a virtual view field, and subsequently, design casted rays between the surface and these views, including their origin, direction, and length.

*Virtual camera placement for determining ray origins*. This virtual camera placement problem (Hardouin et al. 2020; Shuang Song, Cui, and Qin 2021) considers sampling a small number of observing cameras, such that their view space completely cover the scene with optimal angles. To maximize the coverage, we propose to place virtual cameras with a panoramic field of view (FoV), and since it covers all directions of concern. We propose a heuristic virtual camera placement algorithm: it first builds a coarse occupancy grid from the mesh model (green wireframe in Figure 3.3). Then, the algorithm (shown in Algorithm 3.1 samples virtual camera centers based on the occupancy grid. Since outdoor scenes mostly follow a ground-plane assumption (XOY plane sets the ground, and Z relays the height), the camera placement starts with the highest cell for each XOY cell (top height layer of the voxel grid $G_z$). To avoid empty cells (holes), instead of placing the camera for each cell of $G_z$, we apply a window size $\phi$ (set as 3 voxels) and look for the highest cell within the window. For cells at facades (large height difference), we sample along the vertical direction per voxel (For details please see Algorithm 3.1).

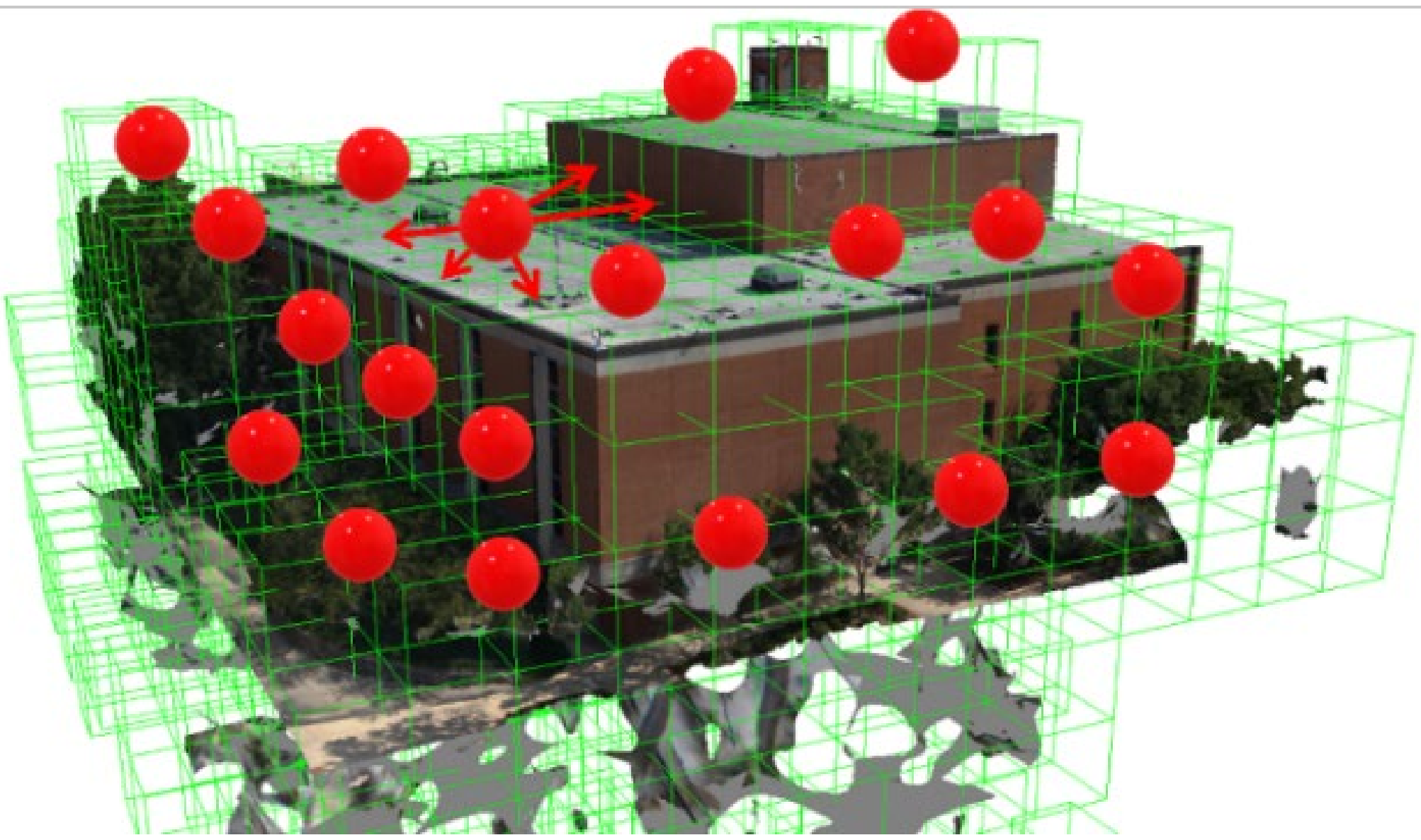

Figure 3.3 Virtual panoramic cameras (red spheres) based on the occupancy of the aerial model.

Algorithm 3.1 Virtual Camera Placement.

```
input Gz: Top height layer of Voxel Grid.
input ϕ: window size
procedure 1: find_high_cell(Gz, i, j, ϕ)
z = -1
for ii=i- ϕ; ii<i+ ϕ; ii++ do
    for jj=j- ϕ; jj=j+ ϕ; jj++ do
        z = max(z, Gz[ii,jj])
return z

procedure 2: sample_camera_centers (Gz, ϕ)
// Gz: p,q matrix. p,q for x,y index
for i=0; i<p; i++ do
    for j=0; j<q; j++ do
        begin_k[i,j] = find_high_cell(Gz, i, j, ϕ)
for i=0; i<p; i++ do
    for j=0; j<q; j++ do
        end_k[i,j] = max(begin_k[i- ϕ:i+ ϕ,j- ϕ:j+ ϕ])+1
        for k=begin_k[i,j]; k<end_k[i,j]; k++ do
            yield i,j,k  // i,j,k refers to camera location
```

With the determined virtual cameras, the *Ray direction* can be sampled from the unit sphere $\mathcal{S}^2$ (i.e., panoramic FoV) of each virtual camera. This is typically done by sampling rays using regular latitude and longitude grids, while such a method can lead to a higher concentration of samples near the poles. Instead, we use the Fibonacci lattice method (González 2010), which was proven to be more uniformly distributed on the $\mathcal{S}^2$.

*Ray length*, defined as t in Equation (3.1), describes the traveling distance from the origin to the object's surface. Given the sampled virtual camera position $o_i$ and direction $v_i$, t is determined by the intersection test between the ray and the geometry. This is supported by modern graphics-card based raytracing, in which billions of rays can be tested in a second.

### 3.4.3 TSDF-based Conflation

To adapt the model conflation application into the TSDF integration framework (Curless and Levoy 1996), we proposed additional modifications to accommodate data with different quality, by using adaptive weights associated with the bandwidth of the TSDF (Equation (3.2) - (3.3)): by definition, when converting the mesh model into TSDF, only the signed distance (Equation (3.2)(2.2)) within the truncated band has non-zero values, so as the weights (Equation (3.3)). Zero weights outside the truncated band may cause ambiguities. Therefore, the bandwidths (as part of the weight definition) of TSDF from different sources of mesh can be adjusted in referencing to the quality of different meshes, which can be determined based on prior knowledge about the dataset itself (e.g., resolution and accuracy). A weighted average over these TSDFs can be applied to achieve the final conflation.

In addition, since the TSDF is discontinuous at truncated boundary, in certain cases, it generates a branched line as illustrated in Figure 3.4d on occasions when non-convex shapes are formed when two triangles intersect. When in 3D it will yield artifacts causing missing details.

We proposed a novel approach to further improve the adaptiveness of our method on the truncated band: First, we define the maximal bandwidth based on the widest uncertainty bounds of the input datasets. Then, instead of using this fixed bandwidth value, we gradually reduce the bandwidth from one side (in our case we take the negative side) to ensure the bandwidth of one triangle does not interfere with another. This process is shown in Figure 3.5: based on the surface point $P$, we extend its ray at $\lambda$, to which the negative bandwidth is defined. The final conflated surface can then be extracted with well-known marching cubes algorithm (Newman and Yi 2006). The resulting conflation output is presented in Section 3.5.2.

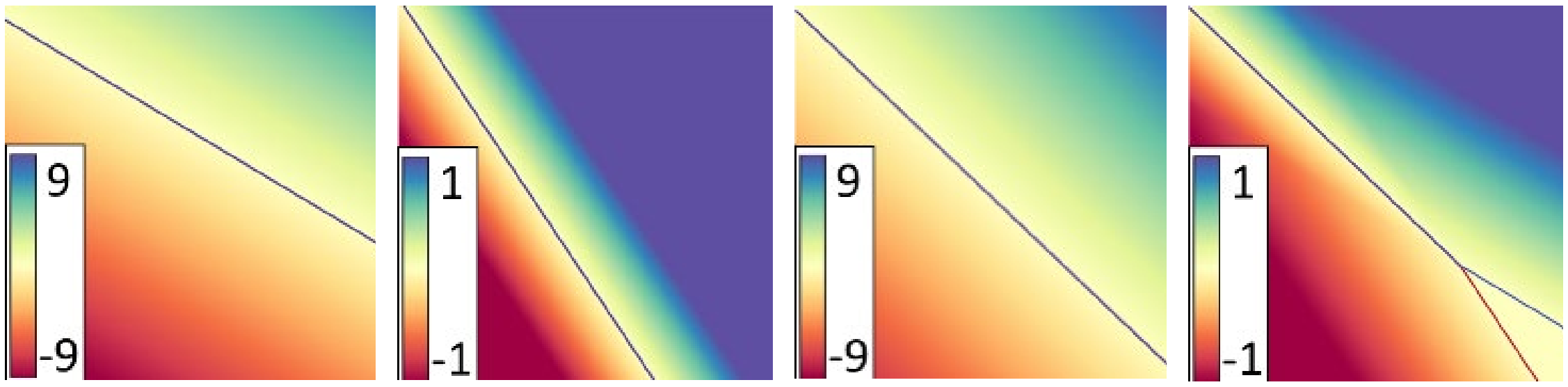

(a) SDF of Line a (b) TSDF of Line b (c) Blended SDF (d) Blended TSDFs

Figure 3.4 Illustration of truncated boundary in terms of TSDF fusion. Color legends represent normalized units.

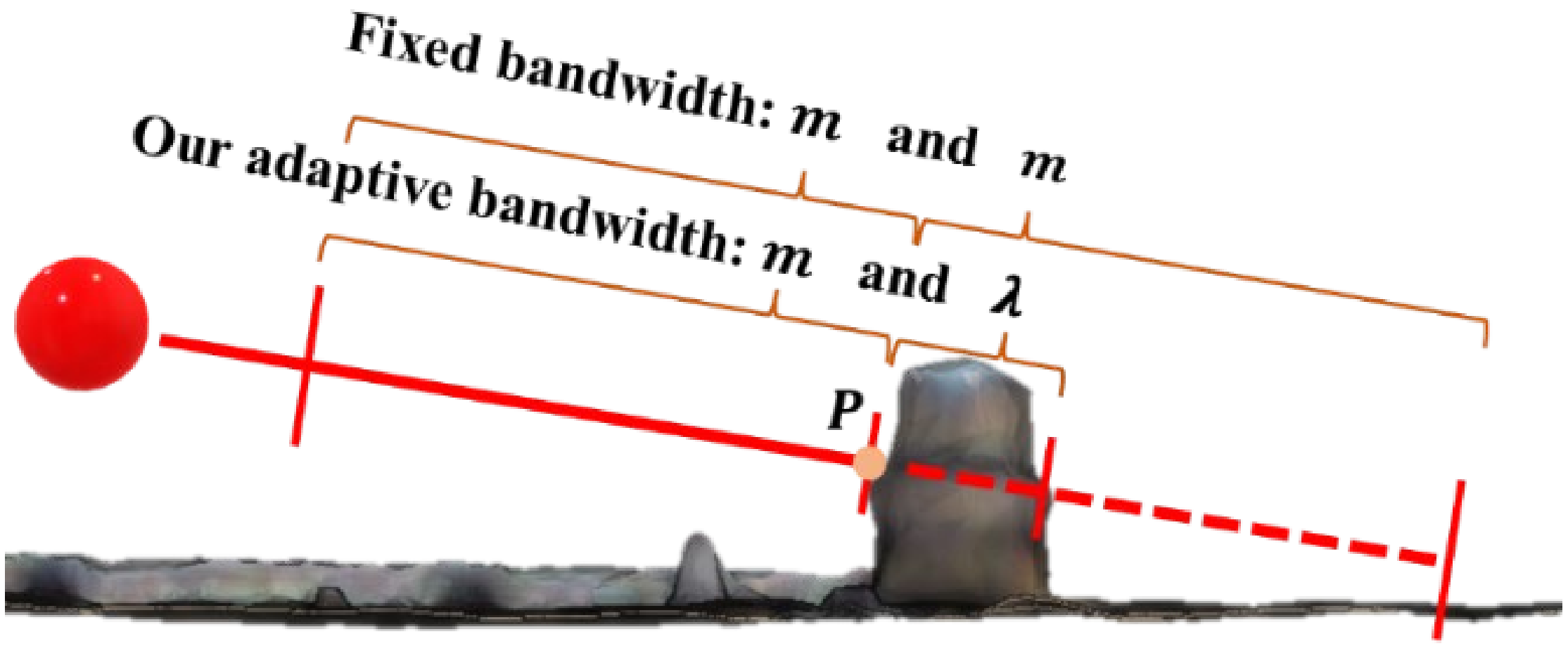

Figure 3.5 Adaptive bandwidth determination based on surface bounds. The areas crossed by the solid line represent the positive distance (outside of the object) while the dashed line indicates the negative distance (inside of the object). A fixed bandwidth in the classic TSDF would result in the contamination of empty regions, whereas our adaptive bandwidth does not suffer from this issue.

### 3.5 Experiments

#### 3.5.1 Implementation

Our implementation makes use of the OpenVDB (Museth 2013), a B+trees based data structure and library for efficient storing and processing large-scale volumetric data. VDBFusion (Vizzo et al. 2022) is a utility library that implements the TSDF integration pipeline (Section 3.4.1) using OpenVDB as the data structure. We verified the effectiveness of our approach by testing it on two oblique photogrammetric meshes that were captured by a drone at different altitudes representing different resolution and quality. The reconstructed meshes are of varying levels of uncertainty, resolution, and coverage. Our experiments were conducted on a Windows 10 workstation equipped with an Intel i7-8700, 32GB RAM, and the maximal bandwidth was set as 5 meters.

#### 3.5.2 Experiments on Mesh Model Conflation

In this experiment, we demonstrate the conflation of two mesh models with different resolution and accuracy. The first model encompasses a larger area, yet it exhibits lower resolution. With 81K vertices and 167K triangles, it spans an area of 24,422 $m^2$. Conversely, the second model comprises 514K vertices and 1M triangles, covering an area of 2,961 $m^2$, which is approximately 12% of the first model. The second model is characterized by finely detailed structures, as illustrated in Figure 3.6a.

We compare the performance of three methods: simple overlay, a remeshing algorithm called APSS (Algebraic Point Set Surfaces) (Guennebaud and Gross 2007), and our approach using both qualitative and quantitative metrics. APSS is a local method based on Moving Least Squares (MLS), which computes an implicit function from local point structures and converts these to a triangle mesh using the marching cubes algorithm (Newman and Yi 2006). The reference data is generated by photogrammetry software through a single and higher resolution collection of the region. It has 21M vertices and 9M triangles.

As shown in Figure 3.6a, a simple overlay method results in meshes intersecting with each other due to the data uncertainties (yellow and grey represent two mesh models). Since the simple overlay keeps individual meshes in their original form, the resulting topology of the meshes is incoherent. We also compared the APSS implemented in MeshLab and decimate it to have the same number of faces as our method, as shown in Figure 3.6b. While this method has merged the surfaces and preserved high-resolution mesh, it generates holes indicating the bad topology quality in overlapping regions. In contrast, Figure 3.6c show our conflated mesh, which is coherent, smooth, complete, and compact.

We analyze the distance between results and the reference mesh in two ways: the mean distance to the reference mesh, as well as the F-Score (Knapitsch et al. 2017). F-score is the harmonic mean between precision and recall, reflecting the proximity of the reconstructed points to the reference mesh and its completeness. Using a threshold of 0.5 meters (equal to the voxel size in our experiments and the highest resolution of source meshes), differences smaller than 0.5m are considered acceptable. Table 3.1 verifies that our method achieves the best result, yielding comparable accuracy to the original mesh, with better F-score.

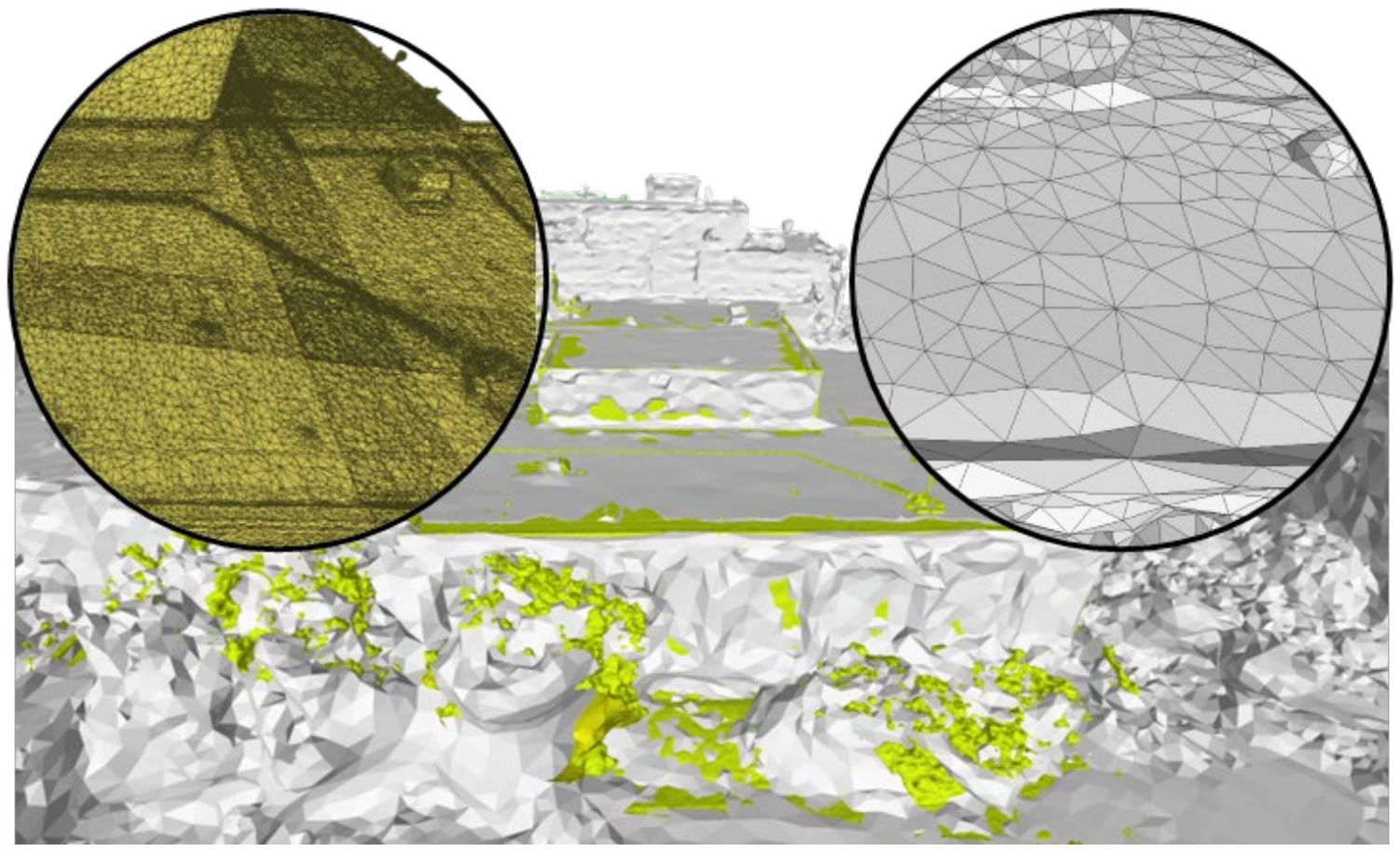

(a) Simple overlay method.
High-res dataset (yellow) and low-res dataset (grey).

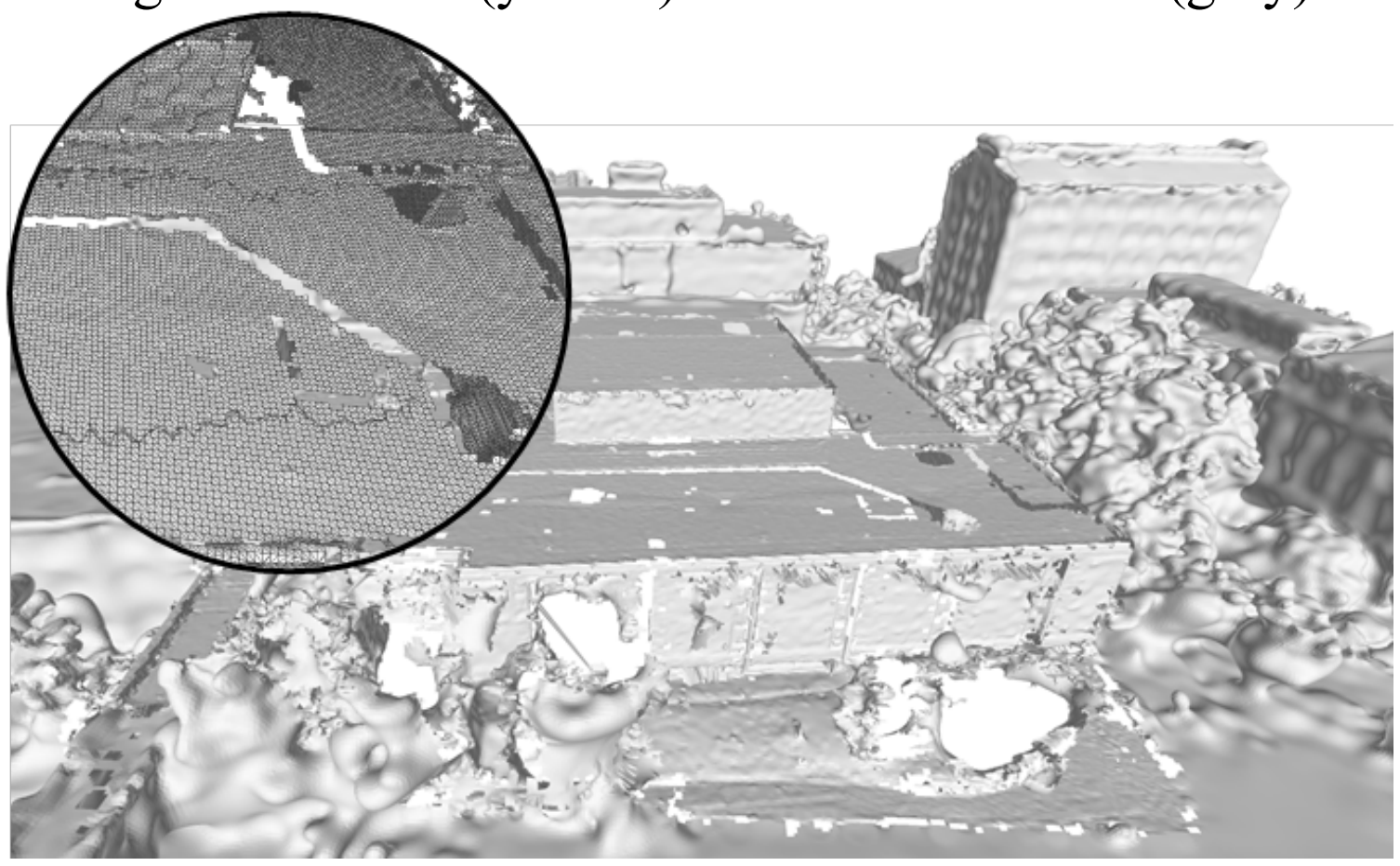

(b) APSS remeshing method in MeshLab (Cignoni et al. 2008).

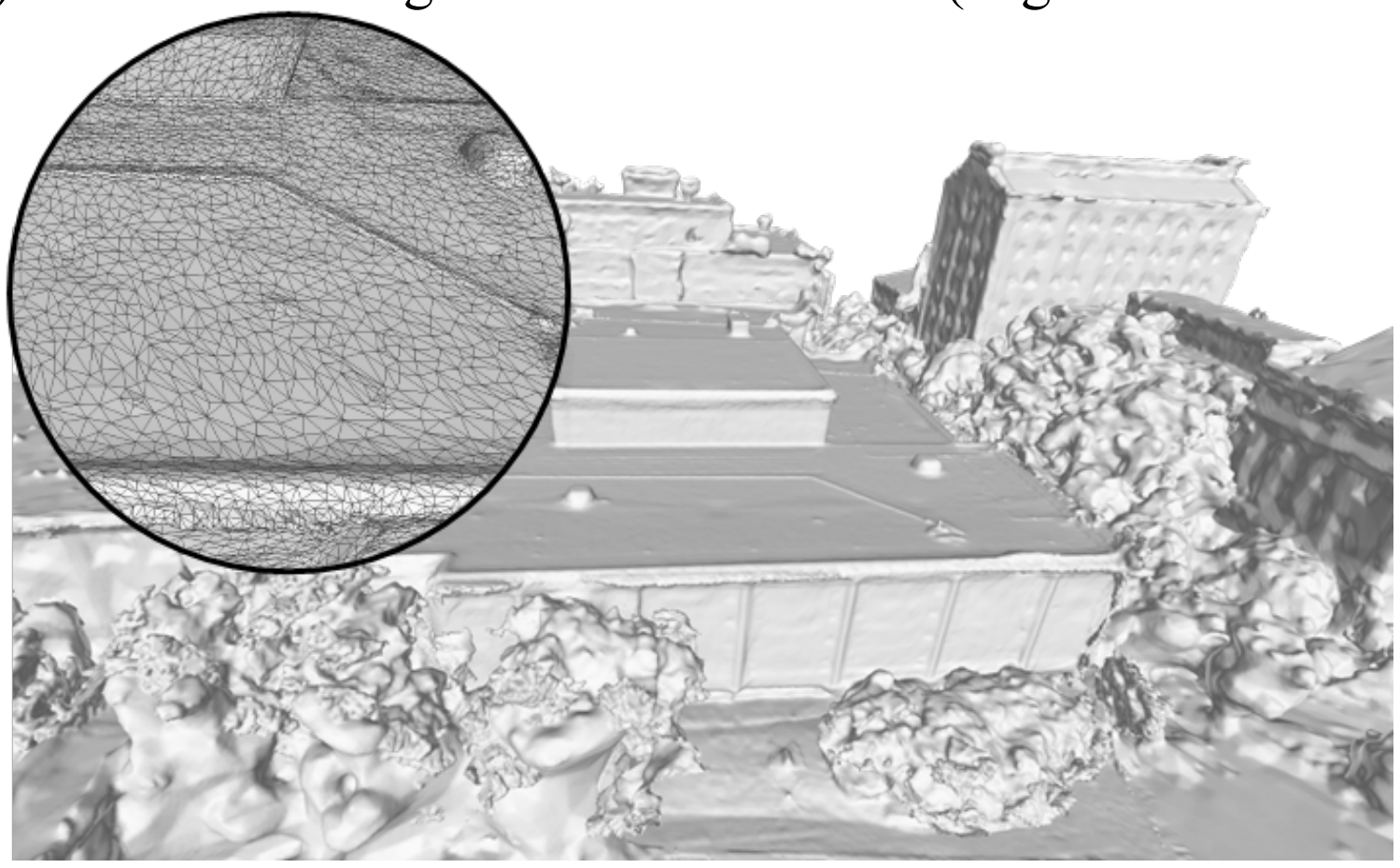

(c) Our conflated mesh.

Figure 3.6 Qualitative comparison of our conflation method with APSS (Guennebaud and Gross 2007) remeshing method.

| Methods | Mean distance | F-score | # Vertices | # Faces |
|---|---|---|---|---|
| **Overlay** | 0.488 m | 0.364 | 594 K | 1.2 M |
| **APSS** | 0.514 m | 0.365 | 569 K | 1.1M |
| **Ours** | 0.503 m | 0.367 | 773 K | 1.1M |

Table 3.1 Quantitative results of model conflation methods comparing with the reference mesh.

### 3.5.3 Computational Complexity

The computational complexity of our method is relevant to several steps: ray casting, TSDF conflation, and marching cubes. Ray casting utilizes highly optimized methods, with a complexity that could be $O\left(N * log(M)\right)$ or even lower, where $N$ is the number of rays and $M$ represents the number of triangles (Y. Zhang, Gao, and Li 2016). Leveraging OpenVDB, which can operate voxels in $O(1)$, the complexity of TSDF fusion process used by our proposed method is $O(N)$, as each ray could potentially influence a constant number of grid cells. The marching cubes algorithm, which scans over the activated voxels (truncated band) and visits only their local neighbors, carries a complexity of $O(P)$, where $P$ is the total number of voxels in the truncated band. This count correlates with the surface area of the input meshes. To sum up, our proposed method demonstrates better than linear complexity with respect to $N$, $M$, and $P$, offering significant scalability for larger datasets.

### 3.6 Conclusion

In this chapter, we presented a novel approach for conflating full-3D mesh models in support of large site modeling, an underserved task in oblique photogrammetry. Our approach adapts a novel virtual panoramic camera concept under a TSDF implicit surface representation. Our preliminary results demonstrated that the conflated model from our method excels existing methods both visually and statistically for complex scenes. It is 4 times more compact and 1.9% more accurate than a typical remeshing method with a coherent topology. As compared to existing practice that presents the earth surface in 2.5D raster, the approach developed in this chapter has the potential to enable future full-3D mesh modeling for large area by leveraging data from different sources including photogrammetry and LiDAR data. However, it is important to note that our method currently supports only the exterior surface of geospatial models due to the assumption made in virtual camera sampling.

# Chapter 4. A General Albedo Recovery Approach for Aerial Photogrammetric Images through Inverse Rendering

This chapter is based on paper called “A Novel Intrinsic Image Decomposition Method to Recover Albedo for Aerial Images in Photogrammetry Processing” that was presented in the “ISPRS Annals of the Photogrammetry, Remote Sensing and Spatial Information Sciences (XXIV ISPRS Congress 2022)” by S. Song and Qin, 2022, and “A General Albedo Recovery Approach for Aerial Photogrammetric Images through Inverse Rendering” that was submitted to “ISPRS Journal of Photogrammetry and Remote Sensing”. We present the difference appearance between the raw images taken by camera and our processed albedo images in Figure 4.1.

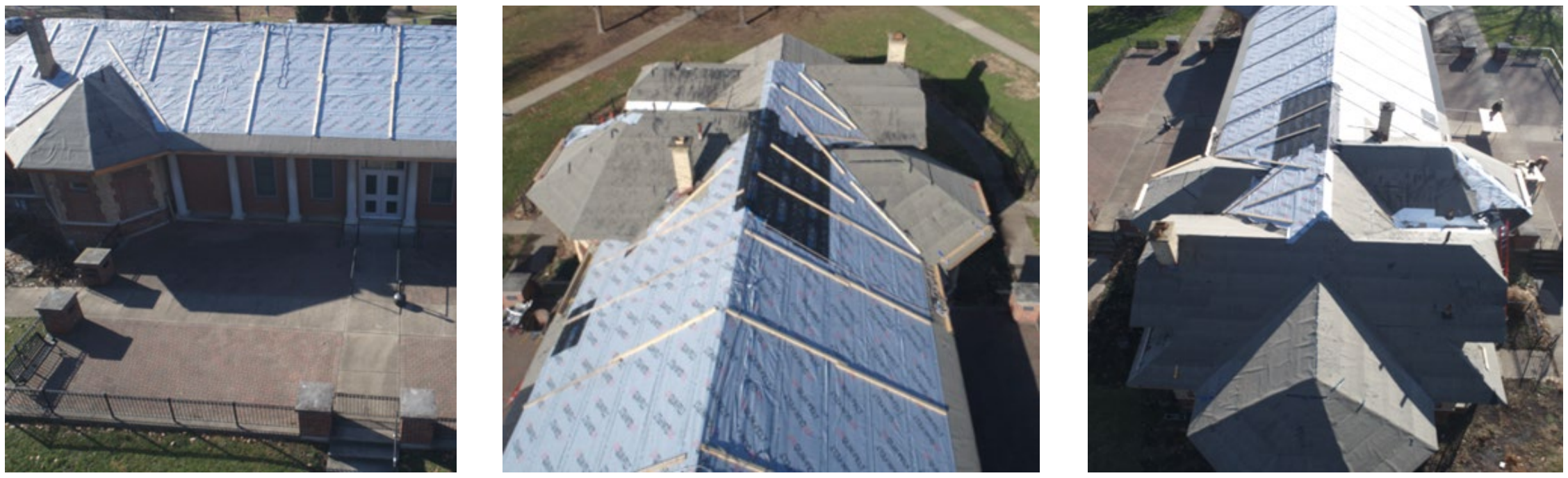

(a) Raw images.

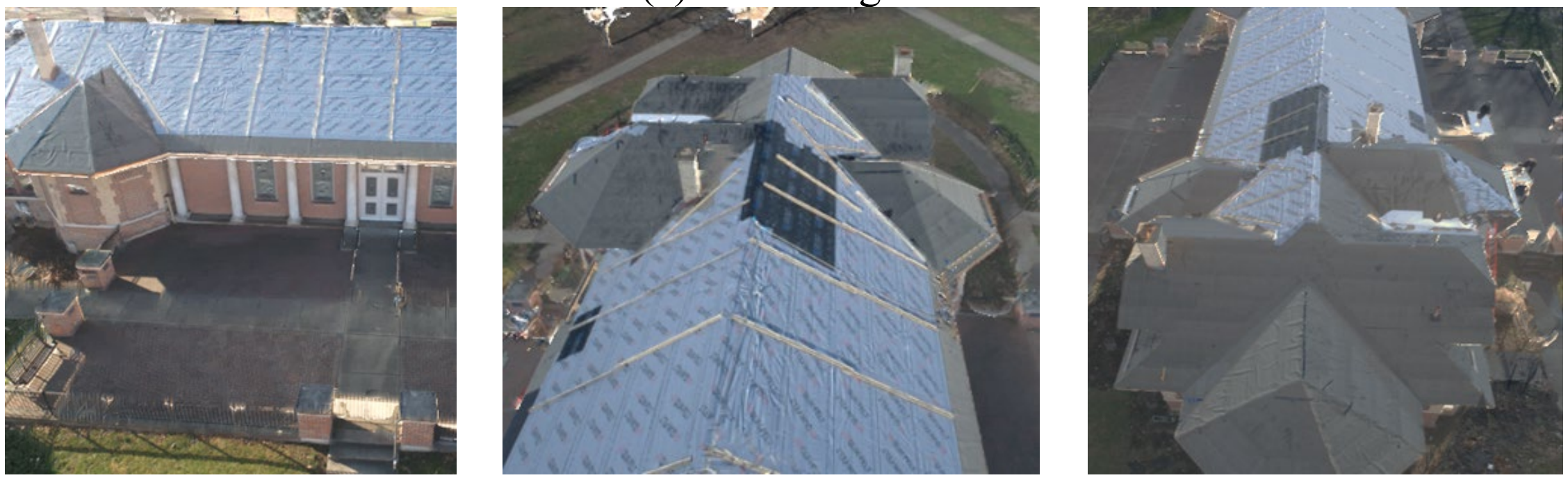

(b) Our albedo images.

Figure 4.1 Comparison between raw images and recovered albedo images using our method.

## 4.1 Chapter Abstract

Modeling outdoor scene for the synthetic 3D environment requires the recovery of reflectance/albedo information from raw images, which is an ill-posed problem and remain unexplored. The recovered albedo can facilitate model relighting and shading, which can further enhance the realism of rendered models and the applications of digital twins. Typically, photogrammetric 3D models simply take the source images as texture materials, which inherently embed unwanted lighting artifacts (at the time of capture) into the texture. Therefore, these "polluted" textures are suboptimal for a synthetic environment to enable realistic rendering. In addition, these embedded environmental lightings further bring challenges to photo-consistencies across different images that cause image-matching uncertainties. This paper presents a general image formation model for albedo recovery from typical aerial photogrammetric images under natural illuminations and derives the inverse model to resolve the albedo information through inverse rendering intrinsic image decomposition. Our approach builds on the fact that both the sun illumination and scene geometry are estimable in aerial photogrammetry, thus they can provide direct inputs for this ill-posed problem. This physics-based approach does not require additional input other than data acquired through typical drone-based photogrammetric collection and was shown to favorably outperform existing approaches. We also demonstrate that the recovered albedo image can in turn improve typical image processing tasks in photogrammetry such as feature and dense matching, edge, and line extraction.

## 4.2 Introduction

Aerial photogrammetry nowadays has been sufficiently automated that it can (almost) generate high-resolution photorealistic models from well-collected images with a few clicks, using commercial/open-source software packages (Agisoft 2023; Bentley 2022) . Beyond its well-known applications to support foundational mapping, the reality-based models from photogrammetry are gaining thrusts in domains that require simulation and immersive sciences and engineering, such as virtual, augmented, extended, mixed reality (VR/AR/XR/MR), metaverse, and digital twin applications (Alidoost and Arefi 2017). However, the use of photogrammetric models in these domains is still very limited, part of the reason is that the texture materials of these models are not the actual "albedos" needed by the rendering pipeline in computer graphics (Innmann, Susmuth, and Stamminger 2020; Lachambre 2018). Rather, these texture materials are often directly inherited from the source images, in which the environmental lighting is unavoidably present and regarded as artifacts. For example, an albedo texture image is free of shadows so the graphics rendering pipeline can relight the model in a simulated environment, while the texture materials from source images may contain unwanted shadows that hinder the realism of the rendered views of the model. Moreover, recovering the albedos from the images may bring added benefit for necessary steps in photogrammetric processing, such as feature extraction & matching, and dense image correspondences (S. Song and Qin 2022). Therefore, this not only enhances the use of photogrammetric models for extended applications but also benefits the photogrammetric process.

Albedo recovery for high-resolution images is mostly studied in the computer vision and graphics community, where given an input image, the per-pixel albedo map can be recovered through a process called inverse rendering, or Intrinsic Image Decomposition (IID). However, among all the existing literature, albedo recovery for images of outdoor scenes remains an unexplored problem, mainly due to that capturing the complex outdoor lighting information is extremely difficult and thus it is often hard to decouple the albedo information from the shading caused by the unknown lighting (Duchêne et al. 2015; Laffont, Bousseau, and Drettakis 2013).

Despite these challenges, we notice that this lighting and shading information may be partially estimated in the photogrammetry context (Chi et al. 2023): first, photogrammetric collection tasks often come with auxiliary data that contains information about the capture time and location (e.g., GPS (Global Positioning System) data), with which the solar illumination at the data capture time may be estimated. Second, photogrammetric images are often collected as aerial or oblique blocks, where reasonably accurate surface geometry can be derived to simulate shading cast by solar illumination.

Therefore, we consider the albedo recovery of photogrammetric images a tractable problem, and thus, propose a general physics-based albedo recovery approach that performs inverse rendering, or intrinsic image decomposition. The proposed method takes the above-mentioned specifics of photogrammetric images as cues to model the in-situ illumination and shading and invert the albedo from the source images. Earlier our published work (S. Song and Qin 2022) presented a solution of this idea, and this work further extends it with the following contributions.

1. We present a more general lighting model under photogrammetric collections, which makes full use of the local geometry to model the environmental lighting.

2. We further extend our experiments by comprehensively evaluating its scalability towards large and diverse scenes, with added experiments demonstrating the benefit of the albedo recovery for various low-level vision tasks, to inform its practical potential as part of the photogrammetric data processing pipeline.

The rest of this chapter is organized as follows: Section 4.3 introduces related works, and Section 4.4 elaborates on the proposed outdoor lighting model consisting of directed sunlight and hemispheric skylight. Section 4.5 describes our approach to estimating the proposed outdoor lighting model from a multi-view image set. In Section 4.6, we evaluated our method both quantitatively and qualitatively based on a synthetic dataset and a multi-temporal real-world UAV dataset. Section 0 demonstrates three promising applications using our albedo imagery in fields of research and industry, and Section 4.8 concludes this chapter.

### 4.3 Related Works

There have been many works in the literature that aimed to recover the albedo reflectance from single or multiple-view images in the field of Computer Vision (CV), primarily under the umbrella of Intrinsic Image Decomposition (IID), which estimates albedo, shading, and normal in the view space (Barron and Malik 2015; Garces et al. 2022). Major efforts are to integrate learned geometric cues, or end-to-end deep learning-based estimation (Das, Karaoglu, and Gevers 2022; Z. Li and Snavely 2018; Janner et al.

2017). However, these works reflect mostly indoor scenes and with poor generalization capability. Therefore, a physics-driven method is necessary. In our context, as described in Section 4.2, the photogrammetric images indirectly provide the scene geometry with potentially estimable scene lighting, therefore adding the “physics” component into the solution. Moreover, the problem of albedo recovery is relevant to shadow removal, which presents great literature in the domain of CV, photogrammetry, and remote sensing. Considering these facts, in this section, we briefly review works related to both IID and shadow removal.

**Intrinsic image decomposition** studies the decomposition of a single image to intrinsic layers (Barrow H.G. and Tenenbaum 1978) that include diffuse albedo (or reflectance) and shading (or pixel-level illumination). The decomposition is an ill-posed problem since both the geometry and environmental lighting are oftentimes unknown. Therefore, solutions heavily rely on regularization priors. A few early works consume user-supplied strokes to regularize the problem (Bousseau et al. 2009; Shen et al. 2011), providing low-level priors of ambient reflectance samples. Most methods assume a sparse or piecewise constant albedo (Gehler et al. 2011; Sheng et al. 2020) and smooth monochromatic illumination. Automatic methods explicitly detect shadow through supervised methods (Griffiths, Ritschel, and Philip 2022) or estimate complex shading with neural networks (Innamorati et al. 2017; Janner et al. 2017; T. Wang et al. 2021; Yu and Smith 2019; 2021). These methods were proven successful for indoor scenes where the data are considered “in the same domain” as the training data. There are a few works that focus on IID for outdoor scenes, for example, by using multi-view images, Laffont, Bousseau, and Drettakis (2013) use a physics-based ray-tracing engine to calculate shading for the purpose of reflectance recovery, while it requires in-situ measurements on site to record the environment lighting. Duchêne et al. (2015) extracted partial cues of sky irradiance from ground images to estimate the environment irradiance, and this is partly effective since the to-be-corrected images and sky information come from the same images and thus, it does not require inter-sensor/image calibration. In an aerial/drone photogrammetry scenario, since these images do not look up to the sky, such an idea is not directly applicable. In contrast, estimated solar lighting from the position and timing (GPS and clock), which is more practical to obtain, pertains a great interest to scalable solutions like our proposed method.

**Shadow removal** studies the extraction and removal of cast shadows from single or multiple images. Early works focused on images with clean scene structures such as those with clean foregrounds and backgrounds, where single and isolated objects cast distinct shadows (Finlayson, Drew, and Lu 2004). For such a problem, simple heuristics can be applied for shadow detection followed by shadow removal. Recent studies train neural networks to perform end-to-end shadow detection and removal, which has demonstrated great successes (Cun, Pun, and Shi 2020; Qu et al. 2017; J. Wang, Li, and Yang 2018), but like many other deep learning approaches, its generalizability is of concern in practice. More practical solutions involve human intervention such as using strokes as guidance of the shadow region (Gong and Cosker 2017). Regarding aerial images, shadow removal has been one of the core tasks for the photogrammetry and remote sensing community such as orthophotos production (Rahman et al. 2019; Silva et al. 2018; T. Zhou et al. 2021), semantic segmentation and object detection, and 3D reconstruction. One relevant line of work (Q. Wang et al. 2017), makes use of geometry (i.e., Digital Surface Models (DSM)) to predict shaded regions using the known direction of the solar illumination, on which shadow removal can be performed using pixels on both sides of the shadow boundaries (Guo, Dai, and Hoiem 2013; Luo et al. 2019; Luo, Li, and Shen 2017). As mentioned earlier, since the photogrammetric images indirectly provide the geometry (Duchêne et al. 2015; Laffont, Bousseau, and Drettakis 2013), our method will take advantage of such an approach, but with more comprehensive lighting modeling to recover not only shadows but general shadings caused by non-directional solar radiation (Section 4.4).

## 4.4 Modeling: Physics-based Aerial View Rendering with a General Lighting Model

To achieve inverse rendering, we will first formulate the aerial image rendering process with our proposed lighting model. Specifically, given the positing and timing information, we model the environmental lighting and then perform the aerial view rendering using the modeled lighting.

### 4.4.1 The Rendering Equation

The general rendering equation (Kajiya 1986) models the observed radiance of surface point $p$ by a camera $L_o$, which consists of the surface-emitted radiance $L_e$ (e.g., luminescent objects) and reflected radiance from light sources $L_i$ (Equation (4.1)). In the context of daylight optical passive sensing, $L_e$ is often not discussed since its contribution is relatively weak. The reflected radiance of a surface point is generally an integral of incident radiance $L_i$ interacted with surface BRDF (Bidirectional Reflectance Distribution Function) $f_r$ over a considered differential hemisphere Ω.

$$L_o(p, \omega_o) = L_e(p, \omega_o) + \int_\Omega L_i(p, \omega_i) f_r(p, \omega_i, \omega_o)(\omega_i \cdot n)^+ d\omega_i, \tag{4.1}$$

where $\omega_i$ is the inlet direction of radiance from the light sources, $\omega_o$ is the outward direction from the surface to the camera, $(\cdot)^+$ is the ramp function which equivalent to $max(0, \cdot)$. At this stage, for simplicity, specular reflection will not be considered directly in the rendering equation, alternatively, if presented, they will be baked into the albedo as expected artifacts. Therefore, we will adopt the Lambertian model (Koppal 2014) as the BRDF for our rendering model as shown in Equation (4.2).

$$f_r(p, w_i, w_o) = \frac{\rho}{\pi} \;, \tag{4.2}$$

where ρ is albedo, which can be interpreted as the intrinsic color of the material, alternatively regarded as the reflectance in the field of remote sensing and photogrammetry. The Lambertian model describes a perfect diffusion reflection that is only affected by the incidental light but not the viewing direction (Koppal 2014).

### 4.4.2 Camera and Sensor

The rendering equation describes the physics-based modeling of light transport and its interaction with the surface materials resulting in the observed radiance. The sensor (film and its associated electronics, i.e., CCD or CMOS) of the camera records the observed radiances and processes them into machine-readable digital signals (pixel values) through an A/D (Analog / Digital) converting process. On top of this process, various standard internal processes were conducted by the camera, including tune mapping, white balancing, and sometimes compression. This facilitates a perceptually well-balanced image for visualization, however, may add complexities when interpreting radiance out of it. As a result, image pixel values may not be linearly correlated to the scene radiance (Grossberg and Nayar 2004), thus is challenging to model. Luckily, most of the aerial and drone-based cameras allow to export of raw images at the collection, which produce raw pixel values with minimal internal processes. Therefore, raw pixel value intensity $I$ can be assumed the following linear relationship with the radiance $L_o$ (Equation (4.3)), where $\epsilon$ is a scale factor that can be interpreted as the exposure factor.

$$I = \epsilon L_o \,. \tag{4.3}$$

### 4.4.3 Modeling Outdoor Illumination

Outdoor illumination consists of both directional illumination from solar radiation and ambient illumination from the sky (scattered illumination, hereafter called sky illumination). This interprets the inlet lighting $L_i$ as a compounded source from both Sun $L_i^{sun}$ and Sky illumination $L_i^{sky}$ (Equation (4.4)), time-varying lights based on solar radiation (with a time variable $t$).

$$L_i(p, \omega_i, t) = L_i^{sun}(p, t) + L_i^{sky}(p, \omega_i, t) \,. \tag{4.4}$$

It should be noted that indirect lights, such as lighting reflected from other objects, will not be considered in the aerial case since they are ignorable as stated in our earlier work (S. Song and Qin 2022)

### *Modeling Sun Radiance*

Despite the Sun illumination should be theoretically modeled as a point/area light source, due to its far distance to Earth, the solid angle of the sun from the Earth can be as small as $0.68 \times 10^{-4}$ steradians (L. Wald 2018). Thus, it is mostly assumed to be parallel and directional light to a region of interest. Therefore, the inlet sunlight to a surface point $p$ can be effectively modeled in the following (Equation (4.5)):

$$L_i^{sun}(p,t) = \psi^{sun}(p,t) \cdot V^{sun}\big(p, \omega_{sun}(t)\big) \;, \tag{4.5}$$

where $\omega_{sun}$ refers to the Sun angle as a function of time $t$, which represents the lighting direction. $V^{sun}$ refers to the visibility of the sun when considering a surface point $p$ (determined by the local geometry of the scene and sunlight direction), as it can be possibly occluded (e.g., shadows). $\psi^{sun}$ refers to the source intensity of the sun, a function of its incident angle (determined by the location $p$), and data collection time. In the context of albedo correction, $\psi^{sun}$ is represented as $\psi^{sun} \in \mathbb{R}^3$ (RGB) to consistent with the image. In practice, the intensity of $\psi^{sun}$ can be defined relative to unit 1, referring to the largest magnitude of a year. Therefore, given the surface point $p$, local geometry (determined by photogrammetric 3D reconstruction), and the time (determining the sunlight direction and strength), $L_i^{sun}$ is directly calculable.

*Modeling Sky Radiance*

When traveling through the atmosphere, the sunlight can be scattered through the aerosol and various layers, creating a domed light source, with centers around the sunlight direction. An extreme example of such is a cloudy day, where direct sunlight and strong shadows are not observable. Following the definition of a dome light, we model the skylight as the following (Equation (4.6)):

$$L_i^{sky}(p, \omega_i, t) = \psi^{sky} \cdot G\big(\omega_i - \omega_{sun}(t)\big) \cdot V^{sky}(p, \omega_i) \ , \tag{4.6}$$

where $V^{sky}$ is skylight visibility at a surface point $p$ (determined by the local geometry of the scene) observing the direction $\omega_i$. $\psi^{sky}$ is the source intensity of the sky, which is considered a constant value by assuming a uniform skylight. $G(\omega_i - \omega_{sun}(t))$ refers to a Gaussian function (with a controllable, but large variance) indicating the maximal intensity still occurs at the solar illumination direction.

Different from our earlier work (S. Song and Qin 2022), this chapter combined skylight visibility (as shown in Figure 4.2) and uniform source intensity to create a non-homogenous skylight, closer to the heterogeneity nature of such a lighting model.

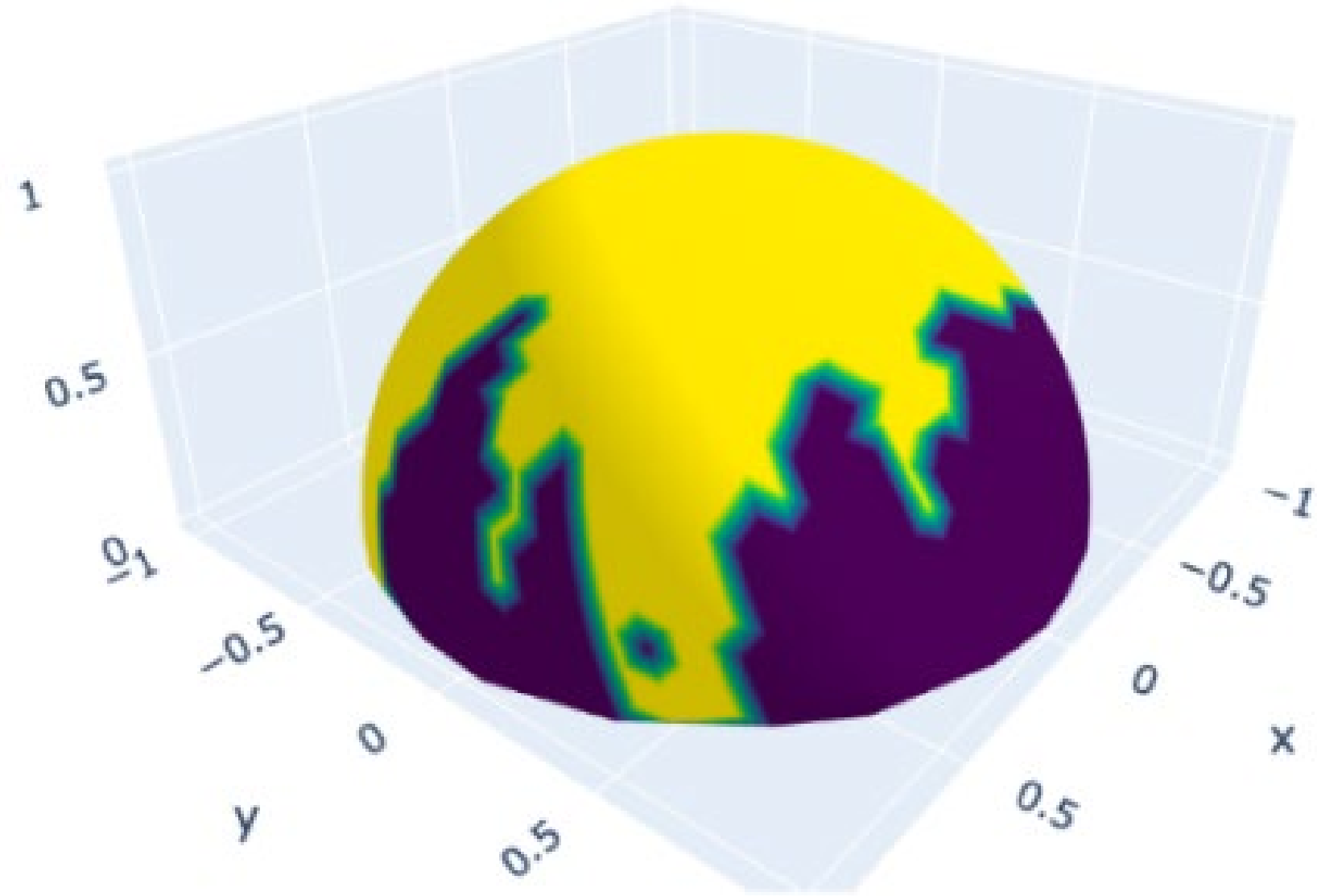

Figure 4.2 Example of $V^{sky}$ from a surface point (center of the hemisphere).

### 4.4.4 Connecting Sun and Sky Illumination

As described in Section 4.4.3, with photogrammetrically reconstructed 3D geometry, $L_i^{sun}$ is calculable given a surface location and time. However, $L_i^{sky}$, as shown in Equation (4.6), is not directly calculable using the same information, primarily due to that the light intensity constant $\psi^{sky}$ is unknown. Considering that this quantity still origins from the sunlight ($\psi^{sun}(p, t)$), our aim is to seek for the relationship between them. We noted a critical fact: in an ideal case, the shadowed region, due to occlusion, contains no sunlight $\psi^{sun}(p, t)$, but it will still be illuminated by the skylight, since the skylight is a domed light that comes from all directions. Whereas non-shadowed region contains both sunlight and skylight. This implies that by using pixel intensity values between shadowed and non-shadowed regions, it is possible to build the relations among these two quantities ($\psi^{sun}$ and $\psi^{sky}$), thus resolving $L_i^{sky}$ as a function of known $\psi^{sun}$. The pixel intensity values in shadowed and non-shadowed regions can be easily observed by means of lit-shadow pixel pairs, where a pair of pixels span between shadowed and non-shadowed regions, such shown in Figure 4.3, can be found. It should be noted that the distance of two pixels $p$ and $p + \Delta$ is best minimal to remove other factors, such as non-homogenous camera sensor responses, as well as the change of sky visibilities $V^{sky}$ due to distant locations. Land and McCann (1971) proposed to assume albedo ρ across the scene should be mostly with low frequency, i.e., piecewise linear or constant. Therefore, taking this assumption, we can reasonably assume that these two points, since they are close enough, share the same albedo ρ. At the same time, we assume that the visibility of the sky $V^{sky}$ of these two points are the same: because the shadow and non-shadow region only reflect the visibility difference of light from a single direction (the same as the direct sunlight), while $V^{sky}$ considers accumulated impacts of all directions, thus differences by a single light direction can be ignored.

These facts can be implemented by building the following observational constraints between lit-shadow points (Equation (4.7)).

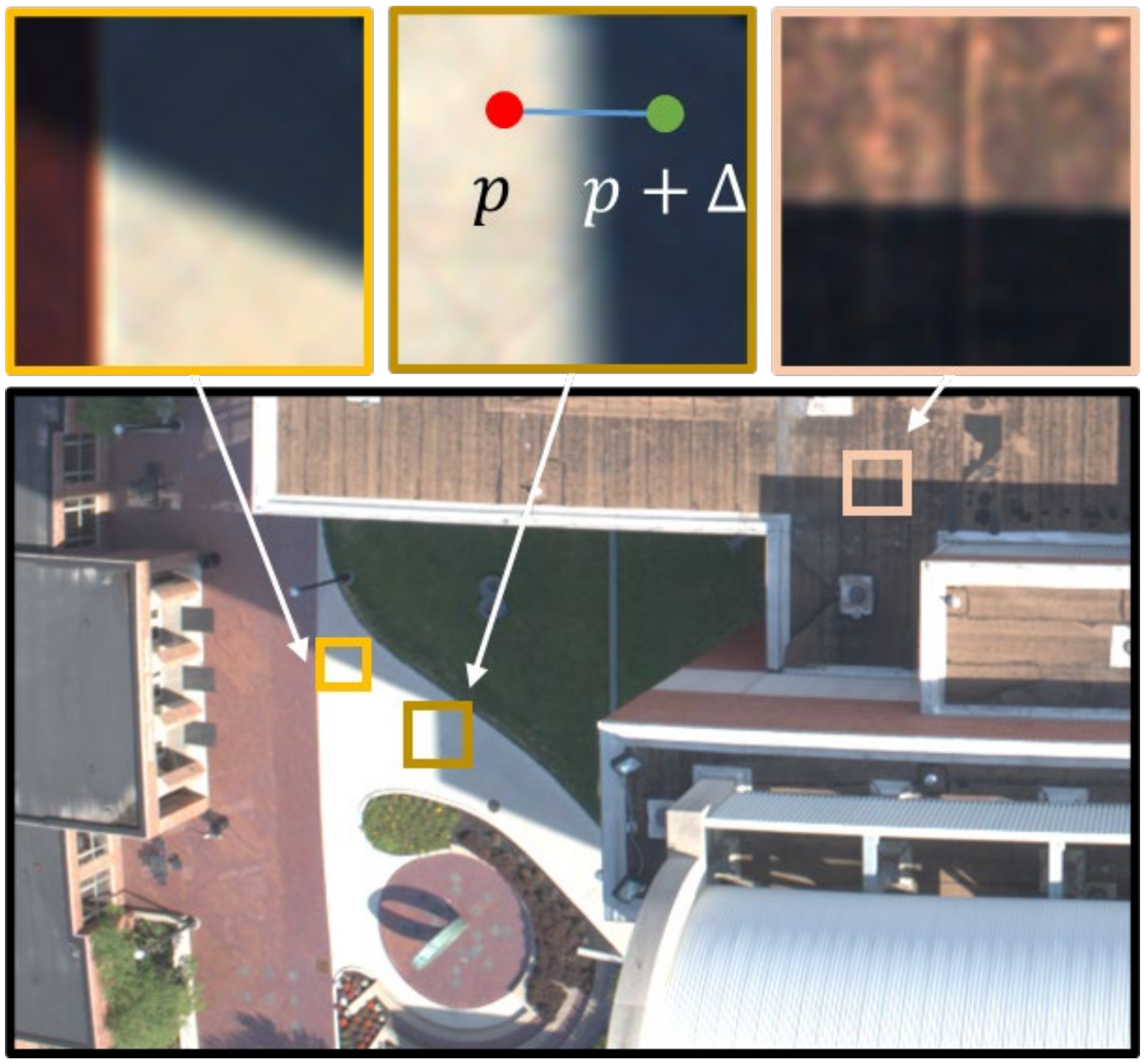

Figure 4.3 Lit-Shadow pair near the shadow boundary. The figure shows 3 patches contains casted shadows with penumbra width (transitions between shadow to non-shadow region).

$$\begin{cases} \rho_{lit} & = & \rho_{shadow} = \rho \\ n_{lit} & = & n_{shadow} = n \\ V_{lit}^{sky} & = & V_{shadow}^{sky} = V^{sky} \\ V_{lit}^{sun} & = & 1 \\ V_{shadow}^{sun} & = & 0 \end{cases} . \tag{4.7}$$

Then, we expand Equation (4.4) for these two lit-shadow points, which leads to the following (Equation (4.8)). Note for simplicity, we ignored the variable $t$, since these two points (in the same image) are collected at the same time:

$$\begin{cases} L_o(p) = & \rho\psi^{sun}(p)(\omega_{sun} \cdot n)^+ + \rho\psi^{sky}\int_\Omega V^{sky}(p,\omega_i)G(\omega_i - \omega_{sun}(t))(\omega_i \cdot n)^+ d\omega_i \\ L_o(p+\Delta) = & \rho\psi^{sky}\int_\Omega V^{sky}(p,\omega_i)G(\omega_i - \omega_{sun}(t))(\omega_i \cdot n)^+ d\omega_i \end{cases}, \quad (4.8)$$

where the sunlight and skylight shading, as part of Equation (4.8), can be denoted as $S^{sun}$ and $S^{sky}$ (Equation (4.9)):

$$\begin{cases} S^{sun} \coloneqq (\omega_{sun} \cdot n)^+ \\ S^{sky} \coloneqq \int_\Omega V^{sky}(p,\omega_i)G\big(\omega_i - \omega_{sun}(t)\big)(\omega_i \cdot n)^+ d\omega_i \end{cases}, \quad (4.9)$$

Here, we assume another simplification: given that the sky visibility is minimally impacted by local geometry, we can assume its visibility component $\int_\Omega V^{sky}(p,\omega_i)d\omega_i$ is close to full visibility (i.e., 1), while the cumulated incident angle $\int_\Omega G(\omega_i - \omega_{sun}(t))(\omega_i \cdot n)^+ d\omega_i$ is biased towards the sunlight incident angle $(\omega_{sun} \cdot n)^+$. This leads to the following conclusion (Equation (4.10)): where the skylight intensity $\psi^{sky}$ and sunlight $\psi^{sun}$ is up to a constant factor calculable using lit-shadow pairs (detectable, introduced in Section 4.5.2).

$$\phi \coloneqq \frac{\psi^{sky}}{\psi^{sun}} = \frac{L_o(p+\Delta)}{L_o(p) - L_o(p+\Delta)} \quad . \quad (4.10)$$

As a result, given a surface location $p$, local geometry (from photogrammetry), and collection time $t$, both the $L_i^{sun}$ and $L_i^{sky}$ are calculable.

## 4.5 Solution: Inverse Rendering for Albedo Recovery

Based on Section 4.4, with photogrammetric images and their associated location and collection time, the inlet source lights, consisting of the direct sunlight $L_i^{sun}$ and skylight $L_i^{sky}$, are calculable. Therefore, based on the general rendering equation (Equation (4.1)), emitted light $L_e(p, \omega_o)$ ignored under our context), since $L_i(p, \omega_i)$ is calculable, as well as the geometry (normal $\boldsymbol{n}$, derived from photogrammetry), the albedo (denoted as the BRDF $f_r$) can be easily inverted. Hence, our solution built on this under the context of photogrammetric images can be depicted as a workflow shown in Figure 4.4. First, we perform standard photogrammetric data processing to calculate the pose of the images and generate the 3D meshes of the scene. It should be noted that for data collection, we require the users to store the raw images, as well as the metadata including the GPS location and the time of the data collection, for the purpose of calculating the inlet light. Second, we prepare these metadata to resolve the Sun visibility for both the sunlight and skylight ($V^{sun}$ and $V^{sky}$), which is used to compute the sunlight and skylight ($L_i^{sun}$ and $L_i^{sky}$). Third, the $L_i^{sky}$ is calibrated in reference to $L_i^{sun}$ is done by using a lit-shadow pairs introduced in Section 4.4.4**.** Fourth, the shadow predicted by Sun visibility $V^{sun}$ is often a binary mask, which cannot depict the penumbra effects (transition between shadow and non-shadow, shown in Figure 4.3), hence we propose to refine the $V^{sun}$ as a soft value (between 0-1 instead binary) to cope with this effect (to be introduced in Section 4.5.3), and finally optimize the recovered albedo ρ.

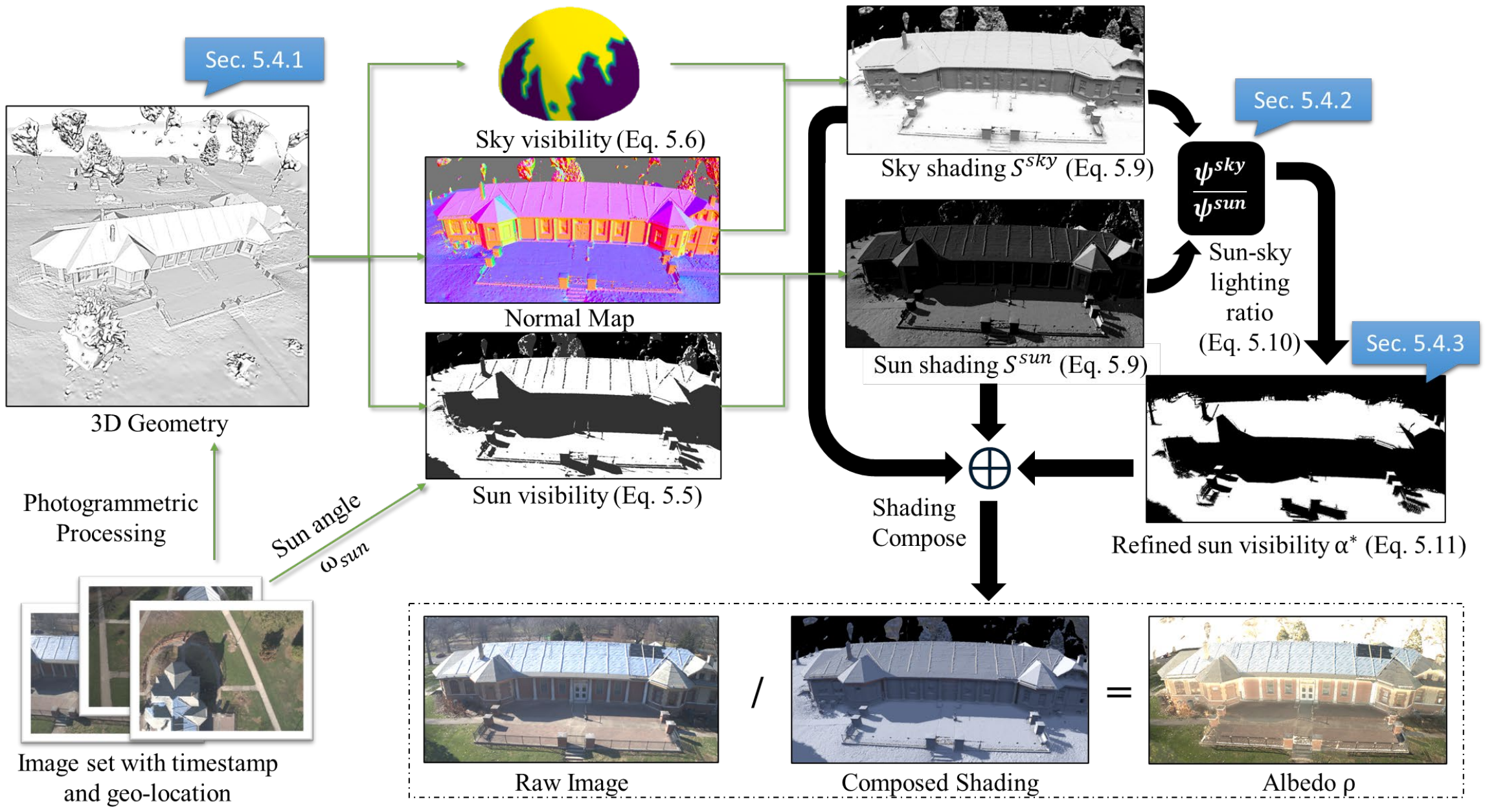

Figure 4.4 The workflow of our solution in albedo recovery.

### 4.5.1 Photogrammetric Data Preparation

We captured a regular photogrammetric block as the input and run through a standard photogrammetric data processing pipeline to orient images and generate the meshes using off-the-shelf commercial or open-source photogrammetric software such as Bentley ContextCapture (Bentley 2022). As described in Section 4.4.2, our albedo recovery method requires the pixel color intensity to be best linearly correlated with the radiance, we store the RAW images. If RAW images are not available, color space calibration algorithms can be performed pre- or post-flight (Lin et al. 2004; Tai et al. 2013).

### 4.5.2 Calculating Sunlight and Skylight

***Computing lighting components***

Using meta information, the sun's position $\omega_{sun}$ can be easily approximated by the astronomical almanac's algorithm (Michalsky 1988) with known geolocation and date time. For each view, we project rays from the camera center and apply the path-tracing technique (I. Wald et al. 2014) to detect their intersecting point to the surface for depth computation and occlusion detection. Surface normal can then be computed from the depth (Figure 4.5b). From the depth image, sun visibility $V^{sun}$ can be computed by emitting rays from every single pixel to the sunlight direction $\omega_{sun}$, and performing occlusion direction. This allows to compute the shading of both sunlight and the skylight ($S^{sun}$ and $S^{sky}$, Equation (4.9)), where $S^{sun}$ is easily computed by taking the sunlight direction, and $S^{sky}$ is computed by a sample of 1024 points over the hemisphere. Results of the shading components are visualized in Figure 4.5(c-d), which perceptually matches the intuition of shadings caused by direct sunlight and that caused by cloudy-day light.

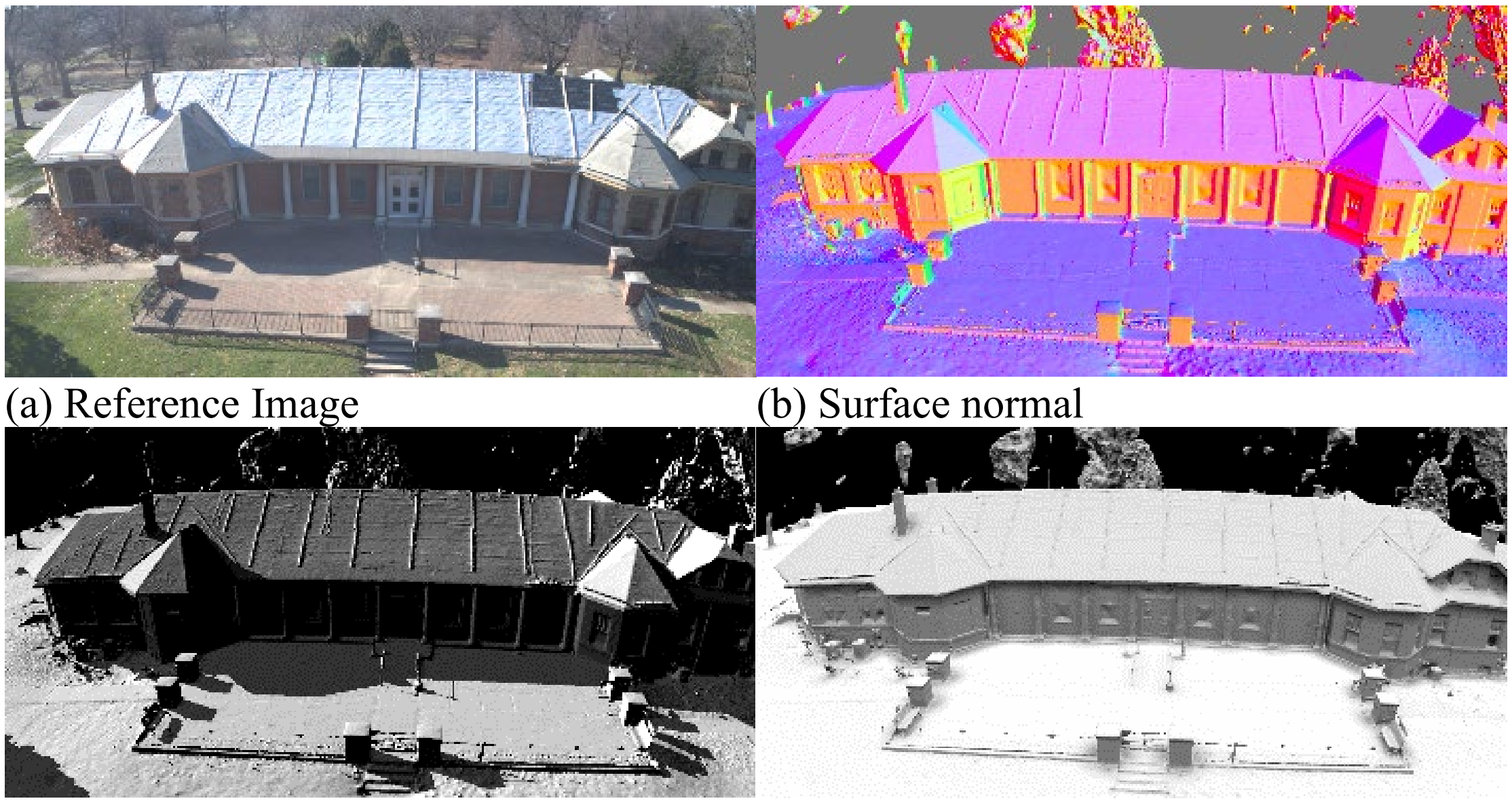

(a) Reference Image (b) Surface normal

(c) Sun shading term $S^{sun}$ (d) Sky shading term $S^{sky}$

Figure 4.5 Visualize shading components of sunlight and skylight.

***Resolving Sun-sky lighting ratio Φ with lit-shadow pairs***

As described in Section 4.4.4, connecting the intensity of sunlight and skylight requires pixel values of paired points sitting in shadow and non-shadow neighbors (following Equation (4.10)). To detect such paired points, we propose a filtering strategy to sample, and then filter pairs for building robust lit-shadow statistics. Firstly, given the sun visibility $V^{sun}$, we extract the boundaries of shadow regions (Figure 4.6a) and sample pairs in the shadow and non-shadow regions (Figure 4.6b). Secondly, we propose a set of criteria to filter out unwanted pairs that are potentially outliers, i.e., those that do not share similar albedo values. The criteria are listed in Algorithm 4.1 to form a criteria-based filter using both pixel intensity and geometry (depth continuity) as intuitive measures, for example, pixels with under or overexposure should not be considered, and pixels with large depth difference should not be considered (since they may not lie on the same surface). Figure 4.6b and Figure 4.6c show results before and after applying the filtering. To ensure these lit-shadow pairs provide a robust estimation of Φ, we calculate Φ for each of the filtered pairs and fit them to a Gaussian distribution. With a p-test, if the null hypothesis is not rejected (meaning the p-value is smaller than 0.05 to achieve 95% confidence level of fitting), we will compute the mean of the samples within the 95% data (adjusted mean) as the final ratio Φ. Figure 4.6d shows outliers identified as in the 5% tail of the distribution.

Algorithm 4.1 Criteria-based filters

1. Remove overexposure (>0.95) and underexposure (<0.05) pixels.
2. The angle between the surface normal of the pair < 5°.
3. The depth difference of the two points is less than < 0.1 meters.
4. $\psi^{sun}, \psi^{sky} \in (0.1, 10.0)$ to ensure numerical stability.

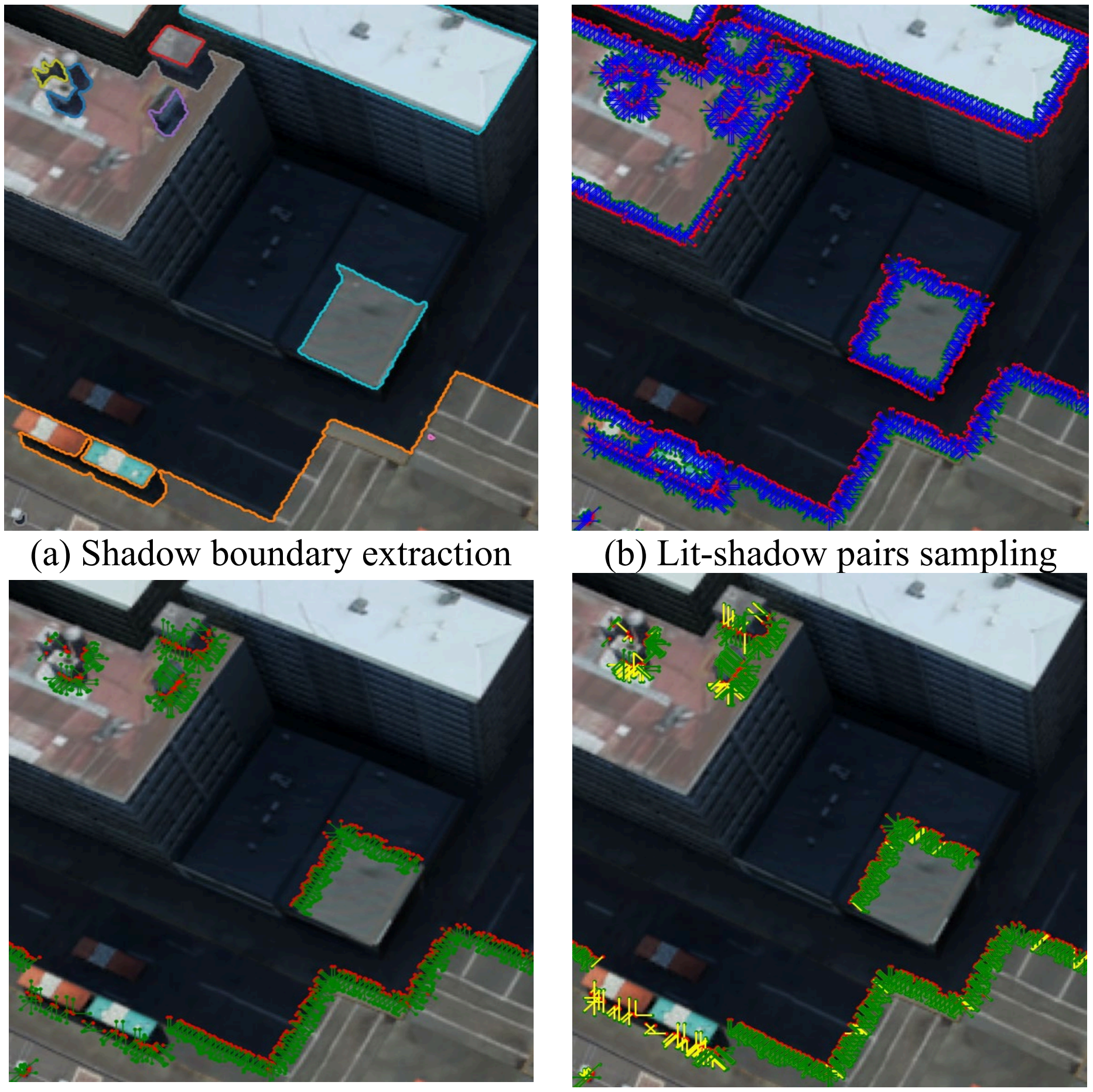

(a) Shadow boundary extraction (b) Lit-shadow pairs sampling

(c) Criteria-based filtering. (d) Statistical filtering (Green: inliers; Yellow: outliers).

Figure 4.6 Find reliable lit-shadow pairs to estimate Φ.

### 4.5.3 Sun Visibility Refinement to Cope with the Penumbra Effect

As mentioned at the beginning of Section 4.5, once the inlet light is known and the geometry is known, the albedo can easily be recovered by inverting the rendering equation by using $\rho = L_o/(\phi^{sun}S^{sun} + \phi^{sky}S^{sky})$. However, since we assume the sun visibility $V^{sun}$, i.e., the shadow, as a binary variable, does not match the actual penumbra effect reflecting the physics of the smooth transition between shadow and non-shadowed region due to sun disc scattering. If not refining the $V^{sun}$, the albedo recovery will produce artifacts at the shadow boundaries (as shown in Figure 4.7a). An analysis of the pixel intensity profile shown in Figure 4.7c indicates that the discontinuity of the visibility profile produced such an artifact. To address this, we aim to recover a continuous $V^{sun}$ using a total variation (TV) regularization, as shown in Equation (4.11):

$$
\begin{aligned}
\alpha^*(x) &= \arg\min_{\alpha(x)} \left|\left|\alpha(x)\text{-}\alpha_0(x)\right|\right|_P^2 + \left|\nabla \frac{1}{\rho(x)}\right| , \\
\frac{1}{\rho(x)} &= \frac{\psi^{sun} S^{sun}(x)}{L_o(x)} \alpha(x) + \frac{\psi^{sky} S^{sky}(x)}{L_o(x)} ,
\end{aligned}
\qquad (4.11)
$$

where $\alpha$ is the sun visibility along the profile, $x$ is the distance on the profile, $\alpha_0$ is the initial binary sun visibility from the directional sun model, $||\cdot||_{\boldsymbol{P}}$ is Mahalanobis distance with weight matrix $\boldsymbol{P}$**,** $\nabla$ is Gradient operator. We choose to minimize $1/\rho$ since compared with directly optimizing regarding $\rho$, the $1/\rho$ yields a closed-form solution due to its linearity with $\alpha(x)$. Weight matrix $P$ is a diagonal positive definite matrix. This formulation adjusts the $\alpha(x)$ that close to the shadow boundary, and $L_o(x)$ in the formulation inherently uses the image information to guide the Sun visibility refinement. By optimizing based on Equation (4.11), the sun visibility becomes a continuous variable and the produced artifacts can be successfully removed (Figure 4.7b and Figure 4.7d).

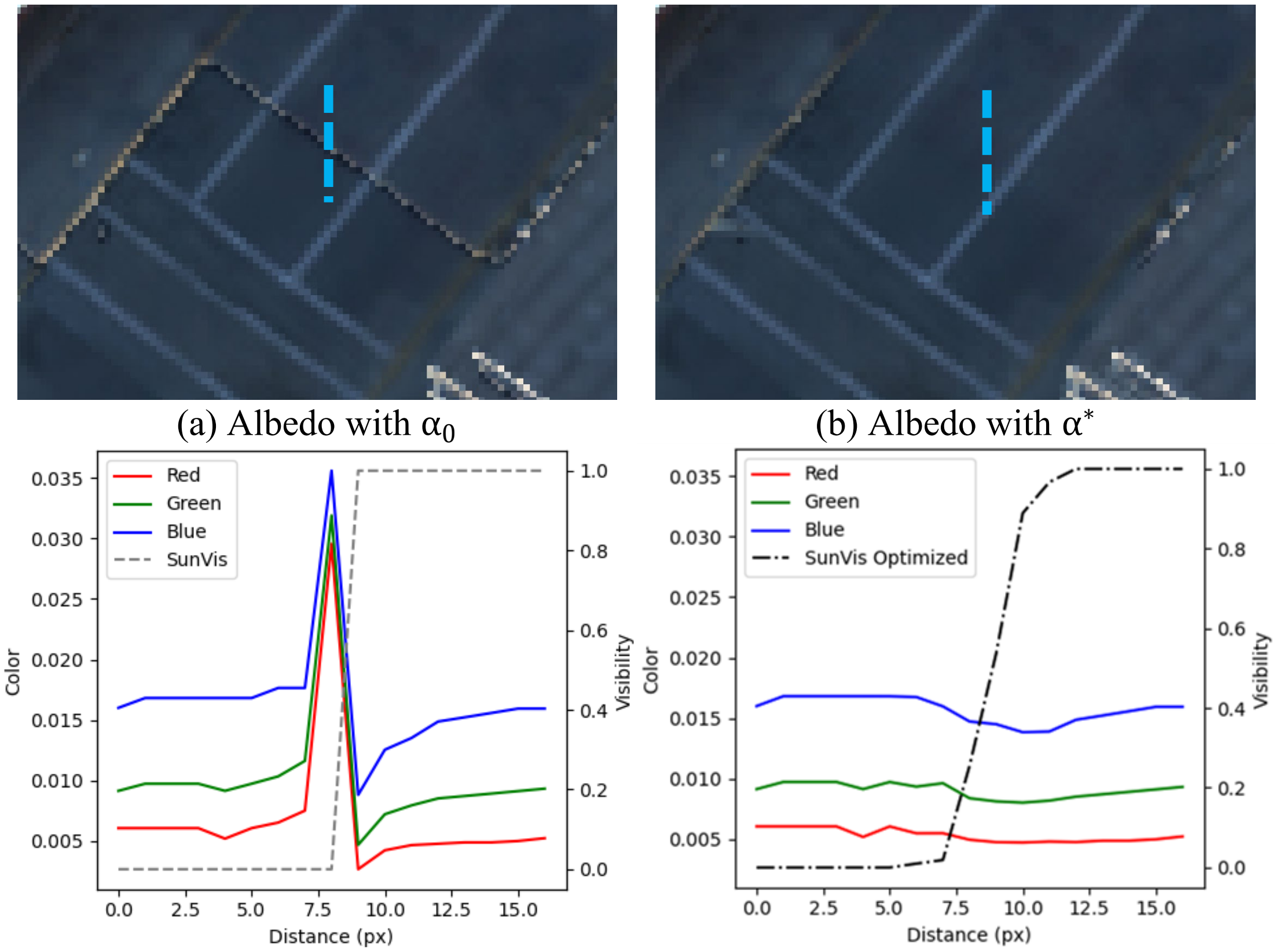

(a) Albedo with $\alpha_0$ (b) Albedo with $\alpha^*$

(c) Profiles of (a) (d) Profiles of (b)

Figure 4.7 (a) and (c) are the recovered albedo with binary $\alpha_0$ showing artifacts in both the figure and the profile; (b) and (d) show the effectiveness of the recovered albedo using our Sun visibility refinement.

Combining $\alpha^*$ and $\Phi$ into the lighting formulation, the inverted albedo is shown in Figure 4.8a and Figure 4.8b. Figure 4.8c shows an example of a simulated frame, where our recovered shadowing and the ground-truth shading are very close to each other, proving the effectiveness of our method.

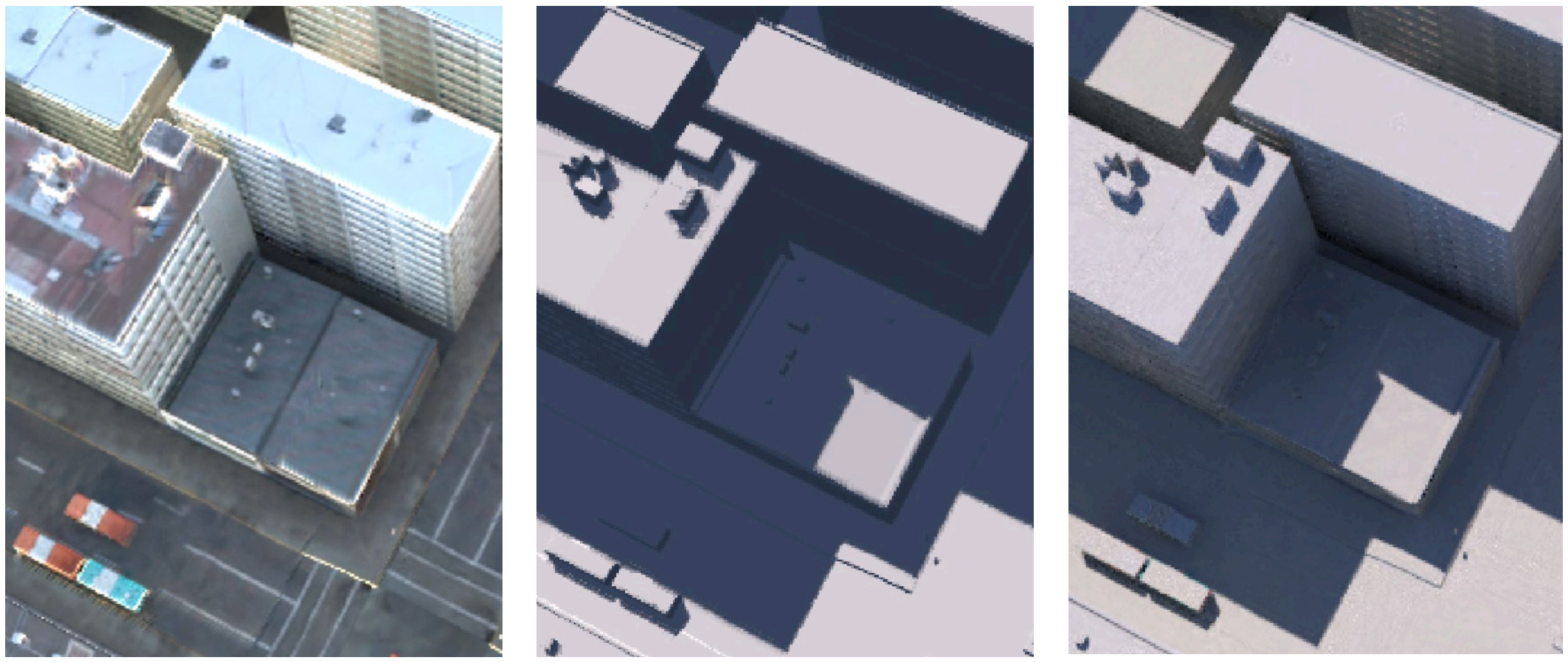

(a) Our estimated albedo image (b) Our estimated shading image (c) Ground truth shading image

Figure 4.8 Our albedo and shading estimation.

## 4.6 Evaluation

In this section, we conduct albedo recovery experiments using both synthetic datasets and real-world datasets to evaluate our method. Since there exists no ground-truth albedo for real-world outdoor images, we only qualitatively and indirectly evaluate the results. Quantitative results are concluded using simulated results through physics-based rendering (PBR). We also compare our results with those of the state-of-the-art methods including InverseRenderNet (IRN) (Yu and Smith 2019), Shadow Matting Generative Adversarial Network (SMGAN) (Cun, Pun, and Shi 2020), InverseRenderNetv2 (IRNv2) (Yu and Smith 2021), and our 2022 version (S. Song and Qin 2022). A comprehensive evaluation protocol for real-world images is adopted to evaluate: 1) if the recovered albedo is free from cast shadow and shading effects (Section 4.6.3); 2) if the recovered albedo shows consistency for images of the same outdoor scene collected at different time of day (Section 4.6.4).

### 4.6.1 Dataset for Experiments

There is no publicly available dataset that provides outdoor aerial views as long as the corresponding ground-truth albedo. Thus, in our experiment, we created a synthetic dataset for the evaluation for the quantitative evaluation. To demonstrate the effectiveness of our method with real-world data, we also collected aerial images with a drone under different lighting conditions. The details of each type of dataset are presented in the following content.

***Synthetic dataset***

To create a synthetic city model to simulate aerial photogrammetry, we generate a synthetic 3D city model using ESRI CityEngine, a procedural modeling software (ESRI 2020). The model textures come from a pre-existing asset library without any baked shadows or shading. Blender Cycles (Blender Online Community 2021) is a ray-tracing render engine that allows us to generate photorealistic samples with ground truth albedo and lighting components. We use built-in Nishita sunlight and skylight models (Nishita et al. 1993) to simulate the natural lighting environment. The synthetic dataset includes 30 virtual camera images, ground truth albedo images, camera orientation parameters, and sun positions. All images use linear RGB color space and are stored in OpenEXR (Academy Software Foundation 2023) format for the wide dynamic range. The average ground resolution is about 20 cm/pixel.

***Real-world aerial dataset***

Capturing ground-truth albedo for all regions in a scene in the wild is extremely challenging (Wu et al. 2023). Thus, we collect drone photogrammetric blocks over three days of time focusing on the same scene. The goal is to apply our albedo recovery method to those individual images and evaluate cross-view consistency under different lighting. We captured aerial images using a DJI FC6310S camera with an 8.8 mm f/2.8 lens. The camera was carried by a DJI Phantom Pro 4 v2.0. Flight height is 70 meters above ground, and average ground resolution is 2.5 cm/pixel. The region of data collection is around a squared area of 200 × 200m. We had 6 flights on 3 consequential days to collect images under various illumination conditions (two time point per day), as listed in Table 4.1. All flights are taken on the same site but on different time of the day. Day-1 data includes nadir images Day-2 and Day-3 data include oblique images as well.

| | | |
|---|---|---|
| Day 1 | 08:00 AM – 09:00 AM \| 29 images | 10:00 AM – 11:00 AM \| 31 images |
| Day 2 | 09:00 AM – 10:00 AM \| 52 images | 06:00 PM – 07: 00 PM \| 53 images |
| Day 3 | 09:00 AM – 11:00 AM \| 68 images | 11:00 AM – 12:00 PM \| 57 images |

Table 4.1 Data acquisition time and number of images of each flight.

We perform photogrammetric 3D reconstruction of the data on Day 3, since all images are taken in the morning and show reflect better geometric reconstruction. Then, we registered Day 1 and Day 2 images with virtual Ground Control Points (GCP) manually collected from the model of Day 3, with registration errors smaller than 0.01m. As shown in Figure 4.9, we generated an orthorectified imagery of each day using a common geometry reconstructed from Day 3 images.

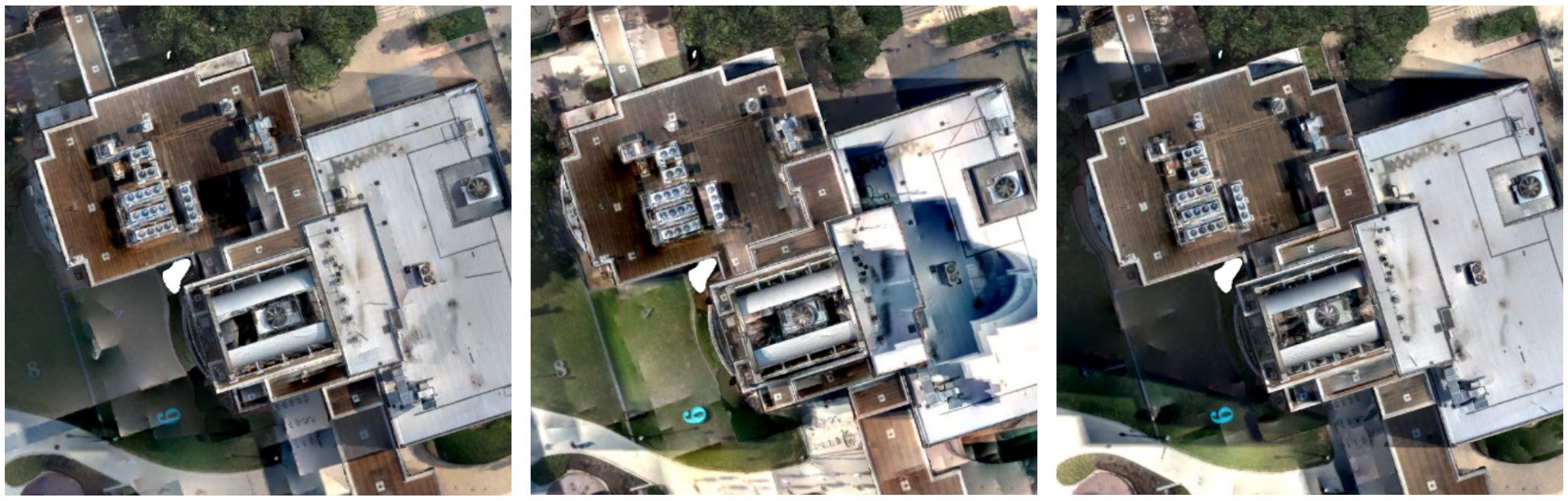

(a) Textured with Day 1 images. (b) Textured with Day 2 images. (c) Textured with Day 3 images.

Figure 4.9 Orthorectified imagery using our multitemporal dataset.

### 4.6.2 Quantitative Evaluation with the Synthetic Dataset

To separately evaluate the performance of our method in chromatic and brightness, we decompose the albedo image into chromaticity and brightness terms as defined in Equation (4.12). An example decomposition from our synthetic dataset is shown in Figure 2.12.

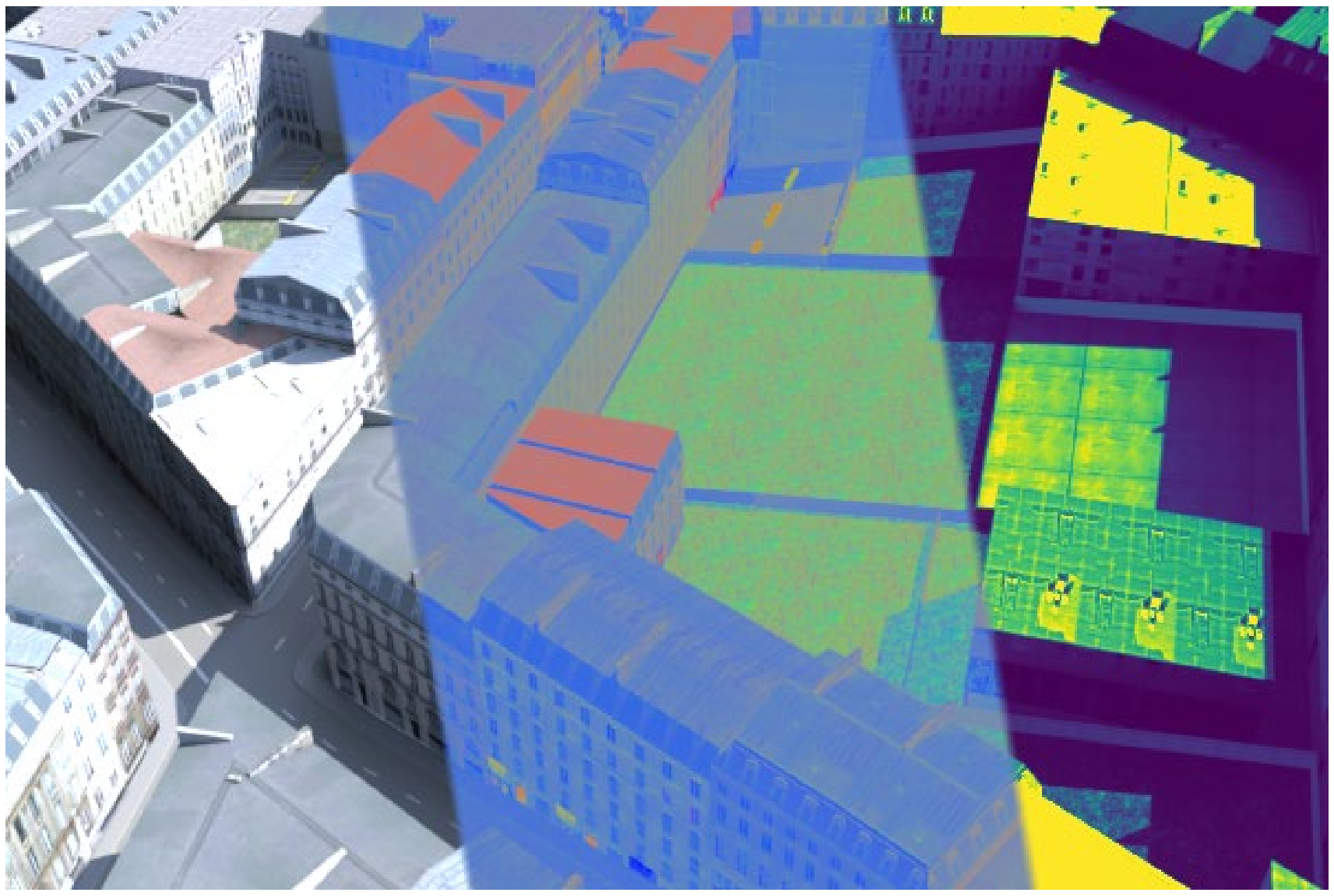

Figure 4.10 Visualization of decomposition of albedo image (left) to chromaticity (middle) and brightness (right).

$$\begin{cases} \text{Chromaticity} \coloneqq \dfrac{\{R, G, B\}}{R + G + B} \\ \text{Brightness} \coloneqq \dfrac{1}{3}(R + G + B) \end{cases} \tag{4.12}$$

We report the PSNR (Peak signal-to-noise ratio), SSIM (Structural Similarity Index), and MAE (Mean Absolute Error) metrics of the result against the ground-truth albedo image in Table 4.2. Visualization of the resulting albedo images is provided in Figure 4.11. Since we explicitly model the outdoor illumination, our result is significantly better than the comparing approaches. Compared with our earlier work (S. Song and Qin 2022), our method shows a slight improvement which can be attributed to the skylight visibility (Section 4.4.3). The baseline in Table 4.2 evaluates the rendered image to the albedo. Among the comparing approaches, only SMGAN shows positive recovery of albedo (better than the baseline), while the IRNv2 is partially better (for PSNR and MAE). We also observe that the PSNR of our method is not among the best, showing that the intensity of the light model should be better calibrated.

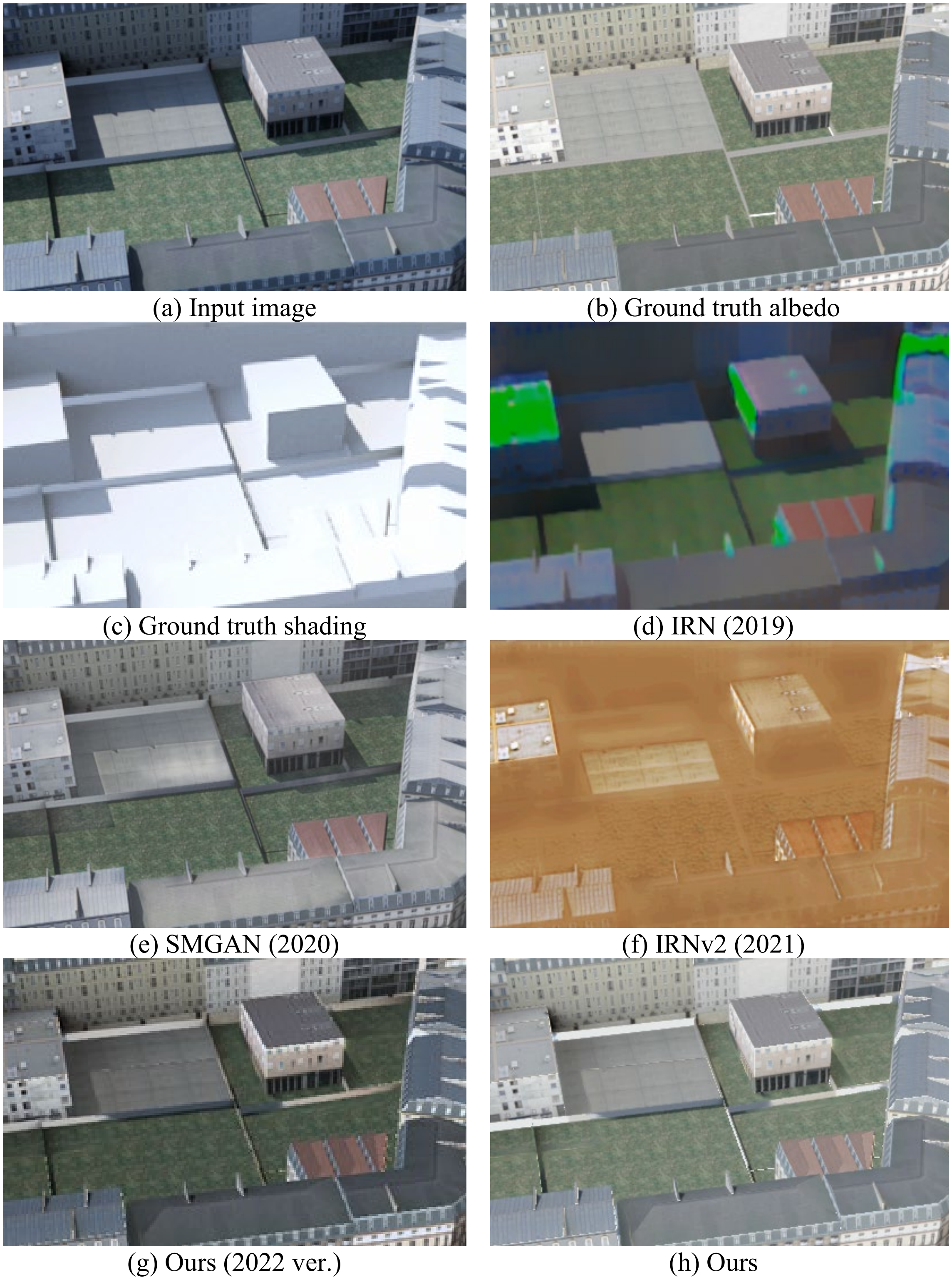

(a) Input image (b) Ground truth albedo

(c) Ground truth shading (d) IRN (2019)

(e) SMGAN (2020) (f) IRNv2 (2021)

(g) Ours (2022 ver.) (h) Ours

Figure 4.11 Albedo decomposition comparison on the synthetic dataset. Images are enhanced by contrast and brightness correction for better visualization.

| Method | Chromaticity | | | Brightness | | |
|---|---|---|---|---|---|---|
| | PSNR ↑ | SSIM ↑ | MAE ↓ | PSNR ↑ | SSIM ↑ | MAE ↓ |
| **Baseline** | <u>22.623</u> | <u>0.943</u> | <u>0.061</u> | <u>15.005</u> | <u>0.631</u> | <u>0.166</u> |
| **IRN (2019)** | 17.669 | 0.807 | 0.087 | 15.821 | 0.550 | 0.123 |
| **SMGAN (2020)** | 27.072 | 0.949 | 0.034 | 16.018 | 0.678 | 0.130 |
| **IRNv2 (2021)** | 24.439 | 0.910 | 0.047 | **17.188** | 0.661 | 0.094 |
| **Ours (2022 ver.)** | 31.97 | **0.978** | 0.020 | 15.73 | 0.777 | 0.091 |
| **Ours** | **32.207** | **0.978** | **0.018** | 15.891 | **0.784** | **0.086** |

Table 4.2 Comparison of albedo decomposition results with ground truth. The Baseline represents the input image without any process.

### 4.6.3 Qualitative Evaluation with Real-world Dataset

We evaluate the performance of our albedo decomposition on real-world aerial datasets by comparing it with existing albedo recovery or related algorithms. Figure 4.12 shows the visual quality of the recovered albedo: Among the results of different approaches, IRN tended to recover an over-smooth albedo image, while IRNv2 preserved more details, but the albedo image was significantly distorted (both color and geometry). The typical qualitative evaluation focuses on the rectangle region in Figure 4.12b and Figure 4.12d, where it is expected that shadows, and over-bright concrete/pedestrian ways (due to direct sunlight) should be corrected. We observe that our method produces the best performance in shadow removal, correcting the over-brightness of concretes. Among the other comparing methods, SMGAN obtained notable results, while it produced artifacts at the rightmost rectangle.

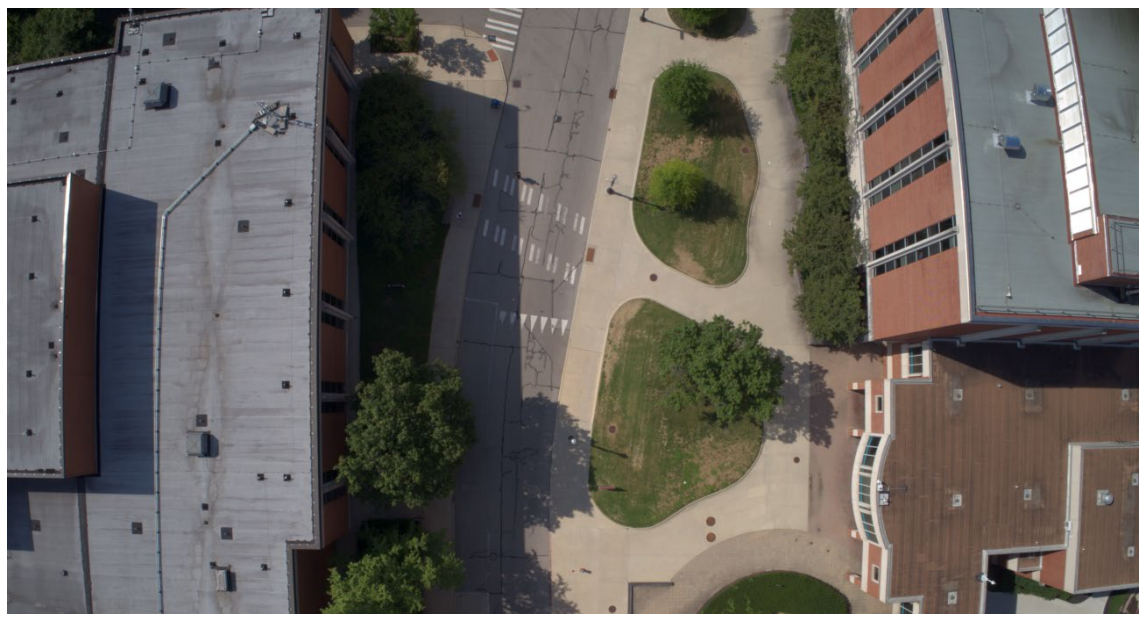

(a) Input

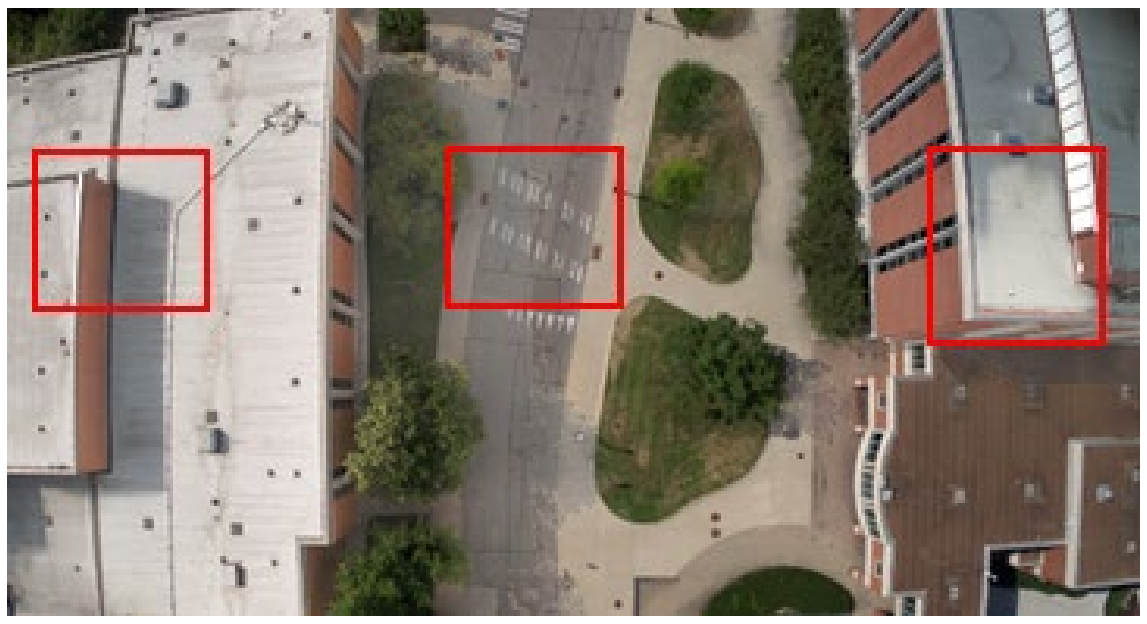

(b) SMGAN (2020)

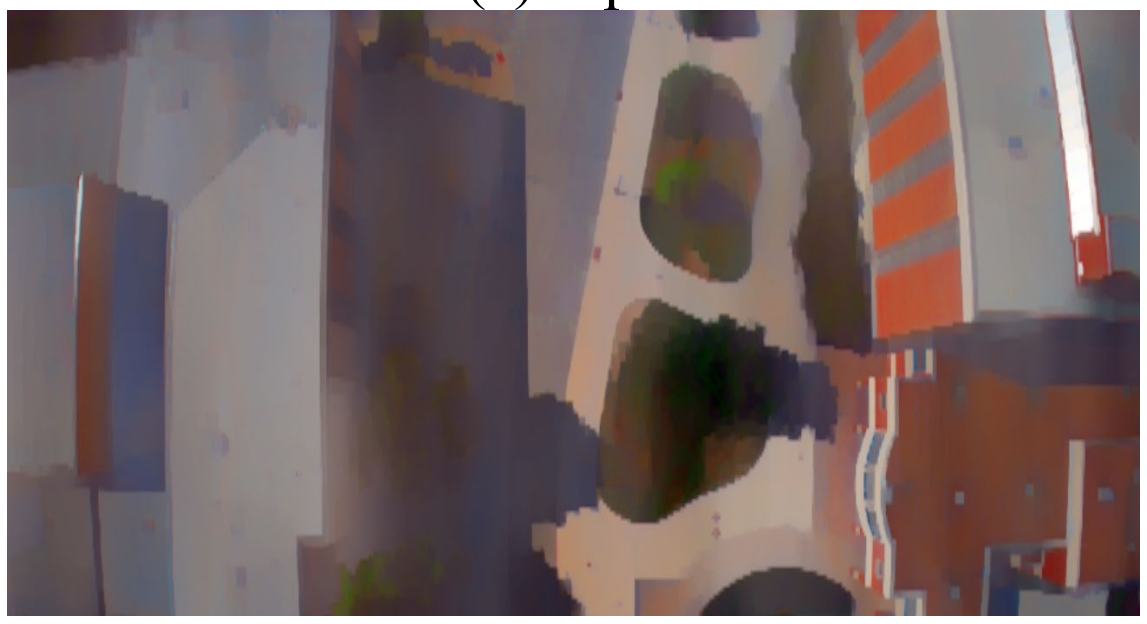

(c) IRN (2019)

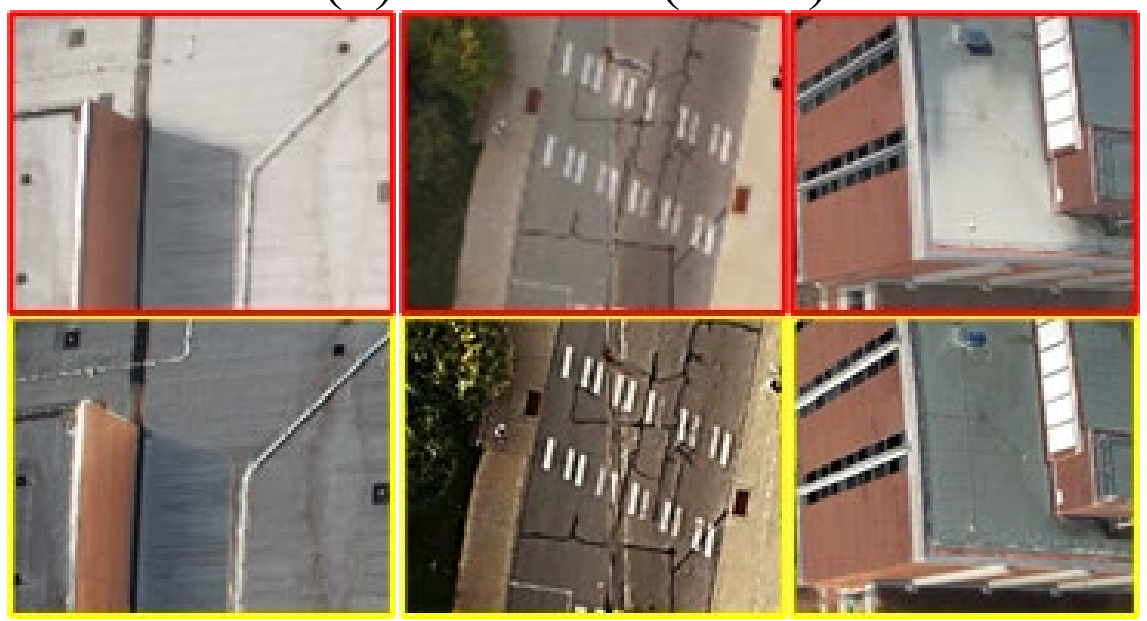

(d) Details of (b) and (f).

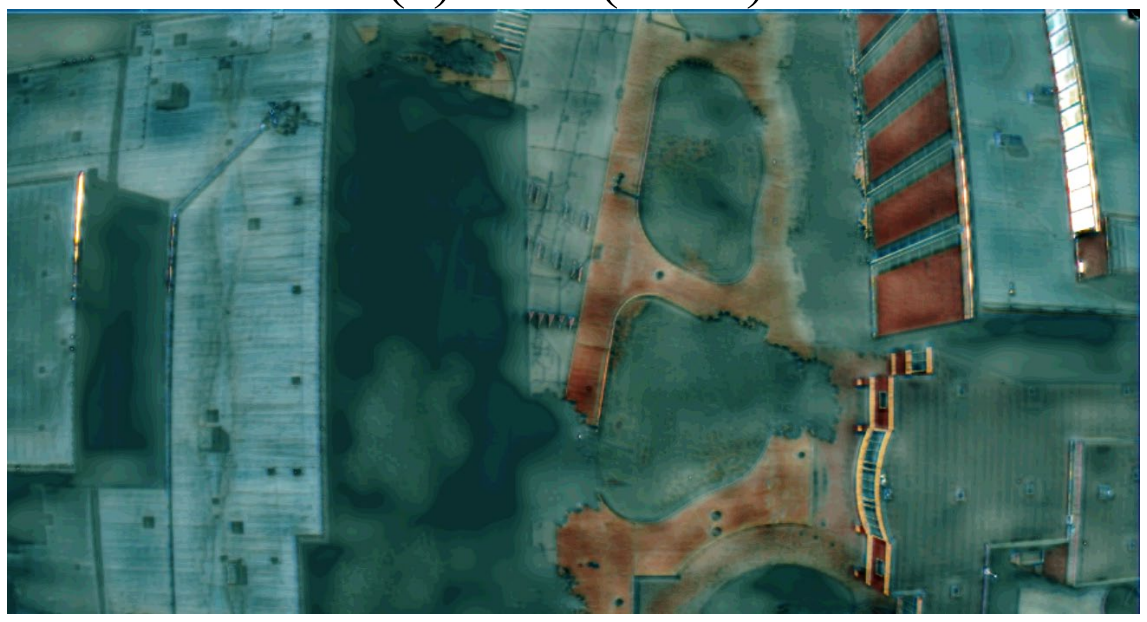

(e) IRNv2 (2020)

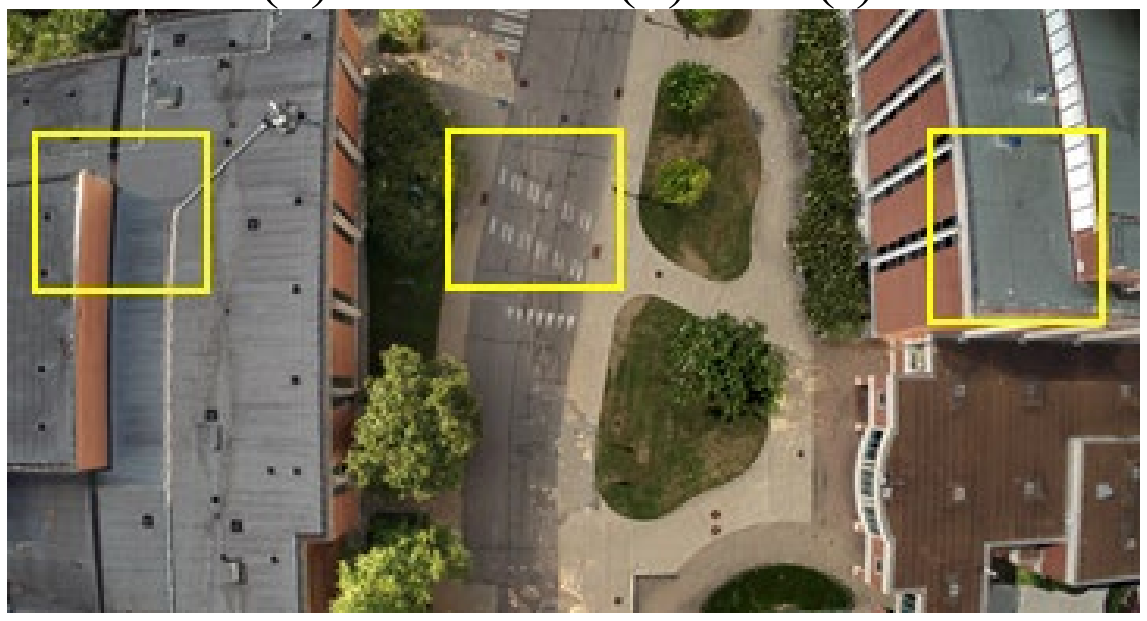

(f) Ours

Figure 4.12 Albedo decomposition comparison on a real-world dataset. Images are enhanced by contrast and brightness correction for better visualization.

### 4.6.4 Multi-temporal Consistency with Real-world Datasets Evaluation

We evaluate the recovered albedo of datasets of the same scene but collected on different days (Section 4.6.1). The goal is to apply our albedo recovery method to these individual images taken at different time and evaluate their consistency. Ideally the recovered albedo will appear the same, albeit their originally images are significantly.

Figure 4.13 shows an example of our result: Figure 4.13(a-b) shows two original images taken at different local time, where drastically different illuminations can be found, with our albedo correction method, as shown in Figure 4.12(c-d), it shows much less shading effect and higher consistency.

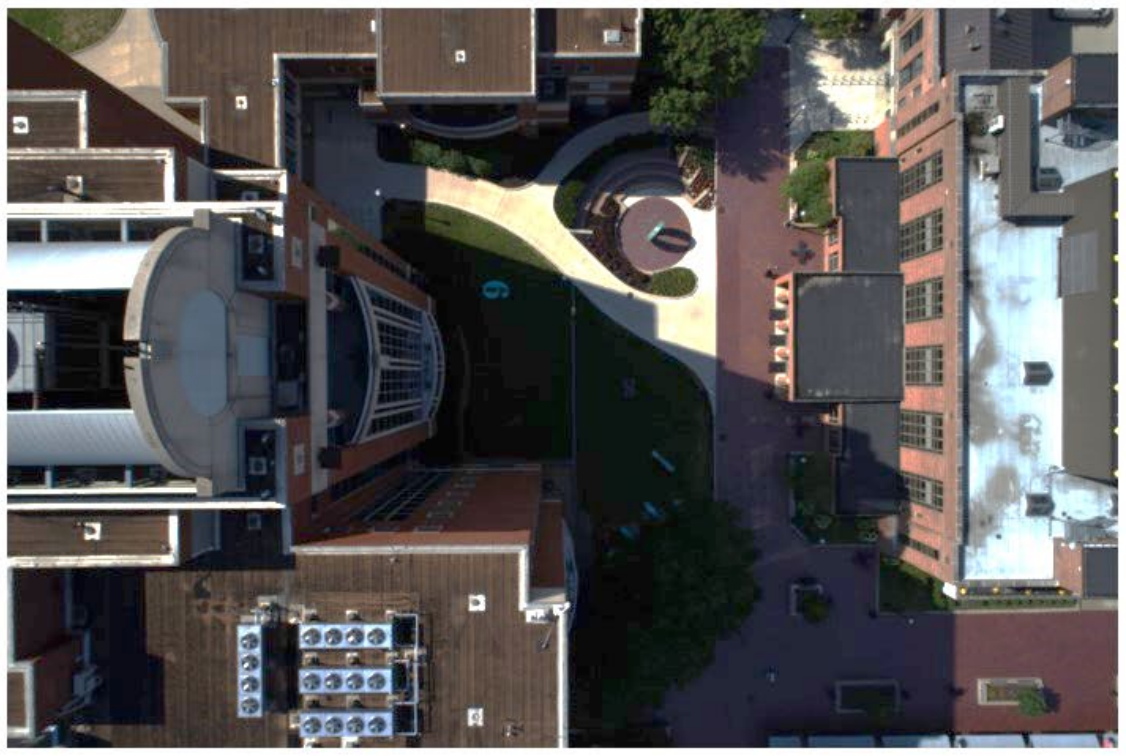

(a) Original image taken at 10 AM local time

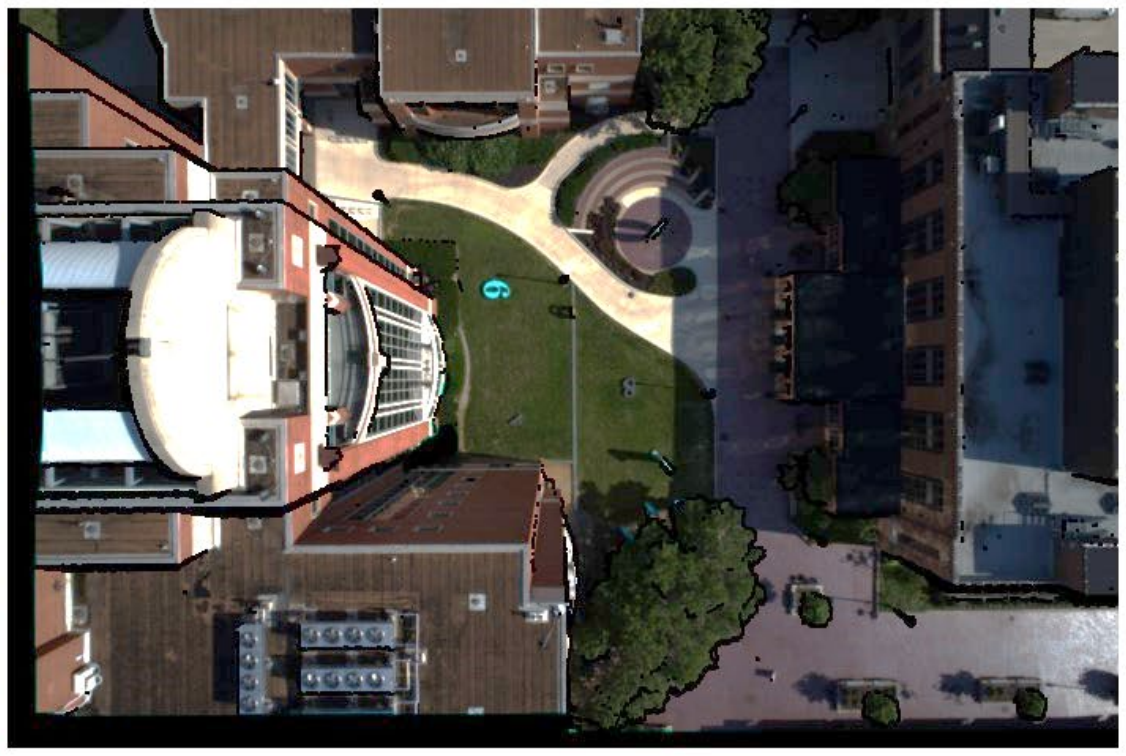

(b) Original image taken at 6 PM local time

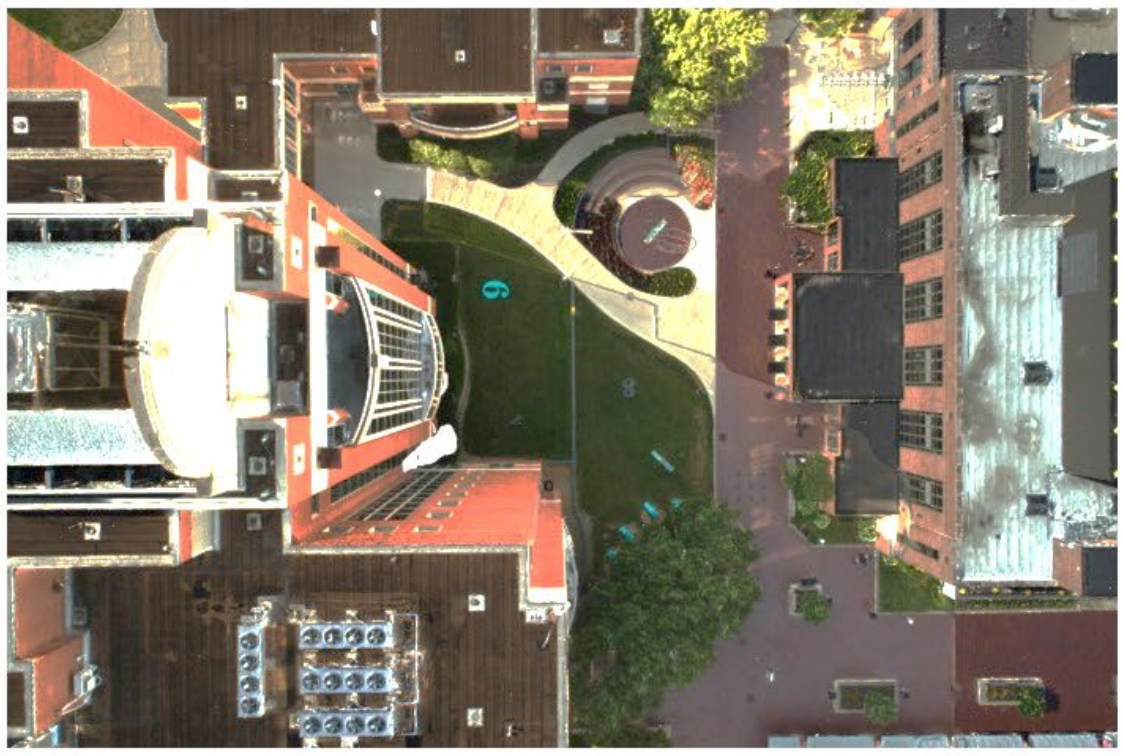

(c) Our albedo recovered image of (a)

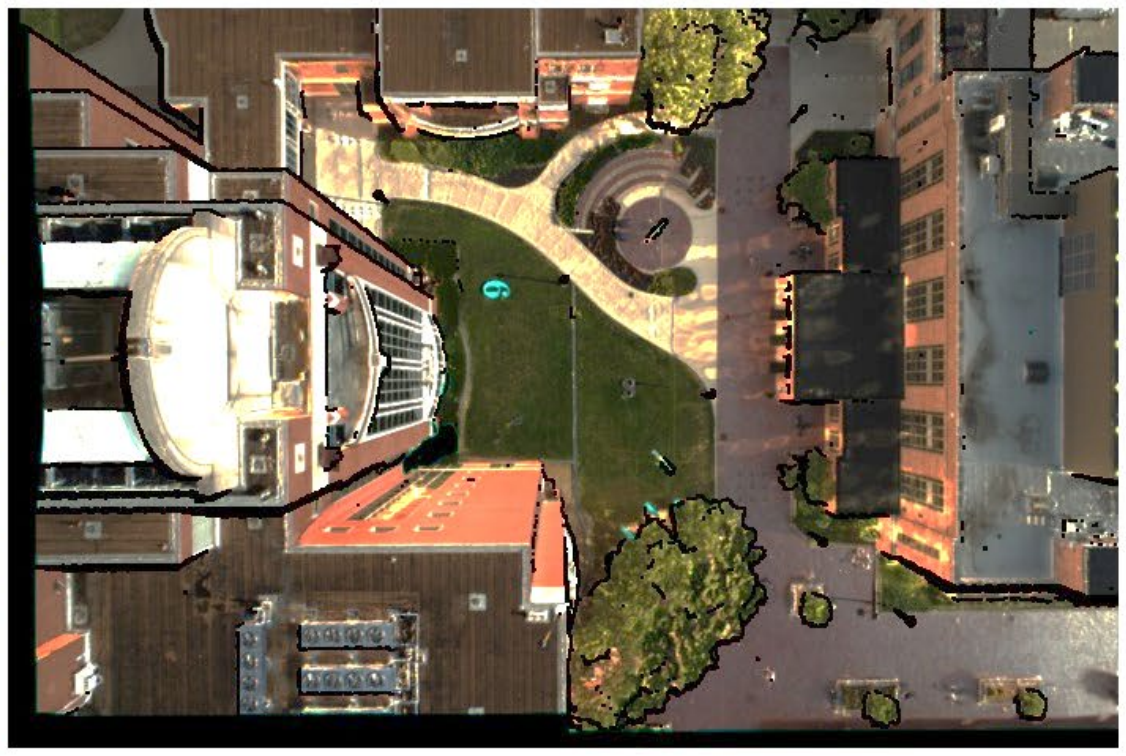

(d) Our albedo recovered image of (b)

Figure 4.13 Evaluate the temporal consistency of our albedo decomposition (c)-(d) compared with original images (a)-(b). All images are geometrically corrected to the same viewpoint. Black regions in (b) and (d) are due to occlusions from view correction.

To quantitatively evaluate this, we computed the standard deviation of images pixels through different time, with and without albedo recovery (i.e., original and albedo recovered image). To facilitate pixel-wise computation, images of different dates and time are geometrically corrected to have the same viewpoint (using the photogrammetric mesh). As expected, the temporal consistency of our albedo decomposition significantly excels the consistency of the original images. Table 4.3 shows that with a scale of 8-bit grey scale, the albedo images of different days and times are 32% more consistent than the original images. It should be noted that the statistics in Table 4.3 are derived from all 6 flights collected across 3 days (as shown in Table 4.1).

| | **Original** | **Albedo recovered image** | **Std drop (improved consistency)** |
|---|---|---|---|
| **Average** | 23.69 | 15.8 | -32% |

Table 4.3 Average standard deviation of temporal images (unitless, pixel value is scaled in 0-255).

### 4.7 Applications

To demonstrate the possibility of utilizing our approach in fields of research and industry, we present four applications benefiting from our albedo recovery method: 1) Model relighting: the model textured with albedo images is more realistic in simulation system relighting; 2) Feature point extraction: our albedo recovered image yield more feature point matches (sparse features) for photogrammetric processing; 3) Dense matching: our albedo recovered image produce more complicate dense matching results in stereo and multi-view reconstruction. 4) Change detection: our albedo recovered image drastically improves the change detection application at ultra-high-resolution images.

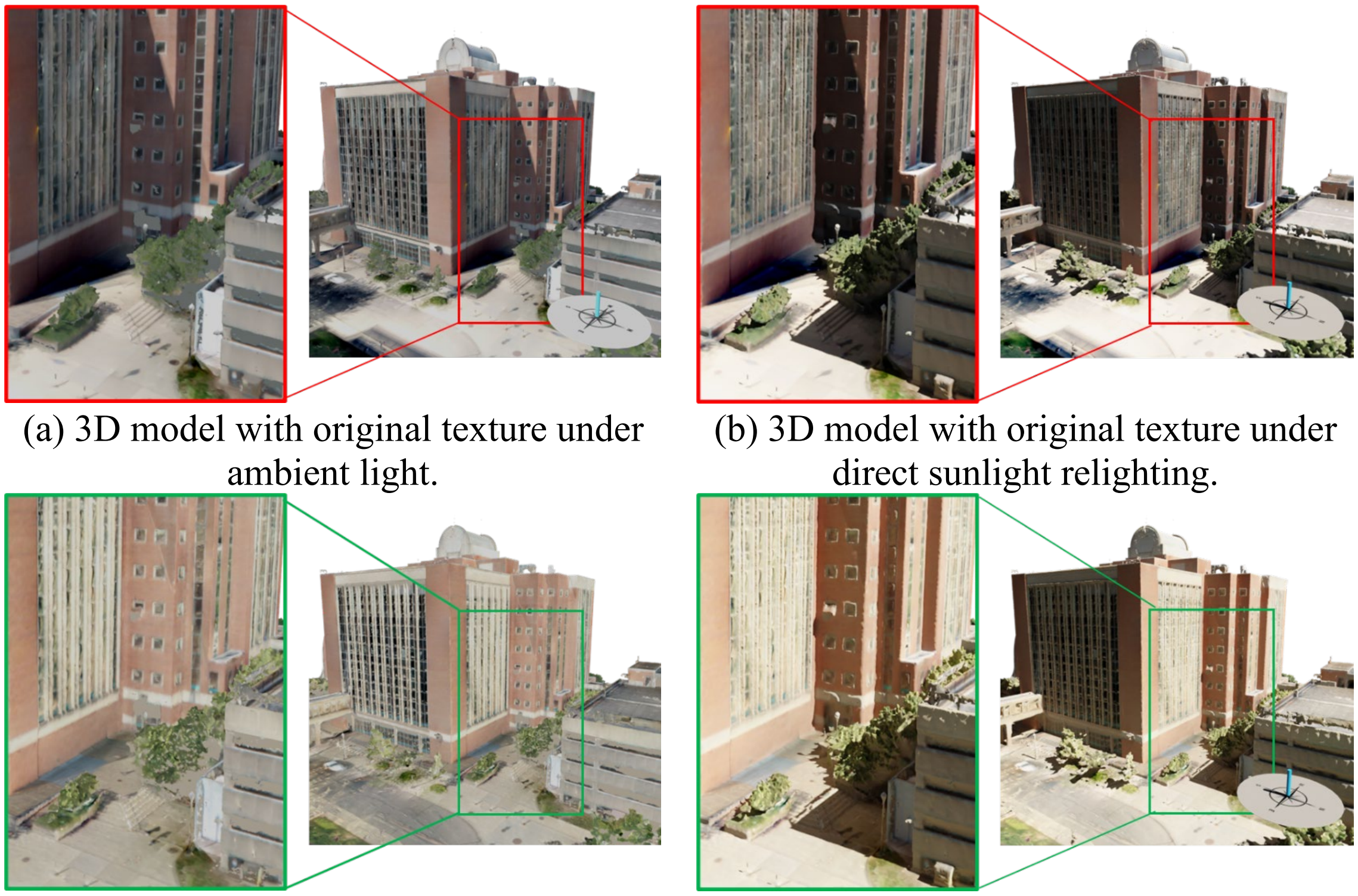

(a) 3D model with original texture under ambient light.

(b) 3D model with original texture under direct sunlight relighting.

(c) 3D model with our albedo recovered texture under ambient light.

(d) 3D model with our albedo recovered texture under direct sunlight relighting.

Figure 4.14 Rendered textured model from novel view with ambient lighting (a, c) and simulated sun-sky lighting (b, d). The compass and cast shadow on the right corner indicate the sun position in image capturing or rendering.

### 4.7.1 Application: Textured Model Relighting

Model relighting is a standard application of 3D textured models in a simulation system, in which views are rendered under different simulated lighting, and thus the realism of the rendered view is the key. In Figure 4.14, we show that, with our recovered image, the rendered views contain much fewer shading artifacts. Figure 4.14a and Figure 4.14c are rendered with ambient lighting (area and homogenous lighting) and Figure 4.14b and Figure 4.14d are rendered with a different sun position than that at the collection time. Ideally, under ambient lighting, no shadow or shading effect should be observed. With sunlight, the shadow azimuth and shading should be coherent with physical law. As can be seen in Figure 4.14, the rendered view using the model texture of the original images contain unwanted shadow under the ambient light and double shadows under the sunlight. Our model shows a much better rendered view since there are no shading artifacts, and colors are consistent for materials that are supposed to be consistent (e.g., the paved road materials).

### 4.7.2 Application: Feature Extraction & Matching, and Edge Detection

In feature matching, ideally, more consistent images yield more feature matches, because the photo-consistency of difference images is a key factor of concern that impact the performance of many feature extractors and matchers (Braeger and Foroosh 2021; Valgren and Lilienthal 2010). Therefore, we compare the performance of feature point matching on a pair of images before and after our albedo correction. Figure 4.15 shows the performance of a classic feature matching, SIFT (Lowe 2004), as it is still the most widely used feature matcher in photogrammetry software packages. We show that the performance of the feature matcher, when applied on our albedo recovered image, significantly outperforms its results on the original image, in terms of the distribution of matches, as well as the number of inliers.

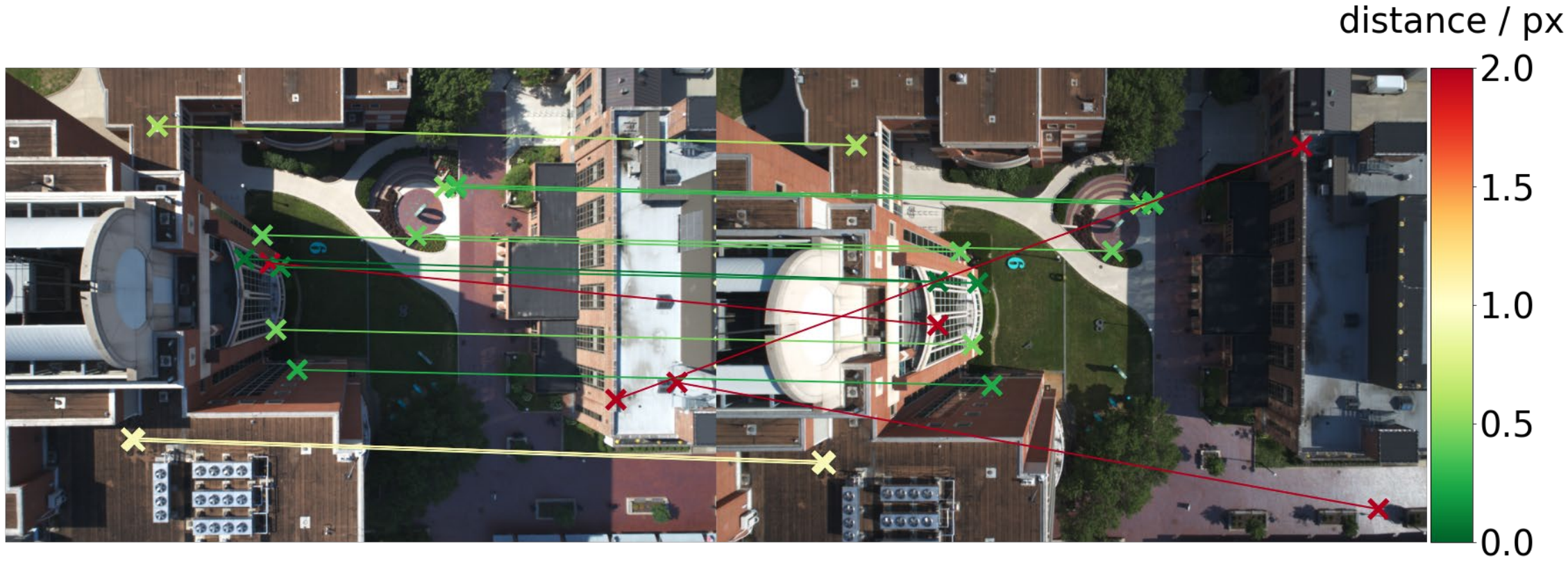

(a) Original image pair
(Number of candidates: 4929, number of Inliers with RANSAC: 17, mean distance to epipolar lines: 35.42 pixels)

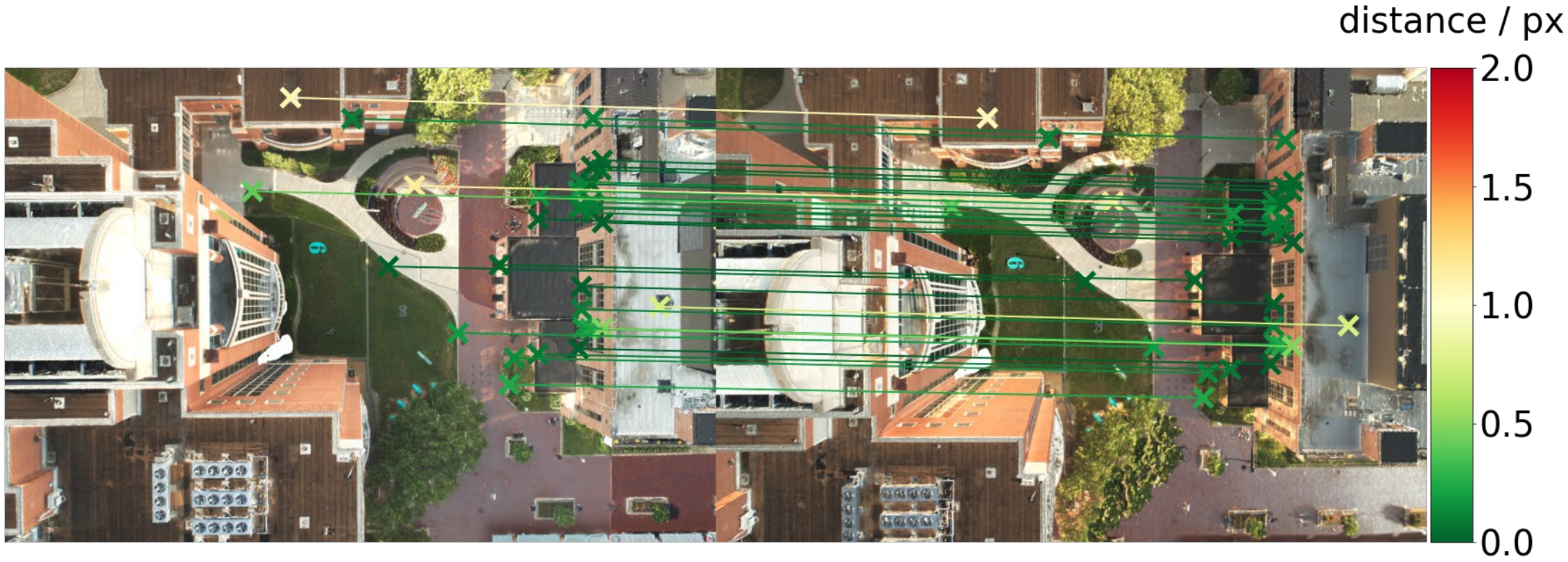

(b) Our albedo image pair
(Number of candidates: 8328, number of Inliers with RANSAC: 31, mean distance to epipolar lines: 0.21 pixels)

Figure 4.15 SIFT matching across different times of the day. Lines and points are colored by distance to corresponding epipolar lines (y-parallax) where the Fundamental matrix is computed from camera poses from the structure-from-motion with all images.

Edge and line extraction is an important task that serves for feature matching and reconstruction, which is sensitive to shading and shadows as well. Thus, we compare the performance of Canny edge detection (Canny 1986) and Line Segment Detector (Grompone von Gioi et al. 2012) on the original and our albedo decomposition image. As shown in Figure 4.16, both canny edges and line segments are less affected by cast shadow, and there are more features under the shadowed region.

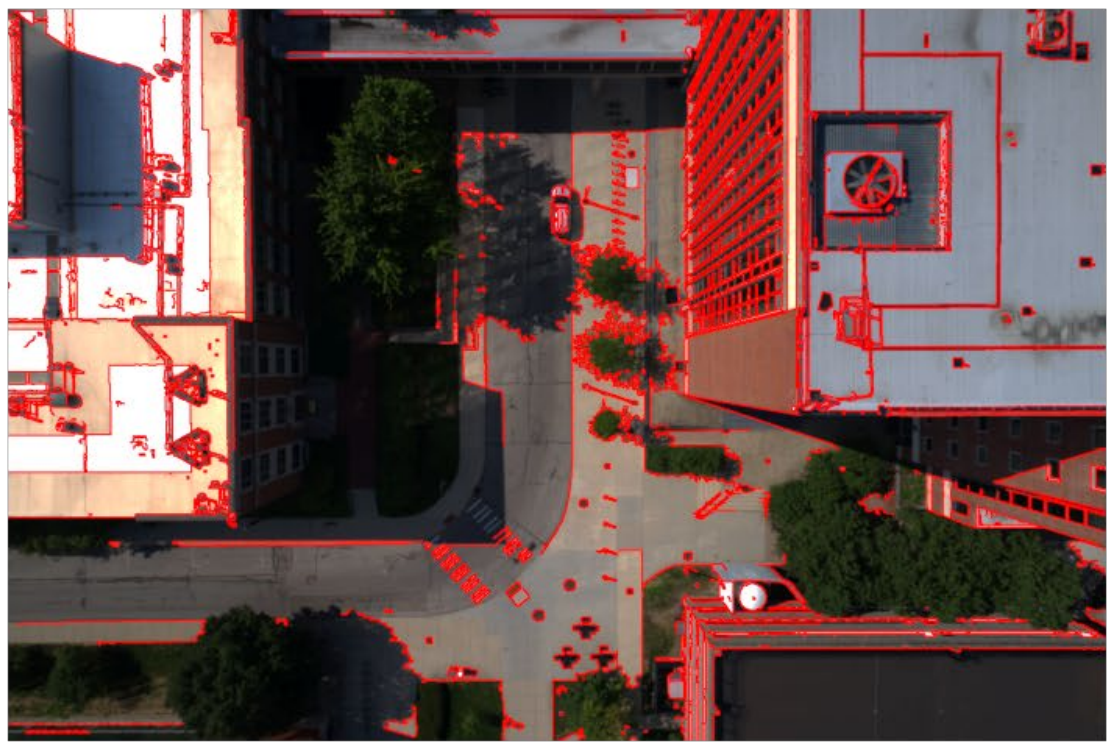

(a) Canny edge of the original image

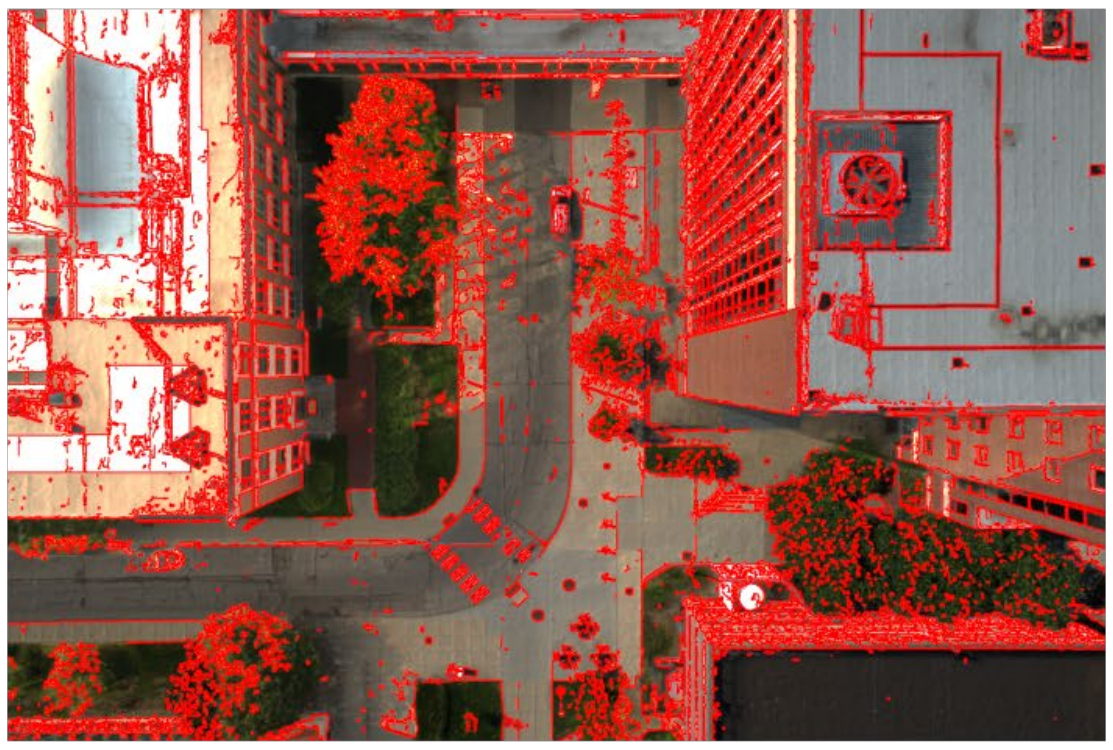

(b) Canny edge of our albedo decomposition

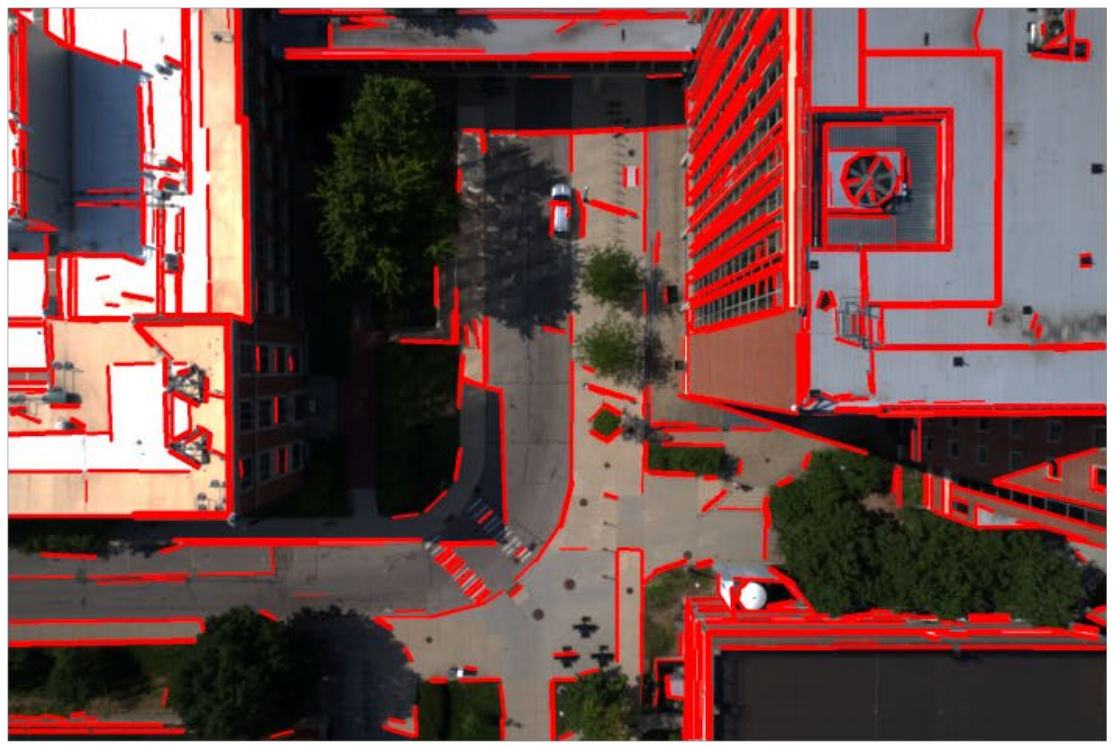

(c) LSD segments of the original image

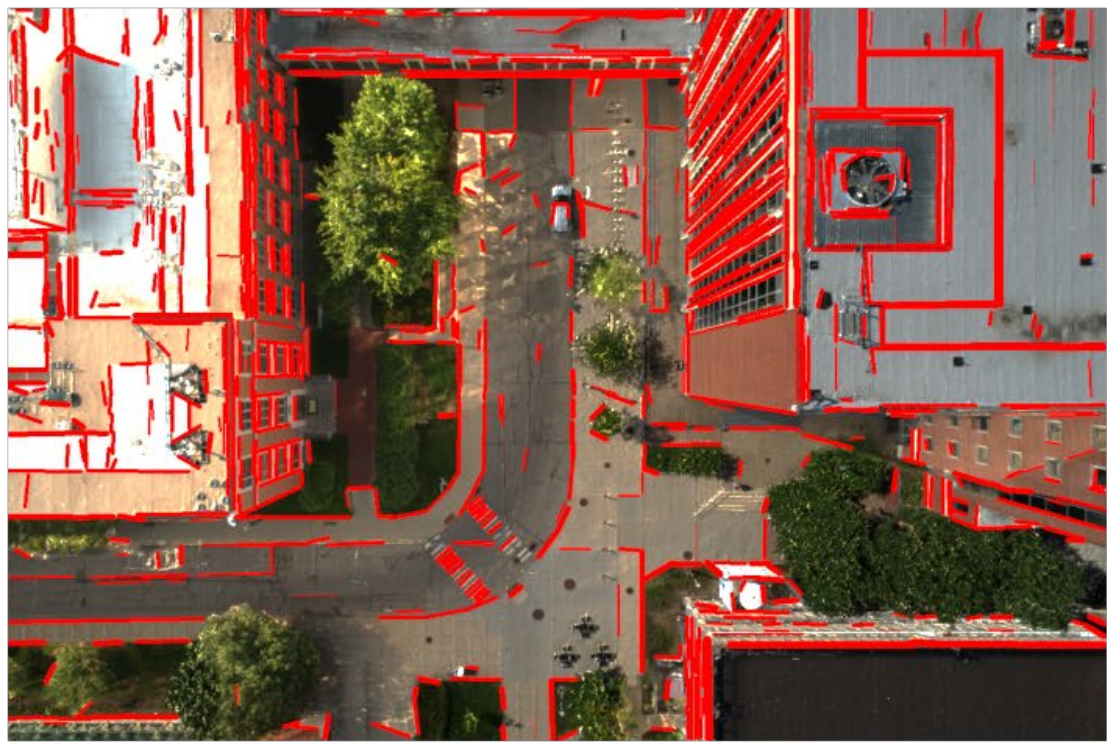

(d) LSD segments of our albedo decomposition

Figure 4.16 Edge and line segment extraction using the original image and our albedo decomposition.

### 4.7.3 Application: Stereo and Multi-view Dense Image Matching

Similarly, more consistent images may yield better dense image matching results. Therefore, our proposed albedo recovery method may augment the dense matching results. Here, by using images before and after our albedo correction method (Figure 4.17a and Figure 4.17c), we perform a multi-view stereo dense image matching using

Patch-Match algorithm (Barnes et al. 2009) followed by a multi-view depth map fusion, both are implemented by OpenMVS (Cernea 2020), a very commonly used open-source software package. The resulting point clouds are shown in Figure 4.17. To verify that the effectiveness of our method is more than just a brightness balancing in the shadowed region. We have also performed a set of experiment by performing a gamma correction (A. R. Smith 1995) prior to dense matching, results shown in Figure 4.17b. It is obvious that our albedo recovered image set yields more details and higher completeness overall.

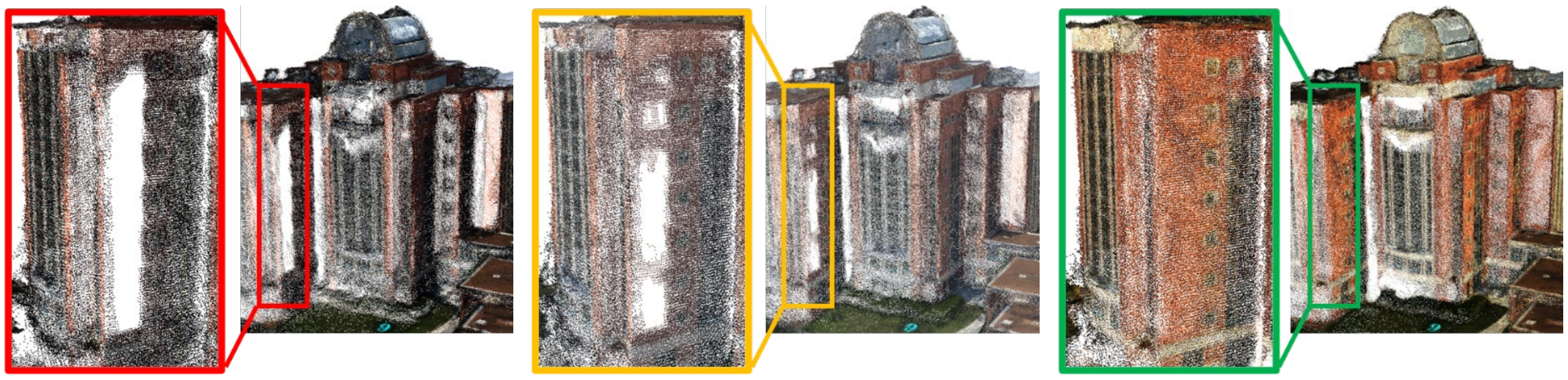

(a) All original images (b) Original images with gamma correction (c) Our albedo images

Figure 4.17 Comparison of multi-view stereo point clouds generated from all images (290 images) of the region. For gamma correction (b), we scale the colors by a power of 1/2.2. Colored boxes indicate the region where our albedo point cloud presents better completeness.

### 4.7.4 Application: Change Detection

Shading effects are one of the major challenges in image-based change detection, which added false positives to change detection algorithms. For example, shadows may be misinterpreted as changes or actual changes may be obscured under the shadowed. Thus, if the shading effects can be reduced, changes could be easily detected by comparing overlapped pixels. Here we use a simple image differencing based change detection algorithm (Algorithm 4.2), and apply it to images with and without our albedo recovery method. Views of different images are corrected to the same viewpoint using the 3D geometry. The results are presented in Figure 4.18. It shows that the change detection algorithm on our albedo recovered image can detect reasonable changes such as transient or moving objects in the scene (Figure 4.18b), while the results on the original images show that it is heavily polluted by the unwanted shadows (Figure 4.18c).

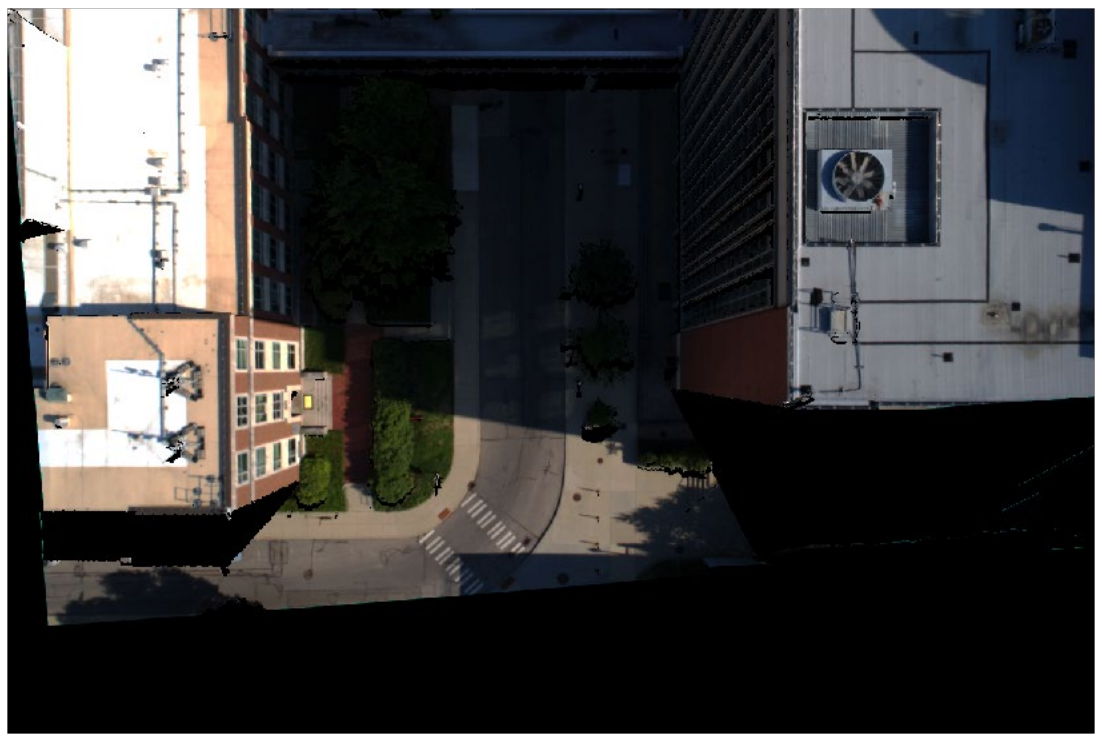

(a) Origin image taken at 6 PM

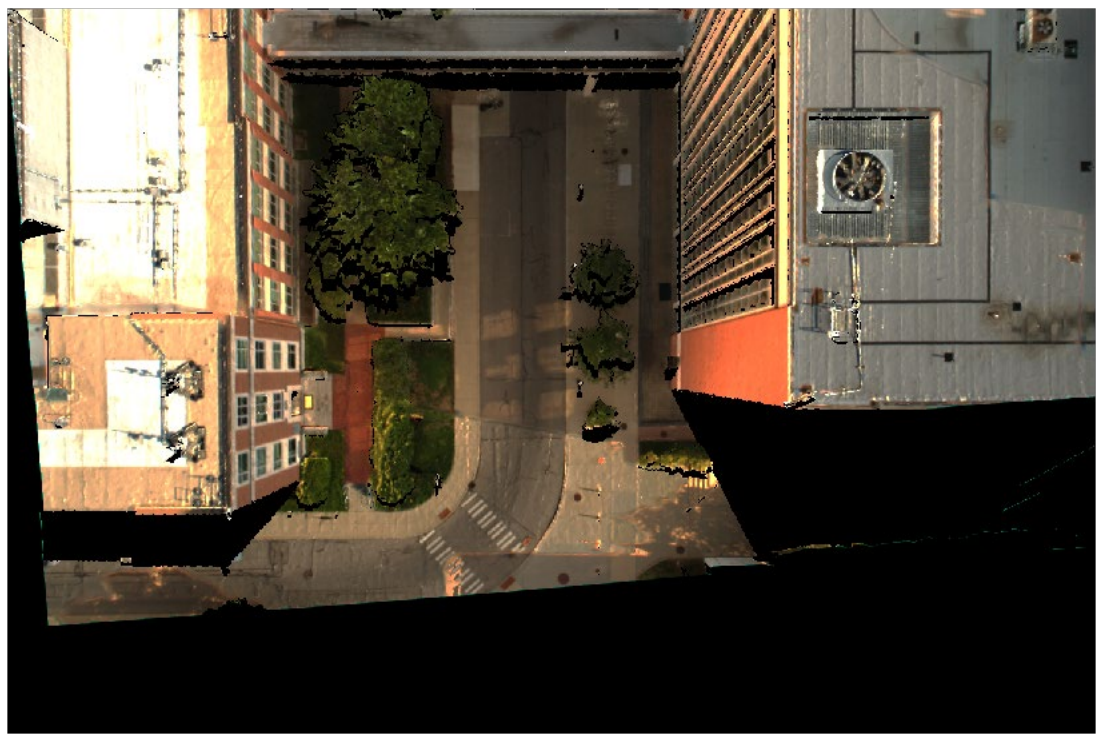

(b) Our albedo recovered image of (a)

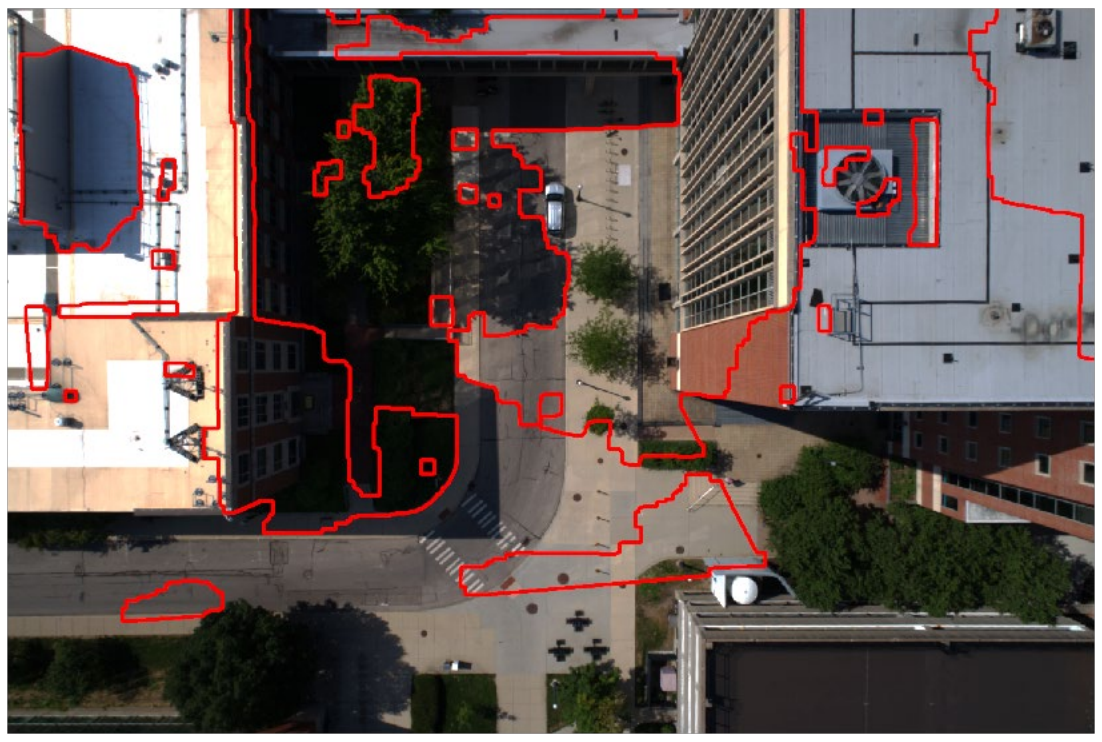

(c) Change detection and origin image taken at 10 AM.

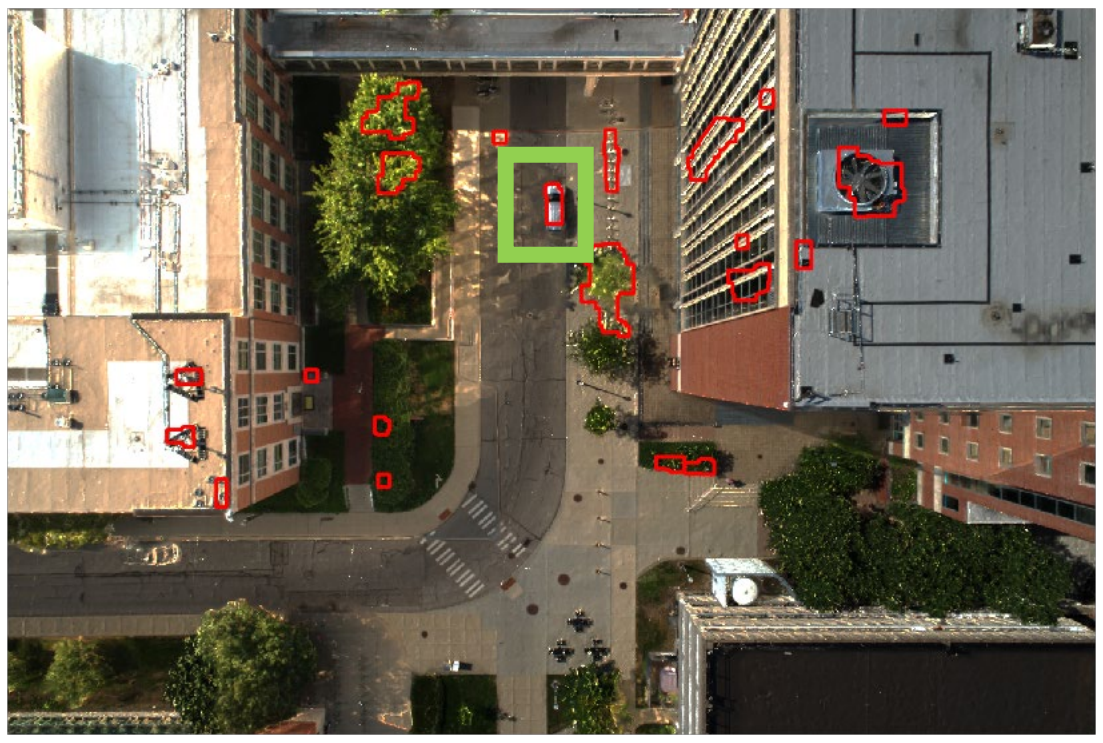

(d) Change detection and our albedo recovered image of (c)

Figure 4.18 Change detection across different times of the day. Change mask from Algorithm 4.2 is visualized as red contours in (c) and (d). All images are geometrically corrected to the same viewpoint. Black regions in (a) and (b) are due to occlusions from view correction. The green box in (d) indicates the changed object between 10 AM and 6 PM.

Algorithm 4.2 A simple change detection method

1. Correct the multi-temporal images to the same viewpoint using 3D model and camera poses.
2. Compute image differencing between the source image and reference image.
3. Convert the difference map into a binary mask by applying a threshold.
4. Apply morphological open & close operations to clean out the binary mask.
5. Output cleaned binary mask.

### 4.8 Conclusion

This paper presents a general albedo recovery approach for photogrammetric images, as an extended work from our earlier work (S. Song and Qin 2022). The core of this approach is a data-agnostic outdoor light modeling, by taking metadata from photogrammetry data collection, the approach directly estimates the sunlight direction, based on which a heterogenous skylight model is estimated by utilizing lit-shadow samples in the image observations and the local geometry. We provide a more comprehensive math framework for light modeling, as well as more comprehensive experimental results demonstrating the effectiveness and scalability of our methods. As compared to existing albedo recovery methods, we show that our proposed method significantly outperforms others under the context of photogrammetric collection both in terms of quantitative metrics (PSNR, SSIM, etc.), qualitative visual comparison, as well as the improvement of downstream applications that the proposed method can drive, including model relighting, feature extraction & matching, dense stereo image matching, and change detection. As compared to many existing works, our approach does not require additional data collection logistics and can be easily scalable. Our future work aims to simplify the procedures and develop more accurate skylight models to cover weathered conditions, as well as more accurate estimation approaches.

# Chapter 5. Conclusion and Future Work

## 5.1 Conclusion

In this dissertation, we advanced the capabilities of scalable 3D reconstruction and mesh processing in large scene photogrammetry by introducing innovative methodologies across three principal chapters.

In Chapter 2, we presented an efficient framework for mesh reconstruction from unstructured point clouds by leveraging the learned visibility from virtual views and traditional graph-cut based mesh generation. A notable advancement is our three-step network, which incorporates depth completion for visibility prediction, and an adaptive visibility weighting in surface determination to improve geometric detail authenticity. This dual-methodological approach has emerged as both efficient and generalizable, outperforming current learning-based and state-of-the-art methods, particularly in large indoor and outdoor scenes.

Chapter 3 addressed the complex task of mesh conflation in full-3D oblique photogrammetric models, introducing a novel approach using virtual cameras and Truncated Signed Distance Field (TSDF). This approach refined the conflation process, producing large, accurate, and seamless 3D site models, thereby enabling enhanced geoscience and environmental applications. Its innovation lies in the use of a panoramic virtual camera field and a non-uniform weighting scheme that respects the quality of individual meshes—advancing the state-of-the-art in full-3D mesh modeling.

Lastly, Chapter 4 proposed a physics-based method for general albedo recovery from aerial photogrammetric images. By using metadata from photogrammetric data collection, we derived a comprehensive lighting model to decouple albedo from shading, improving outdoor scene realism in rendering. This method was validated through extensive experiments and substantiated its potential to enhance various downstream photogrammetric tasks, setting a new benchmark for albedo recovery in photogrammetry.

### 5.2 Limitation

The use of virtual view generators in Chapter 2 relies on the users' decisions, which hinders the workflow from becoming fully automatic. Although in certain applications with similar inputs, it can be streamlined.

In Chapter 3, the use of virtual view samplers faces a similar issue where the suitable scene is limited by the pattern of the virtual view sampler. When applied to different scenarios, it may require customizing the virtual view generator.

Chapter 4 describes a method that requires detailed and accurate geometry, assuming that all illumination remains constant during data collection. This method performs well on man-made structures but struggles with dynamic and thin objects like vegetation.

Moreover, due to our assumption of Lambertian surfaces in Chapter 4, our model is unable to handle complex materials containing metallic, transparent, and shading effects caused by those objects.

## 5.3 Future Work

Future research stemming from this work may include:

**Scalability and Efficiency:** Although our framework in Chapter 2 handles large scenes efficiently, future work could explore incremental learning and parallelized processing to handle dynamic, real-time reconstruction in megacity-scale environments.

**Mesh Conflation Robustness:** Despite the advances in Chapter 3, a deeper investigation into mesh topology and variance in data quality is essential—especially in fusing meshes from heterogeneous sources with different resolutions and sensor modalities.

**Albedo Recovery under Variable Conditions:** Our work in Chapter 4 opens new avenues for research under varying environmental conditions. The adaptability of the albedo recovery process to different weather and lighting conditions requires further exploration, possibly through deep learning techniques that can predict and account for such anomalies.

**Automatization and User-Friendly Solutions:** We plan to simplify the frameworks into more accessible software tools for practitioners, integrating user feedback to ensure practicality across different disciplines and expertise levels.

**Ethical and Ecological Implications:** The digitization of large scenes raises questions about privacy, data security, and environmental impacts. Future work should consider these ethical dimensions, developing guidelines and best practices for data handling and processing.

**Integration with Emerging Technologies:** Integration of this work with emerging technologies like augmented reality (AR), virtual reality (VR), and the metaverse could be explored to enhance user interactivity and visualization capabilities.